\documentclass{article}

 \usepackage[main, final]{neurips_2026}

\usepackage[utf8]{inputenc} 
\usepackage[T1]{fontenc}    
\usepackage{hyperref}       
\usepackage{url}            
\usepackage{booktabs}       
\usepackage{amsfonts}       
\usepackage{nicefrac}       
\usepackage{microtype}      
\usepackage{xcolor}         
\usepackage{amsmath}
\usepackage{tikz}
\usepackage{hyperref}
\usepackage{bm}
\usetikzlibrary{arrows.meta, positioning}
\DeclareMathOperator*{\argmin}{\arg\min}
\DeclareMathOperator*{\argmax}{\arg\max}

\usepackage{graphicx}                        
\usepackage{tikz}
\definecolor{softblue}{RGB}{180,185,230}
\definecolor{softgreen}{RGB}{190,235,190}
\definecolor{softred}{RGB}{240,180,170}
\definecolor{boxgray}{RGB}{235,235,235}
\usetikzlibrary{positioning, arrows.meta}    
\usepackage{booktabs} 
\usepackage{multirow}  
\usepackage{graphicx}  
\usetikzlibrary{fit}
\newtheorem{remark}{Remark}
\title{Amortized Optimal Transport from Sliced Potentials}

\author{%
  Minh-Phuc Truong \\
  Department of Computer Science\\
  Hanoi Univeristy of Science and Technology\\
  Hanoi, Vietnam\\
  \texttt{truongminhphuc08102005@gmail.com} \\
  \And
  Khai Nguyen\thanks{Corresponding author} \\
  Department of Statistics and Data Science\\
  Texas A\&M University \\
  College Station, Texas, USA \\
  \texttt{kbn397@tamu.edu} \\
}

\begin{document}

\maketitle

\setcounter{footnote}{0}

\begin{abstract}
We propose a novel amortized optimization method for predicting optimal transport (OT) plans across multiple pairs of measures by leveraging Kantorovich potentials derived from sliced OT. We introduce two amortization strategies: regression-based amortization (RA-OT) and objective-based amortization (OA-OT). In RA-OT, we formulate a functional regression model that treats Kantorovich potentials from the original OT problem as responses and those obtained from sliced OT as predictors, and estimate these models via least-squares methods. In OA-OT, we estimate the parameters of the functional model by optimizing the Kantorovich dual objective. In both approaches, the predicted OT plan is subsequently recovered from the estimated potentials. As amortized OT methods, both RA-OT and OA-OT enable efficient solutions to repeated OT problems across different measure pairs by reusing information learned from prior instances to rapidly approximate new solutions. Moreover, by exploiting the structure provided by sliced OT, the proposed models are more parsimonious, independent of specific structures of the measures, such as the number of atoms in the discrete case, while achieving high accuracy. We demonstrate the effectiveness of our approaches on tasks including MNIST digit transport, color transfer, supply–demand transportation on spherical data, and mini-batch OT conditional flow matching\footnote{Code for the paper is available at
\url{https://github.com/tmp0810/Sliced-Amortized-OT}.}. 
\end{abstract}

\section{Introduction}
\label{sec:introduction}
Optimal transport (OT)~\citep{villani2003topics,villani2009optimal} has been widely recognized as a fundamental and powerful tool in statistics, machine learning, and data science. It plays a key role across a broad range of applications. For instance, OT has improved the quality of generative models, including autoencoders~\citep{tolstikhin2018Wasserstein,patrini2020sinkhorn}, generative adversarial networks~\citep{arjovsky2017wasserstein,genevay2018learning}, and diffusion and flow-based models~\citep{lipman2023flow,pooladian2023multisample,tong2024improving}, as well as drifting models~\citep{he2026sinkhorn}. It is also widely used in domain adaptation, where it helps establish correspondences between source and target domains~\citep{courty2017joint,courty2016optimal,damodaran2018deepjdot,zhu2024functional}. Beyond these areas, OT has found numerous applications in computational biology~\cite{bunne2023learning,schiebinger2019optimal}, chemistry~\cite{wu2023improving}, image processing~\cite{feydy2017optimal}, fairness~\citep{jiang2020wasserstein}, signal processing~\citep{kolouri2017optimal}, computer graphics~\cite{solomon2016entropic,solomon2015convolutional,bonneel2023survey}, statistical inference~\citep{bernton2019approximate,bernton2019parameter}, and dependency measurement~\citep{catalano2021measuring,catalano2024wasserstein}, among others.

Computational optimal transport (OT)~\citep{peyre2020computational} has emerged as a central yet challenging area within OT. Except for special cases such as OT between Gaussian measures, which admits a closed form solution~\citep{dowson1982frechet}, computing OT is generally expensive. In the discrete setting, where both measures are supported on finite sets, OT can be formulated as a linear programming problem with super cubic complexity in the number of support points~\citep{orlin1997polynomial}. Various approaches have been proposed to mitigate this computational burden, including stochastic approximation with mini-batches~\citep{sommerfeld2019optimal,fatras2020learning,fatras2021unbalanced,nguyen2022improving} and low rank approximations~\citep{scetbon2021low,scetbon2022low}. For general measures, OT is typically solved through its dual formulation, which involves a functional optimization problem over Kantorovich potentials~\citep{villani2009optimal}. This dual problem is constrained and often requires parameterizing the potentials for practical computation~\citep{arjovsky2017wasserstein,makkuva2020optimal}, which can be challenging.

One widely used approach is entropic regularization~\citep{cuturi2013sinkhorn}, which smooths the transport problem by adding an entropy penalty. In the discrete setting, this regularization ensures a unique transport plan that can be efficiently computed via matrix scaling algorithms such as Sinkhorn–Knopp~\citep{sinkhorn1967concerning}, or, equivalently, through iterative Bregman projections~\citep{benamou2015iterative}. Beyond reducing computational complexity to quadratic time, entropic regularization also improves the statistical rates for estimating both the transport cost~\citep{genevay2019sample} and the transport plan~\citep{manole2024plugin,rigollet2025sample}. In the general cases, it removes constraints in the dual formulation, enabling unconstrained optimization~\citep{genevay2016stochastic}. For these reasons, entropic regularized OT is widely used in practice as a surrogate for classical OT.

In many scenarios involving numerous pairs of measures, such as mini-batch generative models~\citep{genevay2018learning,he2026sinkhorn}, dataset comparison~\citep{alvarez2020geometric}, 3D point-cloud autoencoders~\citep{achlioptas2018learning}, point-cloud nearest neighbor classification and regression~\citep{rubner1998metric},  seismic signal analysis~\citep{engquist2014application}, and single-cell perturbation modeling~\citep{bunne2022proximal}, optimal transport (OT) must be solved repeatedly. To accelerate computation in such settings, amortized optimization~\citep{ruishu2017,amos2022tutorial} has been employed to predict OT solutions directly from input measures. Two primary targets arise for amortization: the transportation cost and the transportation plan. Predicting the transportation cost is relatively straightforward, as it reduces to estimating a nonnegative scalar. This direction has been widely explored, for example through learned embeddings or embedding networks~\citep{moosmuller2023linear,courty2018learning,kolouri2021wasserstein,haviv2024wasserstein}, as well as regression onto sliced optimal transport metrics~\citep{nguyen2026fast}. In contrast, amortizing the transportation plan is significantly more challenging, as it involves predicting a joint measure over the input distributions and implicitly determines the cost. To the best of our knowledge, the only existing approach is Meta-OT~\citep{amos2023meta}, which learns to predict entropically regularized Kantorovich potentials from the input measures as a proxy for the  plan.

In the discrete setting, Meta-OT constructs an amortized model, typically a neural network, that maps input measures, represented by their atomic supports and corresponding weights, to a vector whose dimension matches the number of atoms, thereby approximating one of the Kantorovich potentials. In the continuous setting, Meta-OT instead employs a hypernetwork as the amortized model to predict the parameters of a parametric Kantorovich potential. In both cases, Meta-OT adopts an objective-based amortization strategy, where the amortized models are learned by directly optimizing the Kantorovich dual objective. The main limitation of Meta-OT is that it directly takes the raw representation of input measures, hence, its amortized models contains many parameters.  In this work, we propose a new approach to amortizing the transportation plan. The key idea is to use \textit{Kantorovich potentials derived from sliced OT}~\citep{rabin2012wasserstein,nguyen2025introduction} as inputs to the amortized model, rather than the raw input measures. This representation substantially reduces the dimensionality of the input, and consequently lowers the number of parameters required for the amortized model.

\textbf{Contribution.} In summary, our main contributions are three-fold:

1. We introduce a new method to tackle amortized OT problem using sliced OT. In particular, we suggest to predict Kantorovich potentials of the original OT problem from Kantorovich potentials of the corresponding sliced OT problems. Due to the closed-form of sliced OT, the resulting method is computationally fast. Moreover, leveraging expressive sliced OT's potentials, we are able to use parsimonious amortized models with a small number of parameters and are independent of the structure of measures such as the number of atoms of measures in discrete cases. Finally, our proposal can also be seen as the first attempt to reconstruct  OT plan from sliced OT as existing works only form a transporation plan from sliced OT without any notion of approximating the original OT one.

2. We consider two amortization strategies: regression-based amortization (RA-OT) and objective-based amortization (OA-OT). In RA-OT, we formulate a functional regression model, serving as the amortized model, where Kantorovich potentials from the original OT problem are treated as responses and those derived from sliced OT as predictors. The model is then estimated using least-squares methods. In OA-OT, we extend the Meta-OT framework by designing an amortized model that predicts one Kantorovich potential from the sliced OT potentials, and train it by optimizing the Kantorovich dual objective. 

3. While the proposed method is well defined for both discrete and continuous measures, we focus on the discrete setting in this work. This choice is motivated by the fact that discrete OT is more commonly used in practice, whereas computing continuous OT has no exact solutions. We conduct experiments on  MNIST digit transport, color transfer,  spherical supply–demand transportation, and mini-batch OT conditional flow matching to show that the proposed amortized OT approach leads to more favorable prediction of OT plan  than the predicted plan from  Meta-OT and  sliced OT's plans.

\section{Background on Optimal Transport and Entropic Regularization}
\label{sec:background}

We start this section by reviewing the definition of OT, Kantorovich duality, and Kantorovich potentials. Given two measures $(\mu,\nu)$ supported on domains $(\mathcal{X},\mathcal{Y})$ and a ground metric $c:\mathcal{X}\times \mathcal{Y} \to \mathbb{R}_+$, the primal OT~\citep{villani2009optimal} problem is defined as follows:
\begin{align}
\label{eq:OT}
    \pi^\star \in \argmin_{\pi \in \Pi(\mu,\nu)} \int_{\mathcal{X}\times \mathcal{Y}} c(x,y) \mathrm{d} \pi(x,y),
\end{align}
where $\Pi(\mu,\nu)$ is the set of admissible transportation plans (joint measures) between $\mu$ and $\nu$, and $\pi^\star$ is the optimal transportation plan. \eqref{eq:OT} admits a dual problem which is known as Kantorovich duality:
\begin{align}
    \label{eq:dual_OT}
    (f^\star,g^\star) \in \argmax_{(f,g) \in \mathcal{U}_c(\mathcal{X},\mathcal{Y}) } \int_\mathcal{X} f(x) \mathrm{d} \mu(x) + \int_{\mathcal{Y}} g(y) \mathrm{d} \nu(y),
\end{align}
where $\mathcal{U}_c = \{(f,g) \in \mathcal{C}(\mathcal{X}) \times \mathcal{C}(\mathcal{Y})\mid \forall(x,y) \in \mathcal{X}\times \mathcal{Y}, f(x)+g(y) \leq c(x,y)\}$ is the  constraint set, $\mathcal{C}(\mathcal{X})$ and $\mathcal{C}(\mathcal{Y})$ are sets of continuous functions on $\mathcal{X}$ and $\mathcal{Y}$ respectively, and $f^\star,g^\star$ are known as Kantorovich potentials. The connection between $\pi^\star$ and $f^\star,g^\star$ is from complementary slackness condition: $f^\star(x) + g^\star (y) = c(x,y)$ for $\pi^\star$- almost everywhere $(x,y)$.

\textbf{Entropic regularization.} Entropic  OT~\citep{cuturi2013sinkhorn} smoothens the OT problem and it can be conveniently written in a single convex optimization problem as follows:
\begin{align}
    \pi^\star_\epsilon = \argmin_{\pi \in \Pi(\mu,\nu)} \int_{\mathcal{X}\times \mathcal{Y}} c(x,y) \mathrm{d} \pi(x,y) +\epsilon \text{KL}(\pi\mid \mu \otimes \nu),
\end{align}
where $\text{KL}(\pi\mid \mu \otimes \nu) = \int_{\mathcal{X}\times \mathcal{Y}} \left(\log \left(\frac{\mathrm{d} \pi}{\mathrm{d} \mu \otimes \nu}(x,y)\right)\right)\mathrm{d} \pi(x,y)$ and $\mu \otimes \nu$ is the product measure of $\mu$ and $\nu$. The dual problem becomes:
\begin{align}
    (f^\star,g^\star) &\in \argmax_{f \in \mathcal{C}(\mathcal{X}),g\in \mathcal{C}(\mathcal{Y})} \int_\mathcal{X} f(x) \mathrm{d} \mu(x) + \int_{\mathcal{Y}} g(y) \mathrm{d} \nu(y)  \nonumber\\
    &\qquad- \epsilon \log\left(\int_{\mathcal{X}\times \mathcal{Y}} \exp\left(\frac{f(x)+g(y)-c(x,y)}{\epsilon}\right)\mathrm{d}\mu \otimes \nu(x,y)\right), \label{eq:continous_EOT_dual}
\end{align}

For $\epsilon>0$, the optimal transportation plan  is recovered from $(f^\star,g^\star)$ by solving:
$
    \mathrm{d} \pi^\star_\epsilon (x,y) = \exp\left(\frac{f^\star (x) +g^\star(y) -c(x,y)}{\epsilon}\right) \mathrm{d}\mu(x) \mathrm{d}\nu(y),
$
and we can also infer one potential from the other i.e., $g(y) = -\epsilon \log \left(\int_\mathcal{X} \exp\left(\frac{f(x)-c(x,y)}{\epsilon} \mathrm{d} \mu(x)\right)\right)$ and $f(x) = -\epsilon \log \left(\int_\mathcal{Y} \exp\left(\frac{g(y)-c(x,y)}{\epsilon} \mathrm{d} \nu(y)\right)\right)$.

\textbf{Discrete Case.} When $\mu= \sum_{i=1}^n\alpha_i \delta_{x_i}$ and $\nu= \sum_{j=1}^m \beta_j \delta_{y_j}$ with $\sum_{i=1}^n \alpha_i = \sum_{j=1}^m \beta_j$ and $\alpha_i>0,\beta_j>0 \forall i,j$, the entropic OT problem becomes:
\begin{align}
    P^\star_\epsilon = \argmin_{P \in \Gamma(\boldsymbol{\alpha},\boldsymbol{\beta})}\langle C,P\rangle -\epsilon H(P),
\end{align}
where $\Gamma(\boldsymbol{\alpha},\boldsymbol{\beta}) = \{P \in \mathbb{R}_+^{n \times m} \mid P\mathbf{1} = \boldsymbol{\alpha}, P^\top \mathbf{1} = \boldsymbol{\beta}\}$ is the set of discrete plans, $H(P) = -\sum_{i=1}^n \sum_{j=1}^m P_{ij}\log P_{ij}$ is the entropy, and $C_{ij}=c(x_i,y_j)$. The entropic OT dual becomes:
\begin{align}
    (\mathbf{f}^\star,\mathbf{g}^\star ) \in \argmax_{\mathbf{f}\in \mathbb{R}^n,\mathbf{g} \in \mathbb{R}^m} \langle \mathbf{f},\boldsymbol{\alpha}\rangle +\langle \mathbf{g},\boldsymbol{\beta}\rangle -\epsilon\sum_{i=1}^n \sum_{j=1}^m \exp\left(\frac{\mathbf{f}_i+\mathbf{g}_j - C_{ij}}{\epsilon}\right).\label{eq:discrete_EOT_dual}
\end{align}
We can recover a discrete plan from any pairs $(\mathbf{f},\mathbf{g})$ as follows:
\begin{align}
    P_{ij}^\star= \exp\left(\frac{\mathbf{f}_i^\star+\mathbf{g}_j^\star - C_{ij}}{\epsilon}\right).
\end{align}
The mapping between the dual can be rewritten as:
\begin{align}
\label{eq:update_g}
    &\mathbf{g} = \epsilon \log \boldsymbol{\alpha} - \epsilon \log \left(\exp(-C^\top/\epsilon)\exp(\mathbf{f}/\epsilon)\right), \\
    &\mathbf{f} = \epsilon \log \boldsymbol{\beta} - \epsilon \log \left(\exp(-C/\epsilon)\exp(\mathbf{g}/\epsilon)\right)\label{eq:update_f}.
\end{align}
The Sinkhorn algorithm~\citep{cuturi2013sinkhorn} is an efficient iterative method for solving the entropic optimal transport problem. It alternates between updating the dual potentials $\mathbf{f}$ and $\mathbf{g}$ according to~\eqref{eq:update_g} and~\eqref{eq:update_f} until the updates converge. A key property of entropic OT is that knowing only one of the optimal Kantorovich potentials, $\mathbf{f}^\star$ or $\mathbf{g}^\star$, is sufficient: the other potential can be directly computed from it, allowing the optimal transport plan to be fully determined. The time complexity of Sinkhorn algorithm is $\mathcal{O}(nm)$~\citep{altschuler2017near} compared to $\mathcal{O}((n+m)^3 \log (n+m))$~\citep{peyre2020computational} of the conventional OT.
\section{Amortized Optimization of Optimal Transport}
\label{sec:AOT}
In this section, we first review the amortized optimal transport (OT) problem in Section~\ref{subsec:AOT_Review}, including the Meta-OT framework, as well as the definitions of amortized models and amortized losses. We then introduce two new amortization strategies based on sliced OT potentials: regression-based amortized OT (RA-OT) and objective-based amortized OT (OA-OT), and discuss their properties in Section~\ref{subsec:AOT_from_slices}.

\subsection{Amortized Optimal Transport Problem}
\label{subsec:AOT_Review}

In the amortized OT problem, we observe multiple pairs of probability measures, potentially associated with different ground cost functions. We assume that these triplets are drawn from an unknown meta-distribution $\mathcal{D}$, yielding observations $(\mu_1,\nu_1,c_1), \ldots, (\mu_N,\nu_N,c_N)$ ($N>1$). Our objective is not only to compute entropic optimal transport for the given $N$ pairs, but also to leverage these observations to accelerate computation for future pairs $(\mu_1',\nu_1',c_1'), \ldots, (\mu_M',\nu_M',c_M')$ ($M>1$). Meta-OT~\citep{amos2023meta} provides a solution by adopting amortized optimization. We now discuss how Meta-OT address the problem by reviewing the definition of the amortized model and amortized loss.

\textbf{Amortized model.} For each optimal transport (OT) problem, denoted $(\mu_i, \nu_i, c_i)$, the goal is to obtain one of the corresponding Kantorovich potentials, e.g., $f^\star[\mu_i, \nu_i, c_i]$. An \emph{amortized model} is a mapping that takes a pair of measures and the associated cost as input and predicts a Kantorovich potential. Formally, a parametric mapping $\hat{f}_\phi$ is considered in Meta-OT~\citep{amos2023meta}, where $\phi$ belongs to a parameter space $\Phi$. The model is trained such that
\[
\hat{f}_\phi[\mu, \nu, c] \approx f^\star[\mu, \nu, c] \quad \text{for $\mathcal{D}$-almost every $(\mu, \nu, c)$}.
\]
When $\mu$ and $\nu$ are discrete and $c$ is fixed across all realizations from $\mathcal{D}$, the inputs to $\hat{f}_\phi$ can be written as $(\boldsymbol{\alpha}, X)$ and $(\boldsymbol{\beta}, Y)$, representing the weights and atoms of $\mu$ and $\nu$, respectively. In Meta-OT, the authors consider cases where the atoms are fixed, so only the weights $\boldsymbol{\alpha}$ and $\boldsymbol{\beta}$ are used as inputs. In this setting, $\hat{f}_\phi$ can be implemented as a neural network (e.g., an MLP) that takes two vectors of sizes $n$ and $m$ as input and outputs a vector of size $n$. Note that Meta-OT requires the number of atoms, $n$ and $m$, to be the same across all realizations of $\mathcal{D}$, since designing a neural network that accommodates variable input sizes is nontrivial.

\textbf{Amortized loss.} Meta-OT adapts an objective-based optimization approach to estimate the amortized model. In particular, it solves the following optimization problem:
\begin{align}
    \min_{\phi \in \Phi } \mathbb{E}_{(\mu,\nu,c) \sim \mathcal{D}} [J (\hat{f}_\phi[\mu, \nu, c];\mu,\nu,c)],
\end{align}
where $J$ is the Kantorovich dual objective in~\eqref{eq:discrete_EOT_dual} with the plugged in form of the other potential using~\eqref{eq:update_g}. By replacing the expectation with the finite sum across samples  $(\mu_1,\nu_1,c_1), \ldots, (\mu_N,\nu_N,c_N)$,  we obtain the empirical objective, which can be optimized using gradient descent algorithms or stochastic gradient algorithms for a large-scale dataset.

\subsection{Amortized Optimal Transport from Sliced Potentials}
\label{subsec:AOT_from_slices}

We start by reviewing some essential definitions, such as one-dimensional OT plan, sliced potentials, and projection functions in sliced OT.

\textbf{One-dimensional Optimal Transport.} In one-dimension, when $c=h(x-y)$ with $h$ is a strictly convex function, OT between $\mu$ and $\nu$ admits a closed-form~\citep{santambrogio2015optimal}:
\begin{align}
    \pi^\star =(F_{\mu}^{-1},F_{\nu}^{-1})\sharp \mathcal{U}([0,1]),
\end{align}
where $\mathcal{U}([0,1])$ is the uniform distribution on the interval $[0,1]$,  $F_{\mu}^{-1}$ and $F_{\nu}^{-1}$ are quantile functions of $\mu$ and $\nu$ respectively. We then can access to the Kantorovich potential using complementary slackness condition:
\begin{align}
    f^\star (F_{\mu}^{-1}(u)) +  g^\star (F_{\nu}^{-1}(u)) = h(F_{\mu}^{-1}(u)-F_{\nu}^{-1}(u)),
\end{align}
for $\mathcal{U}([0,1])$-almost every $u \in [0,1]$, this fact allows the OT plan and Kantorovich potentials in the discrete case to be computed extremely efficiently, with a time complexity of $\mathcal{O}((n+m)\log(n+m))$, where $n$ and $m$ denote the number of atoms in the two measures. A detailed algorithm for computing the OT plan and Kantorovich potentials in one-dimensional discrete settings can be found in Algorithm 1 of~\citep{sejourne2022faster}. We also note that, alternatively, Kantorovich potentials in discrete cases can be obtained by taking the gradients of the optimal transport cost with respect to the weights of the two measures~\citep{cuturi2014fast}. 

\textbf{Sliced Optimal Transport.} Sliced OT~\citep{rabin2012wasserstein,nguyen2025introduction} leverages the closed-form solution of optimal transport in one dimension through projections. Given the geometry of the space of atoms $\mathcal{X}\cup \mathcal{Y}$ of the measures and the ground metric $c$, a projection function
$
P_\theta^c:\mathcal{X}\cup \mathcal{Y} \to \mathcal{Z}
$
maps the atoms to a one-dimensional space $\mathcal{Z}$, e.g., the real line $\mathbb{R}$, where $\theta \in \Theta$ denotes the projection parameters. Designing effective $P_\theta^c$ remains an active area of research. For instance, in Euclidean spaces ($\mathcal{X}\cup \mathcal{Y} \subset \mathbb{R}^d$) with the Euclidean ground metric $c(x,y)=\|x-y\|_2$, one can use linear projections $P_\theta^c(x) = \langle \theta, x\rangle$ with $\theta \in \mathbb{S}^{d-1}$ (the unit $(d-1)$-sphere). In spherical settings ($\mathcal{X}\cup \mathcal{Y} \subset \mathbb{S}^{d-1}$), spherical~\citep{bonet2022spherical,quellmalz2023sliced} or stereographic~\citep{tran2024stereographic} projections are applicable. Extensions to other domains, such as functions~\citep{garrett2024validating}, manifolds, and products of manifolds~\citep{bonet2024sliced,nguyen2026summarizing}, have also been explored. Using the projection function, sliced OT projects $\mu$ and $\nu$ into $P_\theta^c \sharp \mu$ and $P_\theta^c \sharp \nu$, then obtains the 1D OT plan between them. Recent works have explored lifting the 1D OT plan back to the original space~\citep{mahey2023fast,liu2025expected,tanguy2025sliced}. However, these methods focus on fast approximations and are not intended to recover the exact original OT. Furthermore, they do not operate in an amortized OT setting, as they are applied at the level of individual measure pairs.

\begin{figure}
    \centering
    \includegraphics[width=1\linewidth]{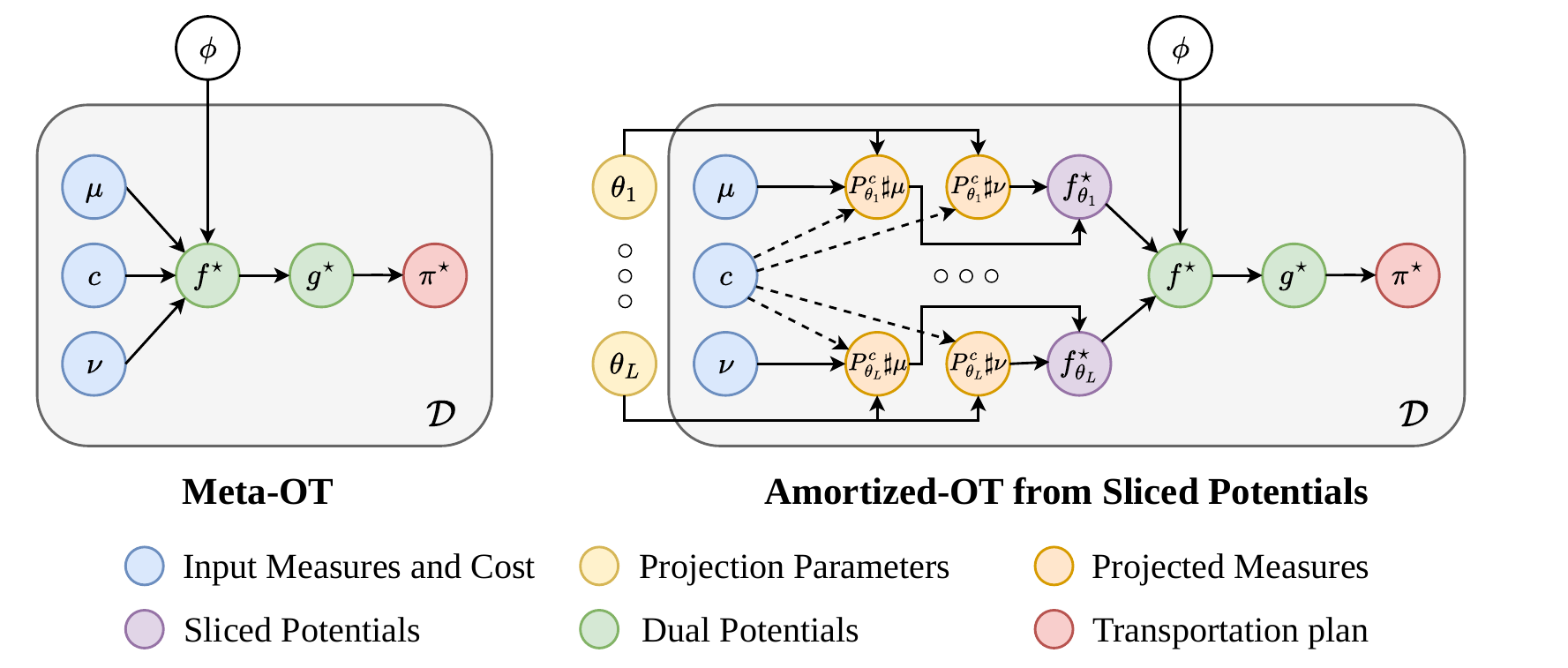}
    \vspace{-2em}
    \caption{{Amortized-OT uses sliced potentials to predict original dual potentials.}}
    \label{fig:AOT}
    \vspace{-1em}
\end{figure}

\textbf{Amortized model with sliced potentials.} Given a set of projection parameters $\theta_1,\ldots,\theta_L$ ($L\geq 1$ is referred to as the number of projections),  which can be created using Monte Carlo sampling~\citep{bonneel2015sliced} or quasi Monte Carlo sampling~\citep{nguyen2024quasimonte}, we are able to obtain a set of sliced potentials $f^\star_{\theta_1}[\mu,\nu,c],\ldots,f^\star_{\theta_L}[\mu,\nu,c]$ using closed-form of one-dimensional OT  problems between $P_{\theta_l}^c \sharp \mu$ and $P_{\theta_l}^c \sharp \nu$ for $l=1,\ldots,L$. We then define an amortized model as follows:
\begin{align}
    \hat{f}_\phi[\mu, \nu, c] = \gamma_\phi (f^\star_{\theta_1}[\mu,\nu,c],\ldots,f^\star_{\theta_L}[\mu,\nu,c]),
\end{align}
where $\gamma_\phi$ is an operator that maps from the space of $L$ functions in one dimension to the space of functions in the dimension of measures $\mu$ and $\nu$.  Since  $f^\star_{\theta_1}[\mu,\nu,c],\ldots,f^\star_{\theta_L}[\mu,\nu,c]$ are "handcraft features" with a lot of information about transportation between $\mu$ and $\nu$ through one-dimensional slices, we propose to consider a simple linear model as follows:
$
     \hat{f}_{\boldsymbol{\omega}}[\mu, \nu, c](x) =\sum_{l=1}^L \omega_l f^\star_{\theta_l}[\mu,\nu,c](P_{\theta_l}^c(x)),
$
where $\boldsymbol{\omega} \in \mathbb{R}^L$ is the linear coefficients. We recall that the above construction of the amortized model is well-defined for both discrete and continuous cases. Importantly, the parameter $\boldsymbol{\omega}$ or $\phi$ of the proposed amortized model is independent of the input measures (e.g., their number of atoms), unlike in Meta-OT. Therefore, we are able to handle settings where the number of atoms varies.

\textbf{Regression-based amortization.} We now discuss the first amortization strategy: regression-based amortized OT (RA-OT). In this strategy, we assume that we can afford the computation of (entropic) OT during training. In particular, we solve the following optimization problem:
\begin{align}
    \min_{\boldsymbol{\omega} \in \mathbb{R}^L} \mathbb{E}_{(\mu,\nu,c)\sim \mathcal{D}}[\| \hat{f}_{\boldsymbol{\omega}}[\mu, \nu, c] -f^\star[\mu, \nu, c] \|_2^2],
\end{align}
where $\|\cdot\|_2^2$ denotes the functional $\mathbb{L}_2$ norm. When $\mu$ is discrete with $n$ atoms $x_1,\ldots,x_n$, the above optimization problem admits a closed-form. In particular, let $X_{\theta_l}[\mu,\nu,c] = (f^\star_{\theta_l}[\mu,\nu,c](P_{\theta_l}^c(x_1)),\ldots,f^\star_{\theta_l}[\mu,\nu,c](P_{\theta_l}^c(x_n))^\top \in \mathbb{R}^n$, $X_\Theta [\mu,\nu,c]=  (X_{\theta_1}[\mu,\nu,c],\ldots,X_{\theta_L}[\mu,\nu,c]) \in \mathbb{R}^{n\times L}$, and $Y[\mu,\nu,c] = (f^\star[\mu,\nu,c](x_1),\ldots,f^\star[\mu,\nu,c](x_n))^\top \in \mathbb{R}^n$, we can rewrite the optimization problem as:
\begin{align}
    \min_{\boldsymbol{\omega} \in \mathbb{R}^L} \mathbb{E}_{(\mu,\nu,c)\sim \mathcal{D}}[\|  X_{\Theta}[\mu,\nu,c]\boldsymbol{\omega}-Y[\mu,\nu,c]\|_2^2],
\end{align}
which has the optimal solution is the root of the following equation:
\begin{align}
\label{eq:normal_equation}
    \mathbb{E}[X_{\Theta}[\mu,\nu,c]^\top X_{\Theta}[\mu,\nu,c]] \boldsymbol{\omega}=\mathbb{E}[X_{\Theta}[\mu,\nu,c]^\top Y[\mu,\nu,c]],
\end{align}
which has the closed-form solution:
\begin{align}
\label{eq:closed_form}
    \boldsymbol{\omega}^\star = \left(\mathbb{E}[X_{\Theta}[\mu,\nu,c]^\top X_{\Theta}[\mu,\nu,c]]\right)^{-1} \mathbb{E}[X_{\Theta}[\mu,\nu,c]^\top Y[\mu,\nu,c]],
\end{align}
where the expectation is under $\mathcal{D}$. When we observe a finite set of $(\mu_1,\nu_1,c_1), \ldots, (\mu_N,\nu_N,c_N)$, we can replace the expectation by the finite sum $\frac{1}{N}\sum_{i=1}^N$  as a Monte Carlo estimate for tractable computation. In practice, we can use other linear solvers of~\eqref{eq:normal_equation} rather than~\ref{eq:closed_form} for computational stability. The time complexity for training RA-OT involves the complexity for computing sliced potentials $\mathcal{O}(ML (n+m) (\log (n+m)+d) )$ ($P_\theta^c$ has the time complexity of $\mathcal{O}(Ld (n+m))$), computing ground truth potential $\mathcal{O}(Nnm)$ for entropic OT or $\mathcal{O}((n+m)^3 \log (n+m))$ for OT, and computing the optimal coefficient $\mathcal{O}(NnL^2+L^3)$. For prediction on a new set $(\mu_1',\nu_1',c_1'), \ldots, (\mu_M',\nu_M',c_M')$, the time complexity of RA-OT involves only the complexity for computing sliced potentials $\mathcal{O}(ML (n+m) (\log (n+m)+d) )$  and applying linear prediction $\mathcal{O}(MnL)$. In the case of conventional OT, we can set up another regression problem for the other potential $g^\star$, while predicting $f^\star$ is sufficient for entropic OT as discussed in Section~\ref{sec:background}.

\textbf{Objective-based amortization.} For the objective-based amortized OT (OA-OT) strategy,  we follow the same training scheme as Meta-OT with the new proposed amortized model. We consider the following optimization problem:
\begin{align}
    \min_{\boldsymbol{\omega} \in \mathbb{R}^L} \mathbb{E}_{(\mu,\nu,c) \sim \mathcal{D}} [J (\hat{f}_{\boldsymbol{\omega}}[\mu, \nu, c];\mu,\nu,c)],
\end{align}
where $J$ is either the Kantorovich dual objective  in~\eqref{eq:continous_EOT_dual} or in~\eqref{eq:discrete_EOT_dual} in the plugged-in form of the other potential. Similar to Meta-OT, OA-OT is designed to address the entropic OT problem only. We also use a gradient-based algorithm for finding an estimate of $\boldsymbol{\omega}$  in OA-OT. For a set of $N$ samples from $\mathcal{D}$, after computing sliced potentials with the time complexity of $\mathcal{O}(NL (n+m) (\log (n+m)+d) )$, an estimate of gradient for $\boldsymbol{\omega} $ has $\mathcal{O}(Nmn)$  in time complexity. The time complexity is then scaled by $T>0$ iterations of gradient updates. Compared to Meta-OT, OA-OT has a much smaller number of parameters; hence, it enjoys faster training. Given an estimate of $\boldsymbol{\omega}$, the time complexity of OA-OT during inference time is the same as RA-OT.

\begin{remark}[Amortization Error and Amortization Gap]
    As we considered linear amortized models, the amortization error $\mathbb{E}_{(\mu,\nu,c)\sim \mathcal{D}}[\| \hat{f}_{\boldsymbol{\omega}}[\mu, \nu, c] -f^\star[\mu, \nu, c] \|_2^2]$  and the amortization gap $\mathbb{E}_{(\mu,\nu,c) \sim \mathcal{D}} [( J (\hat{f}_{\boldsymbol{\omega}}[\mu, \nu, c];\mu,\nu,c)] -J (f^\star[\mu, \nu, c];\mu,\nu,c))^2]$ are minimized when $f^\star[\mu, \nu, c]$ lives in the linear spans of $f^\star_{\theta_1}[\mu,\nu,c],\ldots,f^\star_{\theta_L}[\mu,\nu,c]$ for $\mathcal{D}$-almost every $(\mu, \nu, c)$. Therefore, increasing the number of projections $L$ can potentially enlarge the span generated by the sliced potentials, thereby improving the expressivity of the amortized model.
\end{remark}


\begin{figure}[!t]
    \centering
    \begin{tabular}{cc}
    \begin{minipage}{0.45\textwidth}
        \centering
        \begin{tikzpicture}
            \node[inner sep=0pt] (img3) at (0,0)
                {\includegraphics[width=0.87\linewidth]{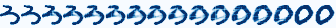}};
            \node[above=0.1cm of img3.north, font=\normalsize]
                {\textbf{Sinkhorn} \textcolor{gray}{(converged, ground-truth)}};
            \coordinate (bl3) at ([xshift=0.3cm,  yshift=-0.1cm]img3.south west);
            \coordinate (br3) at ([xshift=-0.3cm, yshift=-0.1cm]img3.south east);
            \coordinate (bc3) at ([yshift=-0.1cm]img3.south);
            \draw[<->, >=Latex, thick] (bl3) -- (br3);
            \draw[thick] (bc3) -- ([yshift=0.15cm]bc3);
            \node[left=0.05cm  of bl3] {$\alpha_0$};
            \node[below=0cm    of bc3] {$\alpha_1$};
            \node[right=0.05cm of br3] {$\alpha_2$};
        \end{tikzpicture}
    \end{minipage}
    &
    \begin{minipage}{0.45\textwidth}
        \centering
        \begin{tikzpicture}
            \node[inner sep=0pt] (img3) at (0,0)
                {\includegraphics[width=0.87\linewidth]{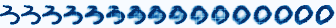}};
            \node[above=0.1cm of img3.north, font=\normalsize]
                {\textbf{Meta-OT}};
            \coordinate (bl3) at ([xshift=0.3cm,  yshift=-0.1cm]img3.south west);
            \coordinate (br3) at ([xshift=-0.3cm, yshift=-0.1cm]img3.south east);
            \coordinate (bc3) at ([yshift=-0.1cm]img3.south);
            \draw[<->, >=Latex, thick] (bl3) -- (br3);
            \draw[thick] (bc3) -- ([yshift=0.15cm]bc3);
            \node[left=0.05cm  of bl3] {$\alpha_0$};
            \node[below=0cm    of bc3] {$\alpha_1$};
            \node[right=0.05cm of br3] {$\alpha_2$};
        \end{tikzpicture}
    \end{minipage}
\\
    \begin{minipage}{0.45\textwidth}
        \centering
        \begin{tikzpicture}
            \node[inner sep=0pt] (img1) at (0,0)
                {\includegraphics[width=0.87\linewidth]{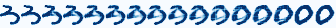}};
            \node[above=0.1cm of img1.north, font=\normalsize]
                {\textbf{RA-OT}};
            \coordinate (bl1) at ([xshift=0.3cm,  yshift=-0.1cm]img1.south west);
            \coordinate (br1) at ([xshift=-0.3cm, yshift=-0.1cm]img1.south east);
            \coordinate (bc1) at ([yshift=-0.1cm]img1.south);
            \draw[<->, >=Latex, thick] (bl1) -- (br1);
            \draw[thick] (bc1) -- ([yshift=0.15cm]bc1);
            \node[left=0.05cm  of bl1] {$\alpha_0$};
            \node[below=0cm    of bc1] {$\alpha_1$};
            \node[right=0.05cm of br1] {$\alpha_2$};
        \end{tikzpicture}
    \end{minipage}
    &
    \begin{minipage}{0.45\textwidth}
        \centering
        \begin{tikzpicture}
            \node[inner sep=0pt] (img2) at (0,0)
                {\includegraphics[width=0.87\linewidth]{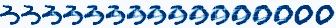}};
            \node[above=0.1cm of img2.north, font=\normalsize]
                {\textbf{OA-OT}};
            \coordinate (bl2) at ([xshift=0.3cm,  yshift=-0.1cm]img2.south west);
            \coordinate (br2) at ([xshift=-0.3cm, yshift=-0.1cm]img2.south east);
            \coordinate (bc2) at ([yshift=-0.1cm]img2.south);
            \draw[<->, >=Latex, thick] (bl2) -- (br2);
            \draw[thick] (bc2) -- ([yshift=0.15cm]bc2);
            \node[left=0.05cm  of bl2] {$\alpha_0$};
            \node[below=0cm    of bc2] {$\alpha_1$};
            \node[right=0.05cm of br2] {$\alpha_2$};
        \end{tikzpicture}
    \end{minipage}

\end{tabular}
    \vspace{-0.3em}
    \caption{ \footnotesize {Wasserstein interpolations between MNIST test digits. (Top) Sinkhorn ground truth
    run to convergence and baseline Meta-OT. (Bottom) RA-OT and
    OA-OT ($M{=}50$, $L{=}100$). Both proposed methods produce interpolation sequences nearly
    identical to the ground truth.}}
    \label{fig:mnist_interp}
     \vspace{-0.5em}
\end{figure}
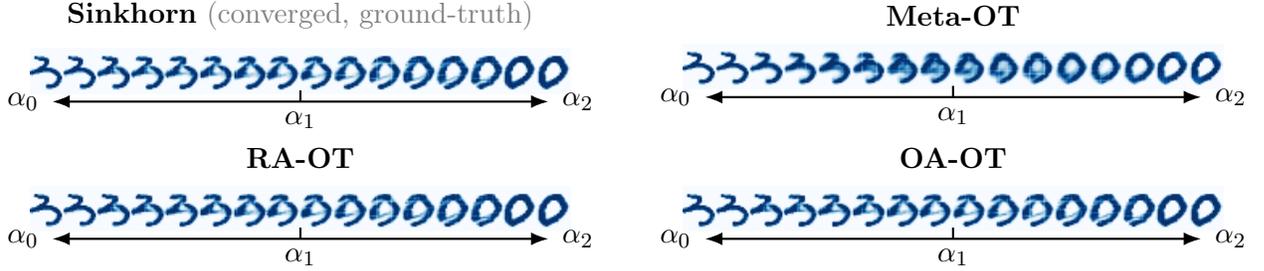

\section{Experiments}
\label{sec:experiments}

Following the experimental pipeline in \citep{amos2023meta}, we evaluate the practical effectiveness of our proposed methods, RA-OT and OA-OT, across three representative settings that span a range of discrete optimal transport applications. First, in \textbf{MNIST grayscale transport}, we compute Wasserstein interpolations between digit images defined on a 2D Euclidean grid. Second, in \textbf{spherical supply–demand transport}, we consider a large-scale, non-Euclidean setting that maps global landmass distributions to population density over a sphere. Finally, in \textbf{color transfer}, we align image color palettes within a 3D RGB space.
For these tasks, all methods are quantitatively assessed using the Root Mean Square Error (RMSE) of the predicted transport plan relative to the converged Sinkhorn ground truth, alongside total training and per-pair inference times. We compare our approaches against three recent baselines: Meta-OT \citep{amos2023meta}, which predicts the plan via an MLP applied to raw measure weights, min-SWGG \citep{mahey2023fast}, a training-free baseline, and Min-STP \citep{liu2025efficient}, which learns an optimized projection direction. To ensure a rigorous evaluation, all trainable methods share an identical pool of $1{,}000$ measure pairs per task with a 70/30 train-test split. Additionally, we evaluate the performance of our methods when applied to mini-batch OT conditional flow matching, focusing on low-dimensional 2D toy datasets to demonstrate accelerated pairing in generative modeling. Since this task deviates from the standard plan-prediction pipeline, we instead measure the training time and the NPE, benchmarking against Independent CFM (I-CFM) and OT-CFM \citep{tong2024improving}. Across all experiments, the number of sliced projections is fixed at $L{=}100$. For the first three standard OT tasks, we report results for varying training sizes $M \in \{10, 20, 50, 200\}$, focusing on $M{=}50$ for qualitative visualizations. In contrast, for the OT-CFM task, we maintain a fixed pre-training size of $M{=}50$. Experiment details are provided in Appendix~\ref{app:impl}. Additional ablation results on the number of projections $L$ are given in Appendix~\ref{Vary_L}, and further visualizations for all tasks are in Appendix~\ref{ab:add_vis}. We also provide additional experiments in Appendix~\ref{sec:additional_experiments}, including evaluations on measures with varying numbers of atoms, heterogeneous measure groups, Sinkhorn warm-starting, distribution shift, marginal feasibility and rounding, and the training-inference time trade-off.

\subsection{MNIST}

Pairs of MNIST digit images offer a standard test for amortized OT, sharing a common $28{\times}28$ grid geometry but varying in mass distributions. Each image is modeled as a discrete measure where pixel intensities dictate the weights of the $n{=}784$ atoms. The transport cost is the squared Euclidean distance between coordinates, and the regularization parameter is set to $\varepsilon{=}0.1$.  As shown in Table~\ref{tab:mnist_vary_m}, RA-OT and OA-OT remain accurate even with very limited training data ($M=10,20$), outperforming Meta-OT in this low-data regime. They also train significantly faster due to their lightweight parameterization. In contrast, single-projection baselines such as Min-STP and min-SWGG struggle to capture the spatial structure of image transport. This improvement is also reflected in Figure~\ref{fig:mnist_interp} (Appendix \ref{ab:add_vis}), where the interpolations better match the Sinkhorn reference.

\begin{table}[!t]
\centering
\caption{ \footnotesize {RMSE, training and inference time on the \textbf{MNIST} grayscale transport task across varying number of training data $M$ ($\varepsilon{=}0.1$, $N{=}300$ test pairs, $L{=}100$). The results are reported as the mean $\pm$ standard deviation. The label ``no train'' indicates that no prior training is required. \textbf{Bold} values denote the best performance in each block.}}
\resizebox{\linewidth}{!}{%
\begin{tabular}{@{} c l c c c @{\hspace{1.5em}} c l c c c @{}}
\toprule
$M$ & Method & RMSE\,($\times 10^{-6}$, $\downarrow$) & Train\,(s) & Infer\,(ms,$\downarrow$) & 
$M$ & Method & RMSE\,($\times 10^{-6}$, $\downarrow$) & Train\,(s) & Infer\,(ms,$\downarrow$) \\
\cmidrule(r){1-5} \cmidrule(l){6-10}

\multirow{5}{*}{10} 
 & Meta-OT \citep{amos2023meta} & $16.16 \pm 4.90$ & $38.05$ & $\mathbf{2.46 \pm 0.20}$ & 
\multirow{5}{*}{50} 
 & Meta-OT \citep{amos2023meta} & $15.54 \pm 4.74$ & $37.11$ & $\mathbf{2.39 \pm 0.29}$ \\
 & Min-STP \citep{liu2025efficient}                     & $93.51 \pm 10.18$ & $199.52$ & $6.18 \pm 0.37$ & 
 & Min-STP \citep{liu2025efficient}                     & $93.50 \pm 10.18$ & $198.01$ & $6.25 \pm 0.48$ \\
 & min-SWGG \citep{mahey2023fast}                     & $89.85 \pm 8.64$ & no train & $10.01 \pm 0.73$ & 
 & min-SWGG \citep{mahey2023fast}                     & $89.85 \pm 8.64$ & no train & $9.72 \pm 0.41$ \\
 & \textbf{RA-OT}  & $8.23 \pm 3.55$ & $\mathbf{1.36}$ & $44.33 \pm 2.91$ & 
 & \textbf{RA-OT}  & $7.77 \pm 3.06$ & $\mathbf{3.03}$ & $39.36 \pm 3.61$ \\
 & \textbf{OA-OT}  & $\mathbf{6.16 \pm 2.53}$ & $16.65$ & $42.54 \pm 2.95$ & 
 & \textbf{OA-OT}  & $\mathbf{6.02 \pm 2.52}$ & $15.78$ & $38.92 \pm 2.23$ \\
\cmidrule(r){1-5} \cmidrule(l){6-10}

\multirow{5}{*}{20} 
 & Meta-OT \citep{amos2023meta} & $16.68 \pm 4.98$ & $37.27$ & $\mathbf{2.44 \pm 0.39}$ & 
\multirow{5}{*}{200} 
 & Meta-OT \citep{amos2023meta} & $14.12 \pm 4.70$ & $37.22$ & $\mathbf{2.37 \pm 0.40}$ \\
 & Min-STP \citep{liu2025efficient}                     & $93.51 \pm 10.18$ & $198.56$ & $6.21 \pm 0.41$ & 
 & Min-STP \citep{liu2025efficient}                     & $93.51 \pm 10.18$ & $197.86$ & $6.17 \pm 0.49$ \\
 & min-SWGG \citep{mahey2023fast}                     & $89.85 \pm 8.64$ & no train & $9.70 \pm 0.47$ & 
 & min-SWGG \citep{mahey2023fast}                     & $89.85 \pm 8.64$ & no train & $9.73 \pm 0.61$ \\
 & \textbf{RA-OT}  & $7.49 \pm 3.19$ & $\mathbf{1.44}$ & $40.29 \pm 2.74$ & 
 & \textbf{RA-OT}  & $7.74 \pm 3.07$ & $\mathbf{11.12}$ & $40.30 \pm 4.39$ \\
 & \textbf{OA-OT}  & $\mathbf{6.04 \pm 2.50}$ & $15.60$ & $40.65 \pm 2.95$ & 
 & \textbf{OA-OT}  & $\mathbf{6.11 \pm 2.54}$ & $16.39$ & $40.48 \pm 4.16$ \\
\bottomrule
\end{tabular}%
}

  \vspace{-1em} 
\label{tab:mnist_vary_m}
\end{table}

\subsection{Spherical Transport}






\begin{figure}[!t]
    \centering
    \begin{tabular}{cc}
    \textbf{Sinkhorn} \textcolor{gray}{(converged, ground-truth)}&  \textbf{Meta-OT}  \vspace{-2em} 
         \\
         \includegraphics[width=0.45\textwidth]{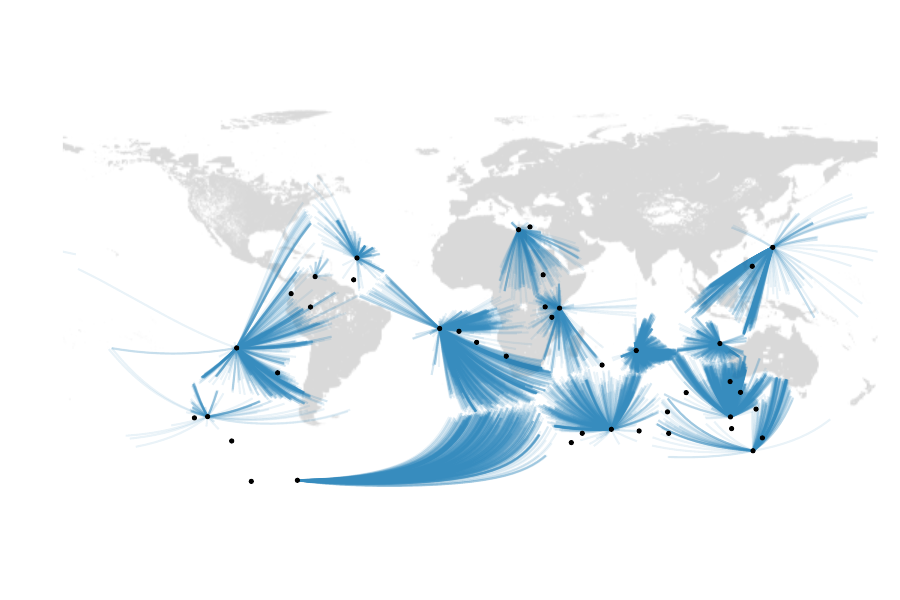} & \includegraphics[width=0.45\textwidth]{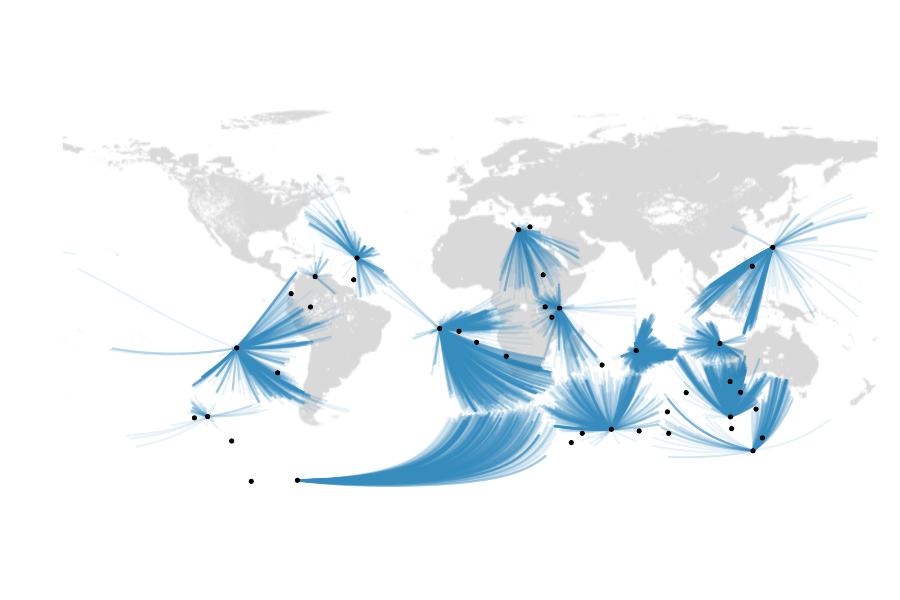} \vspace{-2em} \\
      
          \textbf{RA-OT} & \textbf{OA-OT}   \vspace{-2em} \\
         \includegraphics[width=0.45\linewidth]{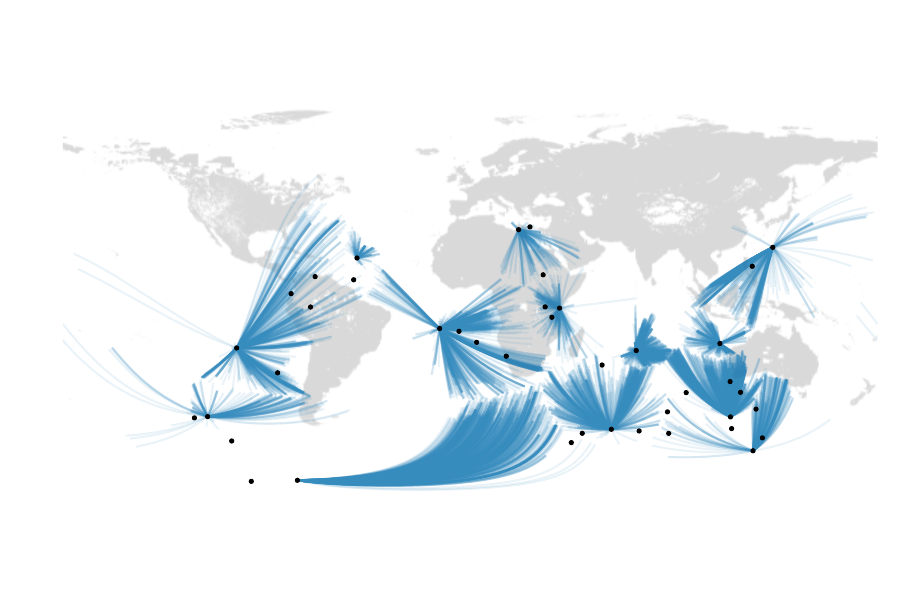} &\includegraphics[width=0.45\linewidth]{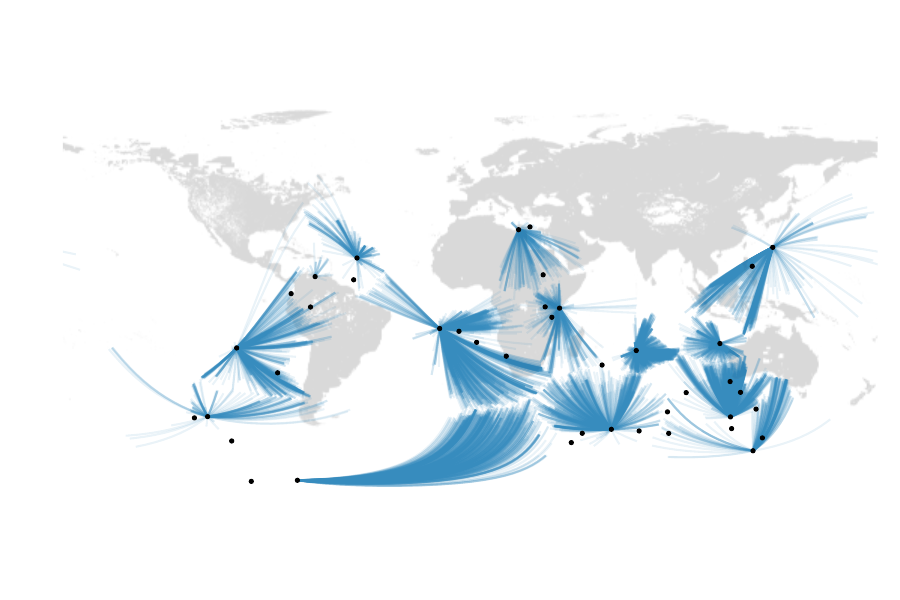} 
       
    \end{tabular}
  \vspace{-3em} 
    \caption{ \footnotesize {Prediction on a held-out spherical supply--demand instance
    ($M{=}50$, $L{=}100$, $\varepsilon{=}0.5$). (Top) Sinkhorn ground truth and baseline Meta-OT. (Bottom) our methods: RA-OT and OA-OT 
    Black dots denote
    active supply locations; blue arcs show great-circle geodesics from each demand
    location to its assigned supply location under the predicted plan.}}
    \label{fig:world_transport}
\end{figure}

To evaluate the proposed methods in a non-Euclidean domain, we construct a supply-demand transportation problem on the unit sphere $\mathbb{S}^2{:=}\{x{\in}\mathbb{R}^3:\|x\|{=}1\}$. Supply capacities are assumed to be uniform across $n{=}100$ locations sampled from Earth's landmass, while demand weights are drawn from Earth's population density across $m{=}10{,}000$ locations based on the CC-licensed dataset \citep{Doxsey-Whitfield03072015}. The ground metric is defined by the geodesic cost $c(x,y){=}\arccos(\langle x,y\rangle)$ with an entropic regularization parameter of $\varepsilon{=}0.5$. To properly accommodate the spherical geometry within our framework, we utilize stereographic projections~\citep{tran2024stereographic} to define the slicing family for both RA-OT and OA-OT. As reported in Table~\ref{tab:world_vary_m}, both methods generalize well with few training samples and converge quickly in practice. This task highlights the importance of geometry: while Min-STP and min-SWGG fail to respect the spherical structure, our approach remains close to the Sinkhorn solution. The transport patterns shown in Figure~\ref{fig:world_transport} (Appendix \ref{ab:add_vis}) clearly illustrate this behavior.

\begin{table}[!t]
\centering
\caption{  \footnotesize {RMSE, training and inference time on the \textbf{spherical supply-demand transport} task across varying number of training data $M$ ($\varepsilon{=}0.5$, $n{=}100$, $m{=}10{,}000$, $N{=}300$, $L{=}100$). The results are reported as the mean $\pm$ standard deviation. The label ``no train'' indicates that no prior training is required. \textbf{Bold} values denote the best in each block.}}
\resizebox{\linewidth}{!}{%
\begin{tabular}{@{} c l c c c @{\hspace{1.5em}} c l c c c @{}}
\toprule
$M$ & Method & RMSE\,($\times 10^{-7}$, $\downarrow$) & Train\,(s) & Infer\,(ms,$\downarrow$) & 
$M$ & Method & RMSE\,($\times 10^{-7}$, $\downarrow$) & Train\,(s) & Infer\,(ms,$\downarrow$) \\
\cmidrule(r){1-5} \cmidrule(l){6-10}

\multirow{5}{*}{10} 
 & Meta-OT \citep{amos2023meta} & $6.83 \pm 2.49$ & $52.08$ & $15.94 \pm 2.14$ & 
\multirow{5}{*}{50} 
 & Meta-OT \citep{amos2023meta} & $4.42 \pm 1.55$ & $52.07$ & $16.09 \pm 2.17$ \\
 & Min-STP \citep{liu2025efficient}                     & $115.07 \pm 0.54$ & $46.06$ & $\mathbf{7.51 \pm 7.23}$ & 
 & Min-STP \citep{liu2025efficient}                     & $115.10 \pm 0.54$ & $46.38$ & $\mathbf{7.88 \pm 1.22}$ \\
 & min-SWGG \citep{mahey2023fast}                     & $112.15 \pm 0.99$ & no train & $34.62 \pm 2.70$ & 
 & min-SWGG \citep{mahey2023fast}                     & $112.15 \pm 0.99$ & no train & $33.25 \pm 1.71$ \\
 & \textbf{RA-OT}  & $7.92 \pm 2.77$ & $\mathbf{1.05}$ & $41.16 \pm 3.11$ & 
 & \textbf{RA-OT}  & $7.82 \pm 1.89$ & $\mathbf{2.53}$ & $41.96 \pm 6.29$ \\
 & \textbf{OA-OT}  & $\mathbf{4.80 \pm 1.96}$ & $18.60$ & $39.92 \pm 3.12$ & 
 & \textbf{OA-OT}  & $\mathbf{3.93 \pm 1.92}$ & $19.37$ & $41.03 \pm 2.76$ \\
\cmidrule(r){1-5} \cmidrule(l){6-10}

\multirow{5}{*}{20} 
 & Meta-OT \citep{amos2023meta} & $6.03 \pm 2.20$ & $51.99$ & $15.57 \pm 1.22$ & 
\multirow{5}{*}{200} 
 & Meta-OT \citep{amos2023meta} & $\mathbf{3.62 \pm 1.14}$ & $52.12$ & $15.17 \pm 2.54$ \\
 & Min-STP \citep{liu2025efficient}                     & $115.06 \pm 0.54$ & $46.39$ & $\mathbf{7.95 \pm 1.41}$ & 
 & Min-STP \citep{liu2025efficient}                     & $115.05 \pm 0.54$ & $46.41$ & $\mathbf{7.93 \pm 1.22}$ \\
 & min-SWGG \citep{mahey2023fast}                     & $112.15 \pm 0.99$ & no train & $33.85 \pm 2.49$ & 
 & min-SWGG \citep{mahey2023fast}                     & $112.15 \pm 0.99$ & no train & $33.59 \pm 1.99$ \\
 & \textbf{RA-OT}  & $7.82 \pm 2.25$ & $\mathbf{1.31}$ & $40.70 \pm 3.25$ & 
 & \textbf{RA-OT}  & $7.55 \pm 1.88$ & $\mathbf{8.74}$ & $39.40 \pm 1.43$ \\
 & \textbf{OA-OT}  & $\mathbf{4.24 \pm 1.98}$ & $18.75$ & $39.65 \pm 3.26$ & 
 & \textbf{OA-OT}  & $3.66 \pm 1.56$ & $21.32$ & $40.03 \pm 3.83$ \\
\bottomrule
\end{tabular}%
}
\label{tab:world_vary_m}
  \vspace{-1em} 
\end{table}

\subsection{Color Transfer}

\begin{figure}[!t]
    \centering
    \setlength{\tabcolsep}{0pt}
    \renewcommand{\arraystretch}{0}
    \newcommand{\imw}{0.2\textwidth}
    \scalebox{1}{
    \begin{tabular}{@{}ccccc@{}}
        \small Source & \small Target & \small Sinkhorn \textcolor{gray}{(GT)} & \small RA-OT & \small OA-OT \\[2pt]

        \includegraphics[width=\imw, height=\imw, keepaspectratio=false]{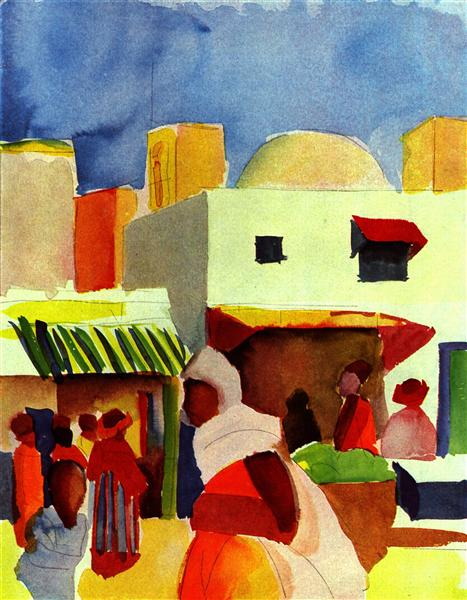} &
        \includegraphics[width=\imw, height=\imw, keepaspectratio=false]{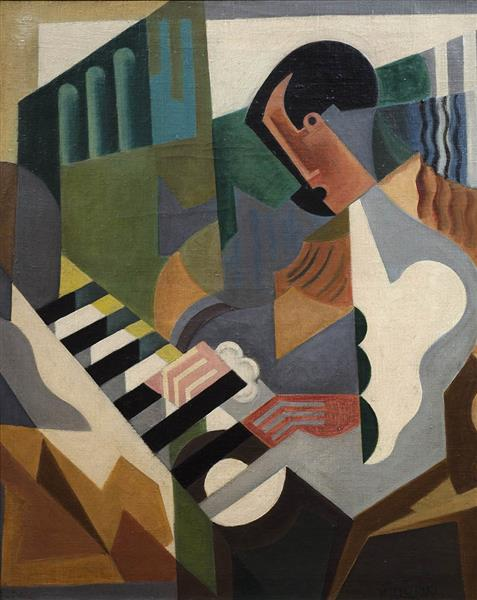} &
        \includegraphics[width=\imw, height=\imw, keepaspectratio=false]{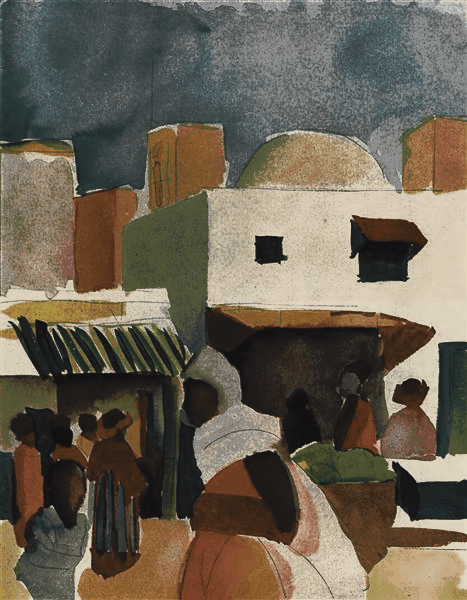} &
        \includegraphics[width=\imw, height=\imw, keepaspectratio=false]{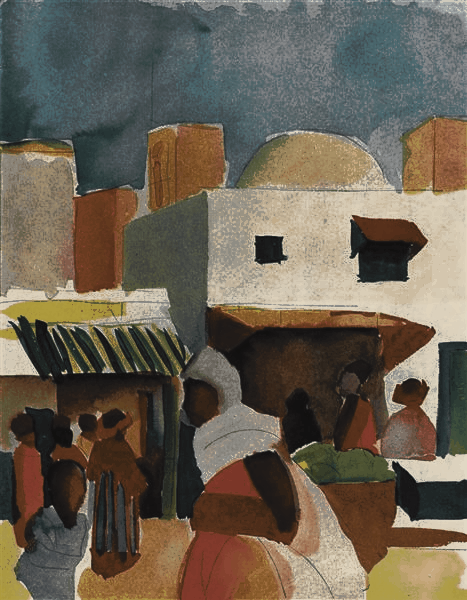} &
        \includegraphics[width=\imw, height=\imw, keepaspectratio=false]{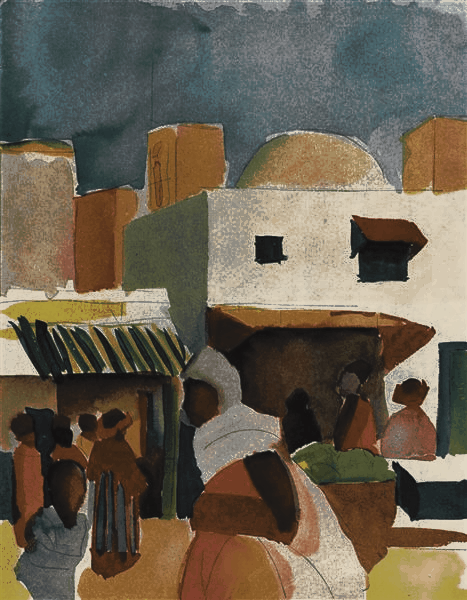}\\[-0.5pt]

        \includegraphics[width=\imw, height=\imw, keepaspectratio=false]{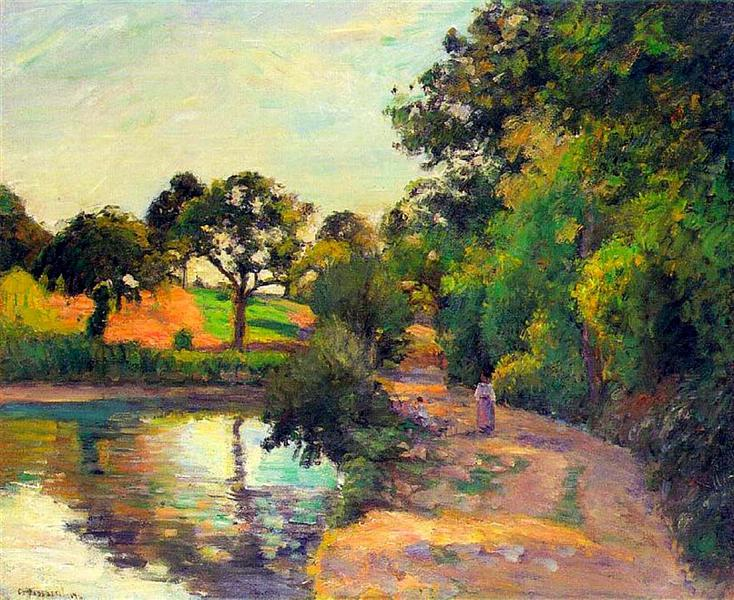} &
        \includegraphics[width=\imw, height=\imw, keepaspectratio=false]{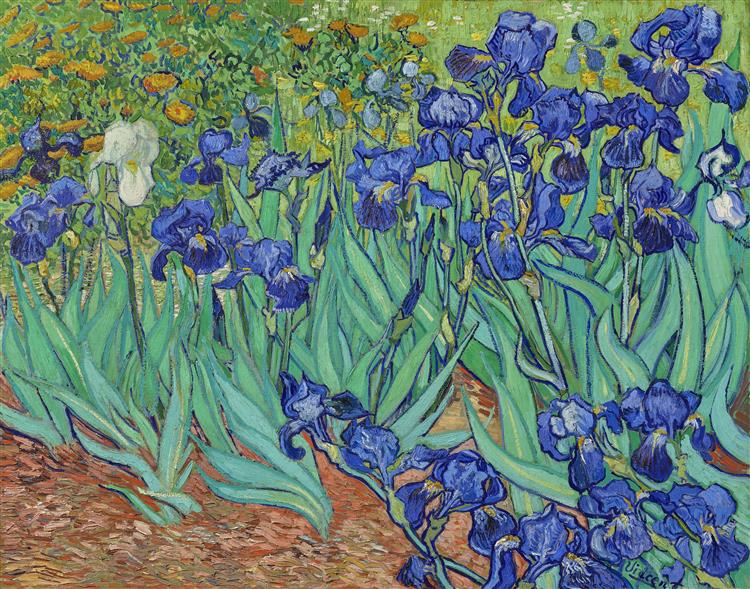} &
        \includegraphics[width=\imw, height=\imw, keepaspectratio=false]{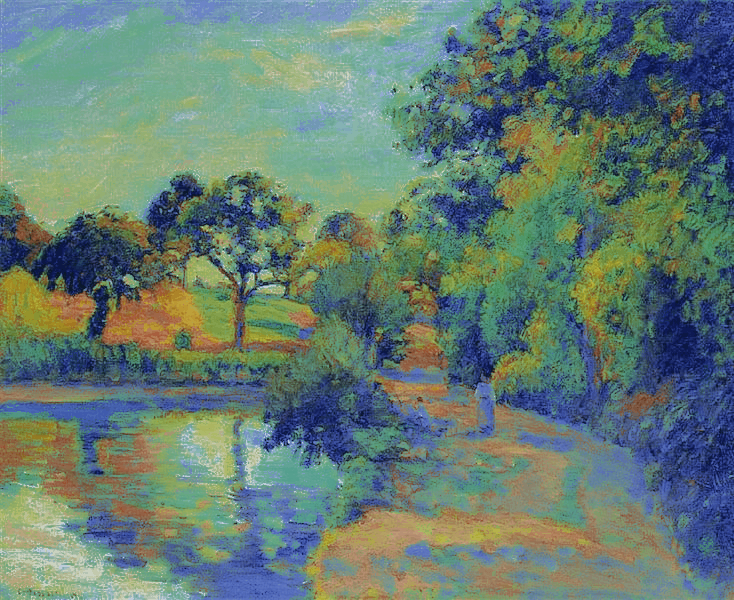} &
        \includegraphics[width=\imw, height=\imw, keepaspectratio=false]{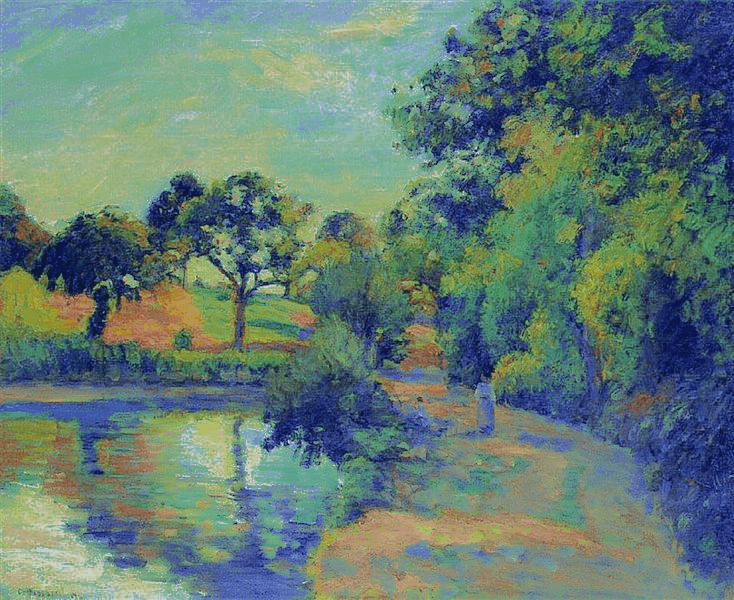} &
        \includegraphics[width=\imw, height=\imw, keepaspectratio=false]{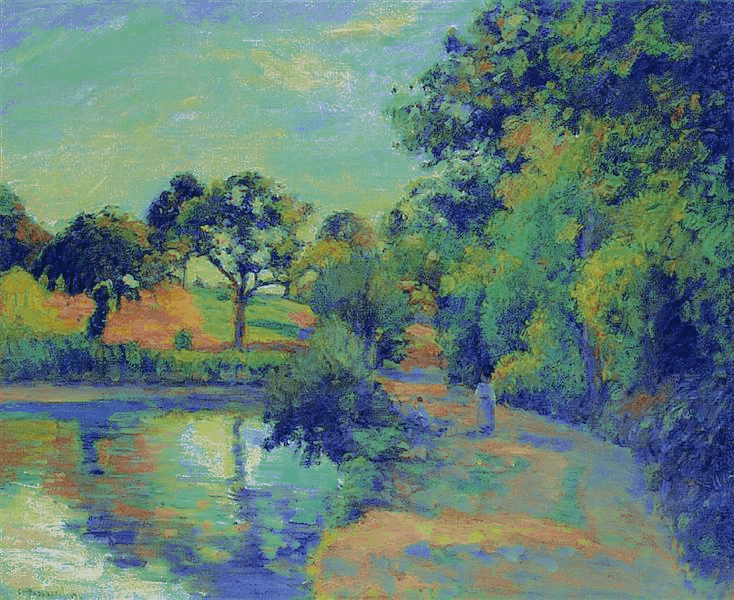}\\

    \end{tabular}
}
    \caption{\footnotesize{Color transfer results on held-out test pairs ($M{=}50$, $L{=}100$, $\varepsilon{=}0.005$, $K{=}500$ clusters). Each row corresponds to one source--target pair. Columns show: source $\alpha$, target $\beta$, Sinkhorn ground truth, RA-OT, and OA-OT.}}
    \label{fig:color_transfer}
    \vspace{-0.5em}
\end{figure}

Color transfer is formulated as an OT problem by representing images as weighted point clouds in the normalized RGB space $[0,1]^3$. Following the experimental setup of Meta-OT, we collect approximately 200 public domain images from WikiArt to construct the dataset. To retain a discrete formulation, images are quantized into $K{=}500$ distinct color clusters using mini-batch $k$-means, where centroids act as atoms and normalized cluster occupancies as weights. The ground metric is the squared Euclidean distance with $\varepsilon{=}0.005$. Transferred images are generated by updating source pixels via the target centroids weighted by the row-normalized predicted transport plan. As shown in Table~\ref{tab:color_vary_m}, RA-OT and OA-OT achieve strong performance across all data regimes, with clear advantages when training data is limited. They also require less training time compared to Meta-OT. The gap with Min-STP and min-SWGG suggests that single projections are insufficient for modeling color distributions. As seen in Figure~\ref{fig:color_transfer}, this leads to more faithful color reconstruction.

\begin{table}[!t]
\centering
\caption{\footnotesize{RMSE, training and inference time on the \textbf{color transfer} task across varying number of training data $M$ ($\varepsilon{=}0.005$, $K{=}500$ clusters, $N{=}300$ test pairs, $L{=}100$). The results are reported as the mean $\pm$ standard deviation. The label ``no train'' indicates that no prior training is required. \textbf{Bold} values denote the best performance in each block.}}
\resizebox{\linewidth}{!}{%
\begin{tabular}{@{} c l c c c @{\hspace{1.5em}} c l c c c @{}}
\toprule
$M$ & Method & RMSE\,($\times 10^{-6}$, $\downarrow$) & Train\,(s) & Infer\,(ms,$\downarrow$) & 
$M$ & Method & RMSE\,($\times 10^{-6}$, $\downarrow$) & Train\,(s) & Infer\,(ms,$\downarrow$) \\
\cmidrule(r){1-5} \cmidrule(l){6-10}

\multirow{5}{*}{10} 
 & Meta-OT \citep{amos2023meta} & $36.83 \pm 10.97$ & $33.32$ & $16.83 \pm 1.33$ & 
\multirow{5}{*}{50} 
 & Meta-OT \citep{amos2023meta} & $33.16 \pm 10.45$ & $32.91$ & $15.16 \pm 0.64$ \\
 & Min-STP \citep{liu2025efficient}                     & $80.93 \pm 3.24$ & $49.51$ & $\mathbf{4.02 \pm 0.27}$ & 
 & Min-STP \citep{liu2025efficient}                     & $81.04 \pm 3.42$ & $50.00$ & $\mathbf{4.10 \pm 0.20}$ \\
 & min-SWGG \citep{mahey2023fast}                     & $77.69 \pm 3.19$ & no train & $4.25 \pm 0.21$ & 
 & min-SWGG \citep{mahey2023fast}                     & $77.69 \pm 3.19$ & no train & $4.34 \pm 0.17$ \\
 & \textbf{RA-OT}  & $10.73 \pm 6.35$ & $\mathbf{1.95}$ & $18.41 \pm 1.21$ & 
 & \textbf{RA-OT}  & $9.99 \pm 5.61$ & $\mathbf{7.40}$ & $17.40 \pm 0.83$ \\
 & \textbf{OA-OT}  & $\mathbf{9.78 \pm 5.78}$ & $17.52$ & $18.77 \pm 1.31$ & 
 & \textbf{OA-OT}  & $\mathbf{9.00 \pm 5.02}$ & $18.13$ & $17.76 \pm 0.79$ \\
\cmidrule(r){1-5} \cmidrule(l){6-10}

\multirow{5}{*}{20} 
 & Meta-OT \citep{amos2023meta} & $34.73 \pm 10.94$ & $34.63$ & $15.03 \pm 0.70$ & 
\multirow{5}{*}{200} 
 & Meta-OT \citep{amos2023meta} & $33.23 \pm 10.89$ & $33.11$ & $15.44 \pm 1.20$ \\
 & Min-STP \citep{liu2025efficient}                     & $80.97 \pm 3.33$ & $49.92$ & $\mathbf{4.07 \pm 0.23}$ & 
 & Min-STP \citep{liu2025efficient}                     & $81.10 \pm 3.38$ & $49.68$ & $\mathbf{4.08 \pm 0.22}$ \\
 & min-SWGG \citep{mahey2023fast}                     & $77.69 \pm 3.19$ & no train & $4.30 \pm 0.17$ & 
 & min-SWGG \citep{mahey2023fast}                     & $77.69 \pm 3.19$ & no train & $4.27 \pm 0.19$ \\
 & \textbf{RA-OT}  & $9.74 \pm 5.68$ & $\mathbf{3.11}$ & $18.02 \pm 1.34$ & 
 & \textbf{RA-OT}  & $10.21 \pm 5.53$ & $26.57$ & $18.08 \pm 1.40$ \\
 & \textbf{OA-OT}  & $\mathbf{9.31 \pm 5.24}$ & $17.92$ & $18.45 \pm 1.20$ & 
 & \textbf{OA-OT}  & $\mathbf{9.00 \pm 5.05}$ & $\mathbf{20.37}$ & $18.25 \pm 0.71$ \\
\bottomrule
\end{tabular}%
}
\label{tab:color_vary_m}
\vspace{-0.5em}
\end{table}

\begin{table}[!t]
\caption{\footnotesize{Quantitative results on 2D Conditional Flow Matching tasks. We report both training and pretraining time.}}
\label{tab:fm_results}
\centering
\resizebox{\textwidth}{!}{
\begin{tabular}{l cccc cccc cccc}
\toprule
& \multicolumn{4}{c}{\textbf{Gaussian $\rightarrow$ 8gaussians}} & \multicolumn{4}{c}{\textbf{Gaussian $\rightarrow$ moons}} & \multicolumn{4}{c}{\textbf{Gaussian $\rightarrow$ scurve}} \\
\cmidrule(lr){2-5} \cmidrule(lr){6-9} \cmidrule(lr){10-13}
\textbf{Method} & $\mathbf{W_2^2} \downarrow$ & \textbf{NPE} $\downarrow$ & \textbf{Train (s)} & \textbf{Pretrain} & $\mathbf{W_2^2} \downarrow$ & \textbf{NPE} $\downarrow$ & \textbf{Train (s)} & \textbf{Pretrain} & $\mathbf{W_2^2} \downarrow$ & \textbf{NPE} $\downarrow$ & \textbf{Train (s)} & \textbf{Pretrain} \\
\midrule
I-CFM & 0.2800 & 0.2590 & \textbf{168.7} & 0.0s & 0.3901 & 0.5380 & \textbf{45.6} & 0.0s & 0.1113 & 1.9230 & \textbf{44.5} & 0.0s \\
OT-CFM & \textbf{0.1956} & \textbf{0.0108} & 1054.7 & 0.0s & \textbf{0.2668} & \textbf{0.0142} & 1513.1 & 0.0s & \textbf{0.0991} & \textbf{0.0243} & 992.0 & 0.0s \\
RA-OT & 0.5072 & 0.0144 & 411.8 & \textbf{2.6s} & 0.5569 & \textbf{0.0619} & 331.7 & \textbf{1.0s} & 0.4675 & 0.0400 & 299.3 & \textbf{6.3s} \\
OA-OT & 0.5041 & 0.0142 & 413.8 & 14.0s & 0.5459 & 0.0631 & 319.7 & 14.6s & 0.4672 & 0.0425 & 294.0 & 14.4s \\
\bottomrule
\end{tabular}
}
\vspace{-1.5em}
\end{table}
\subsection{Mini-batch Optimal Transport Conditional Flow Matching}

\begin{figure}[!htbp]
    \centering
    \begin{tabular}{cccc}
        \textbf{I-CFM} & \textbf{OT-CFM} & \textbf{RA-OT} & \textbf{OA-OT} \\
        \includegraphics[width=0.23\textwidth]{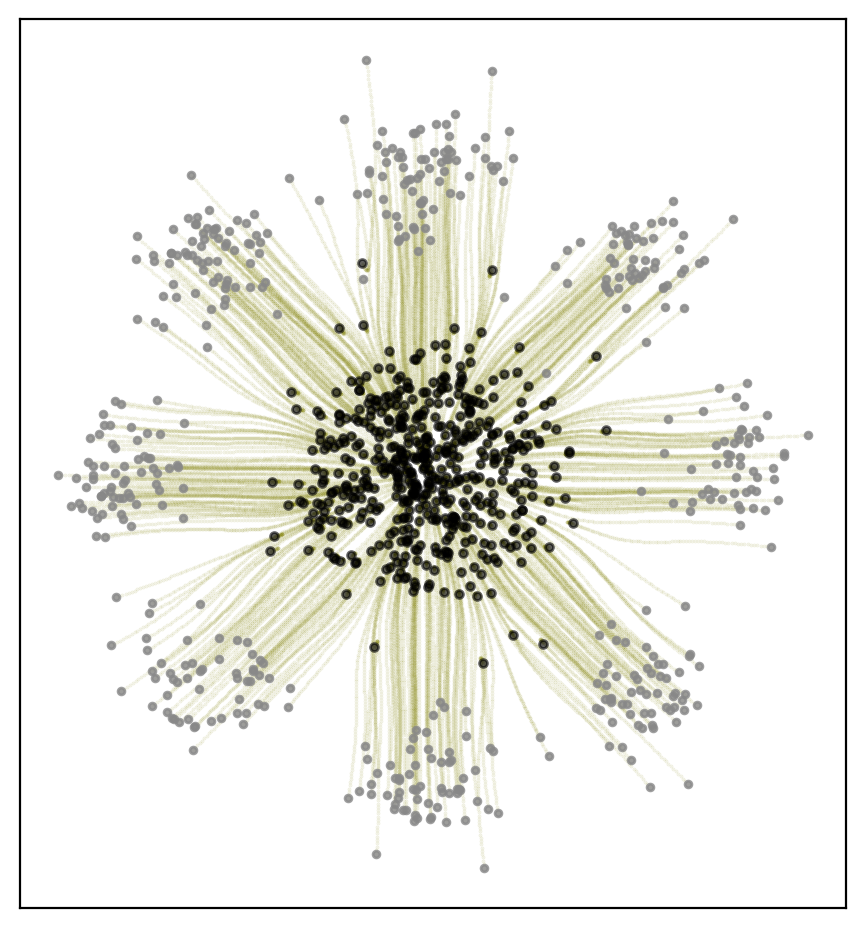} & 
        \includegraphics[width=0.23\textwidth]{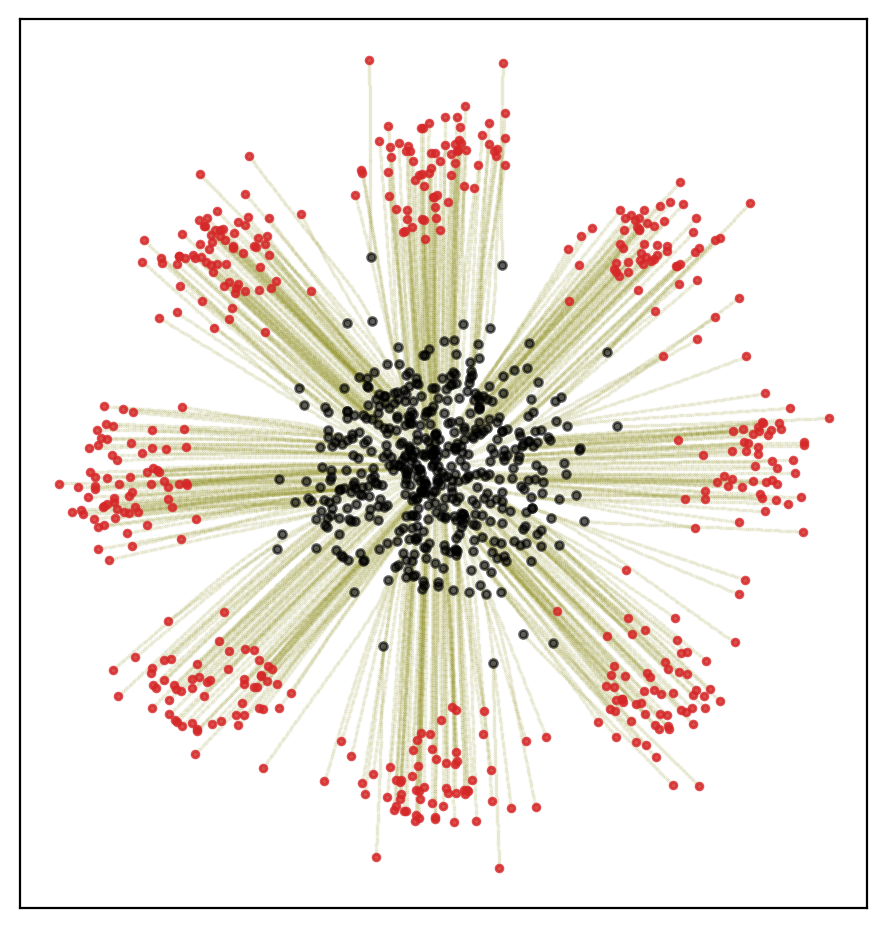} &
        \includegraphics[width=0.23\textwidth]{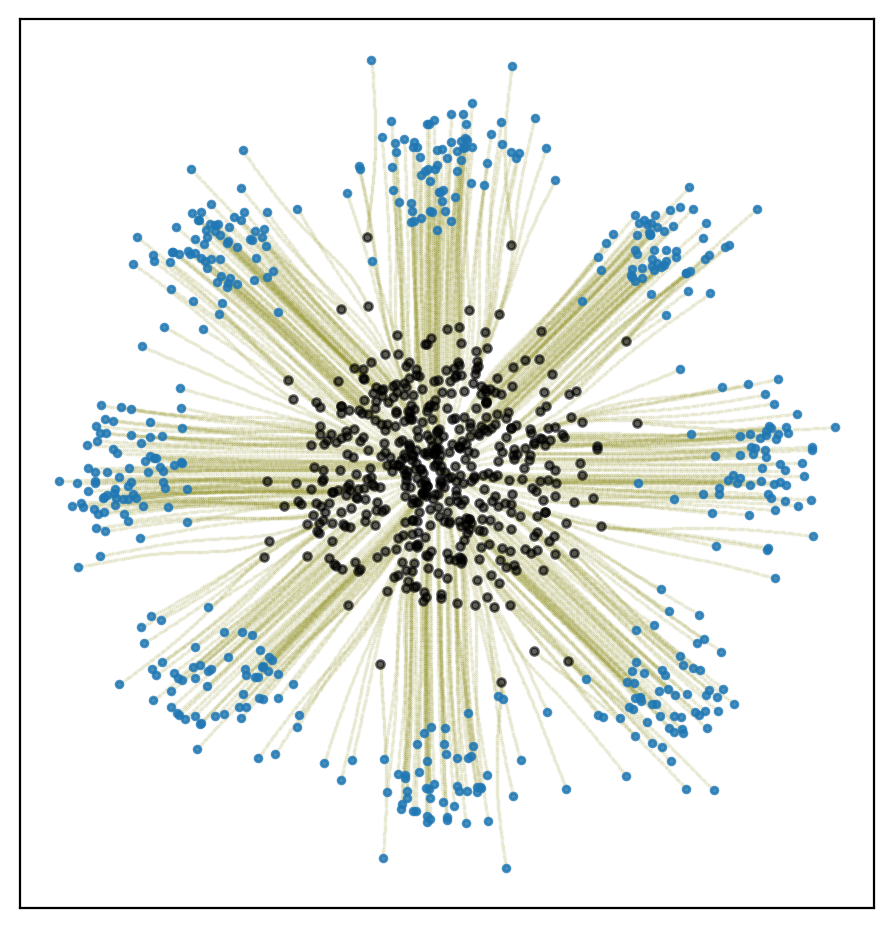} &
        \includegraphics[width=0.23\textwidth]{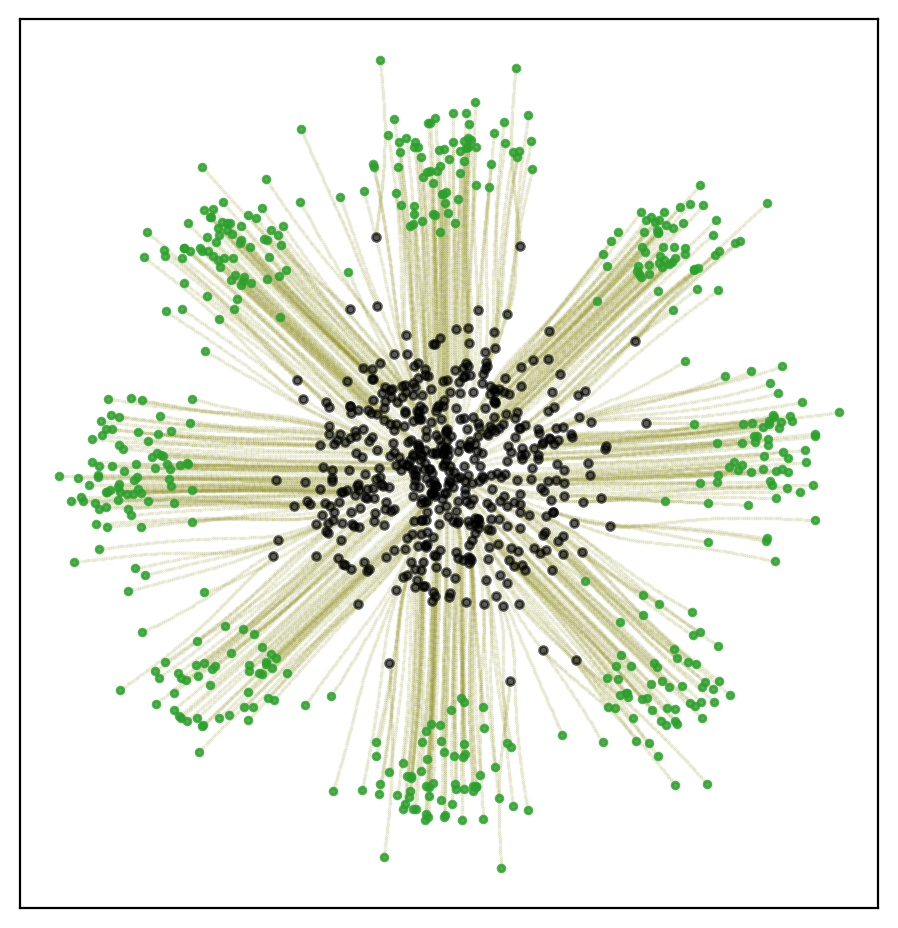} 
    \end{tabular}
    \vspace{-1em} 
    \caption{\footnotesize {Generated flow trajectories moving from a 2D Gaussian prior to the \textbf{8-Gaussians} target. (Left to Right) Baseline I-CFM, OT-CFM, and our proposed amortized methods. Black dots signify source samples. The olive tracks map the predicted integration paths. Our solutions successfully reproduce the straight, uncrossed optimal transport paths of OT-CFM, yielding significantly lower collision energy than standard I-CFM.}}
    \label{fig:traj_8gauss}
\end{figure}

\begin{figure}[!htbp]
    \centering
    \begin{tabular}{cccc}
        \textbf{I-CFM} & \textbf{OT-CFM} & \textbf{RA-OT} & \textbf{OA-OT} \\
        \includegraphics[width=0.23\textwidth]{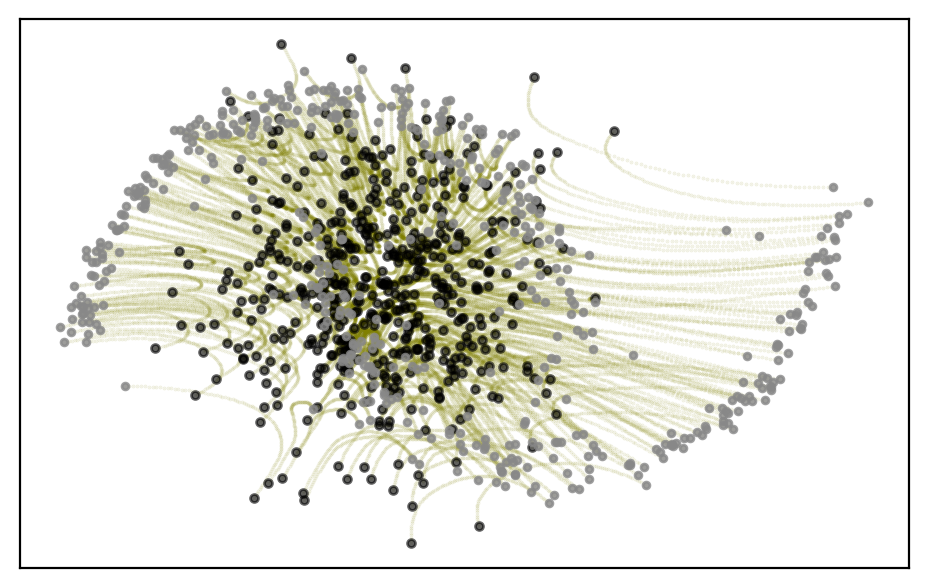} & 
        \includegraphics[width=0.23\textwidth]{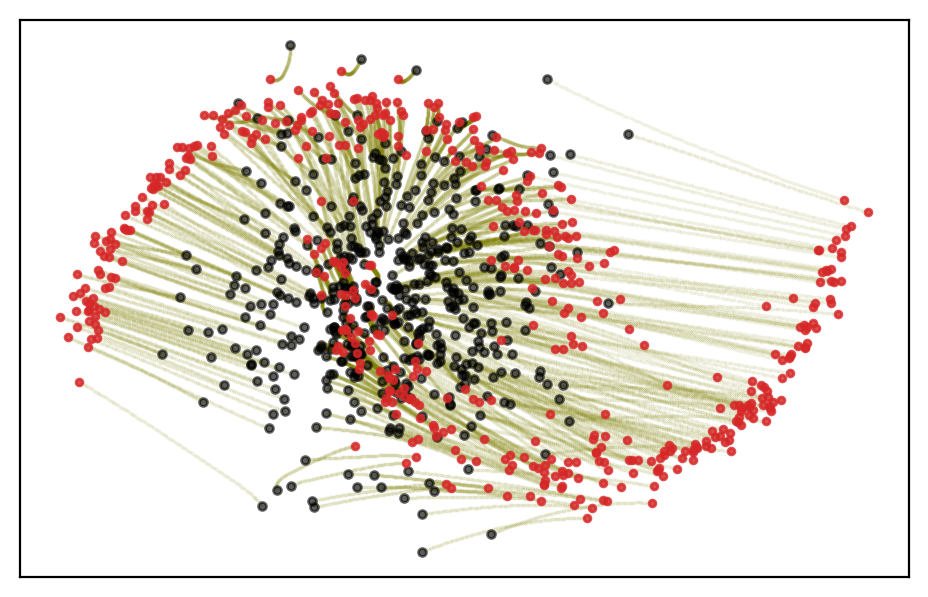} &
        \includegraphics[width=0.23\textwidth]{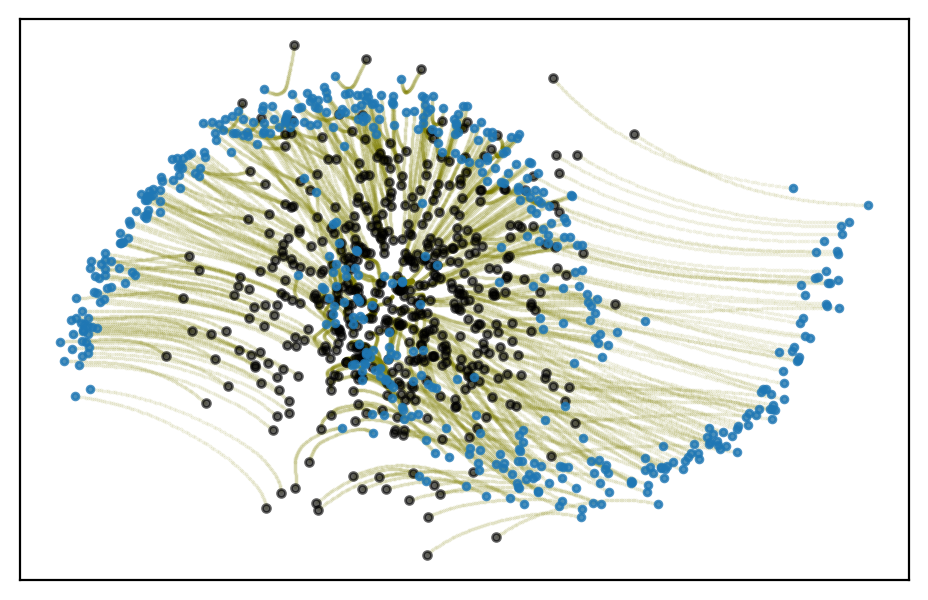} &
        \includegraphics[width=0.23\textwidth]{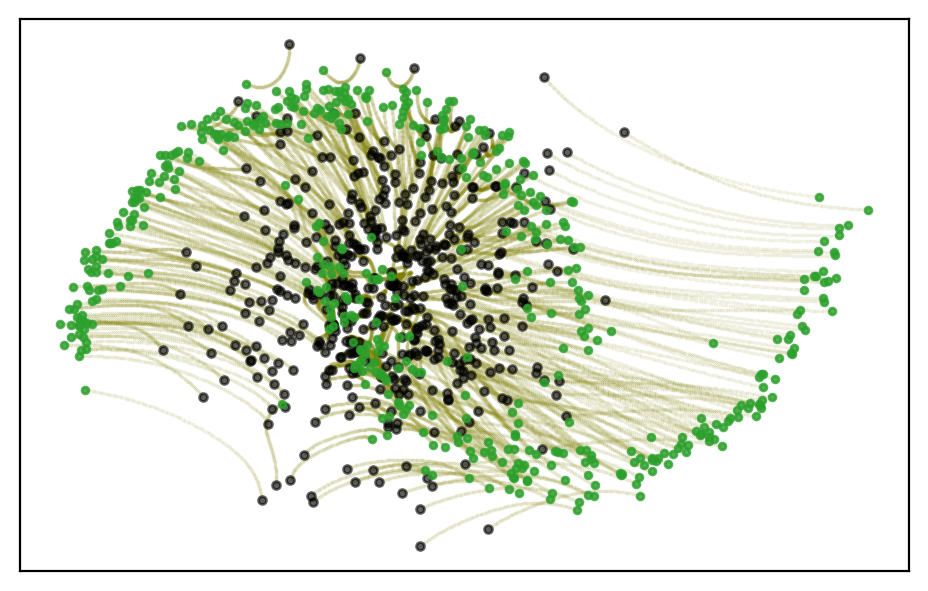} 
    \end{tabular}
    \vspace{-1em} 
    \caption{\footnotesize {Flow trajectories transforming a 2D Gaussian prior to the \textbf{Moons} target distribution. (Left to Right) Baselines I-CFM and OT-CFM, followed by our proposed RA-OT and OA-OT methods. Both amortized variants effectively map unentangled optimal integration flows (olive paths) across the dual clusters, matching exact OT capabilities.}}
    \label{fig:traj_moons}
        \vspace{-0.5em} 
\end{figure}

\begin{figure}[!htbp]
    \centering
    \begin{tabular}{cccc}
        \textbf{I-CFM} & \textbf{OT-CFM} & \textbf{RA-OT} & \textbf{OA-OT} \\
        \includegraphics[width=0.23\textwidth]{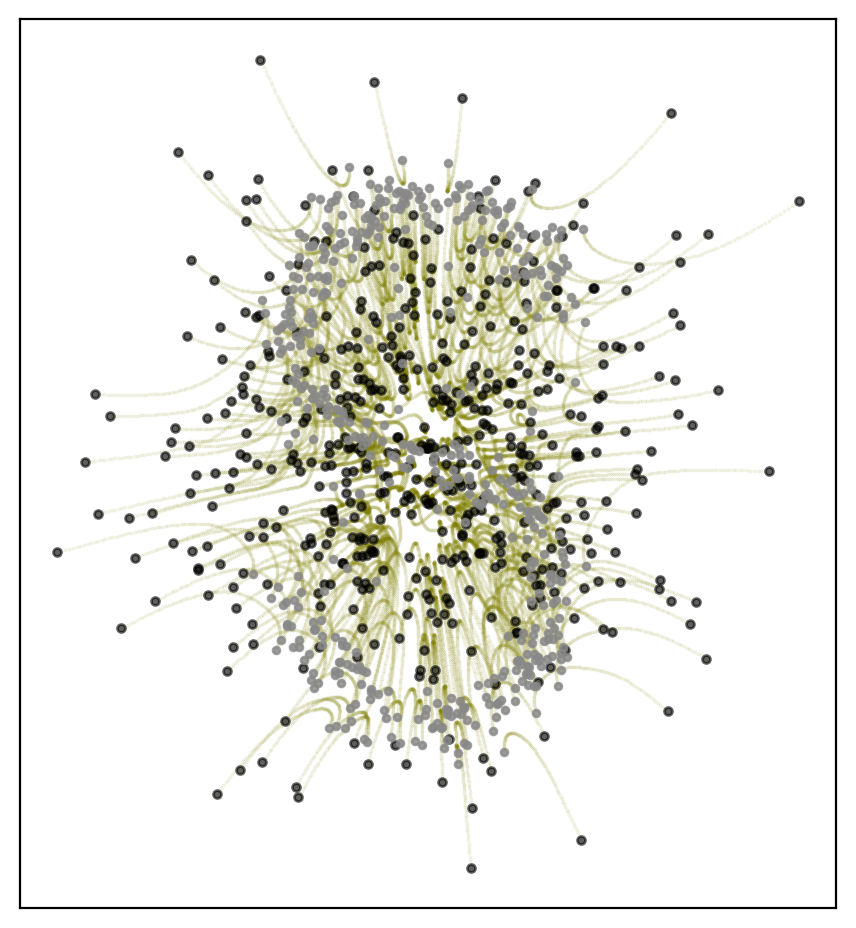} & 
        \includegraphics[width=0.23\textwidth]{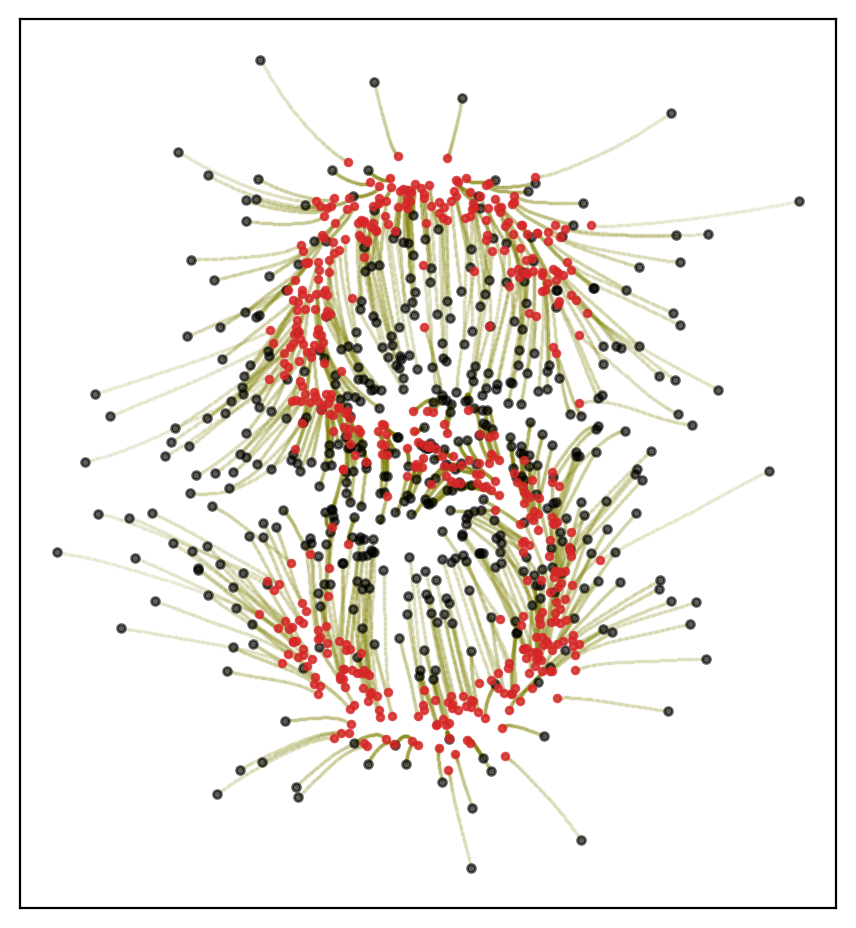} &
        \includegraphics[width=0.23\textwidth]{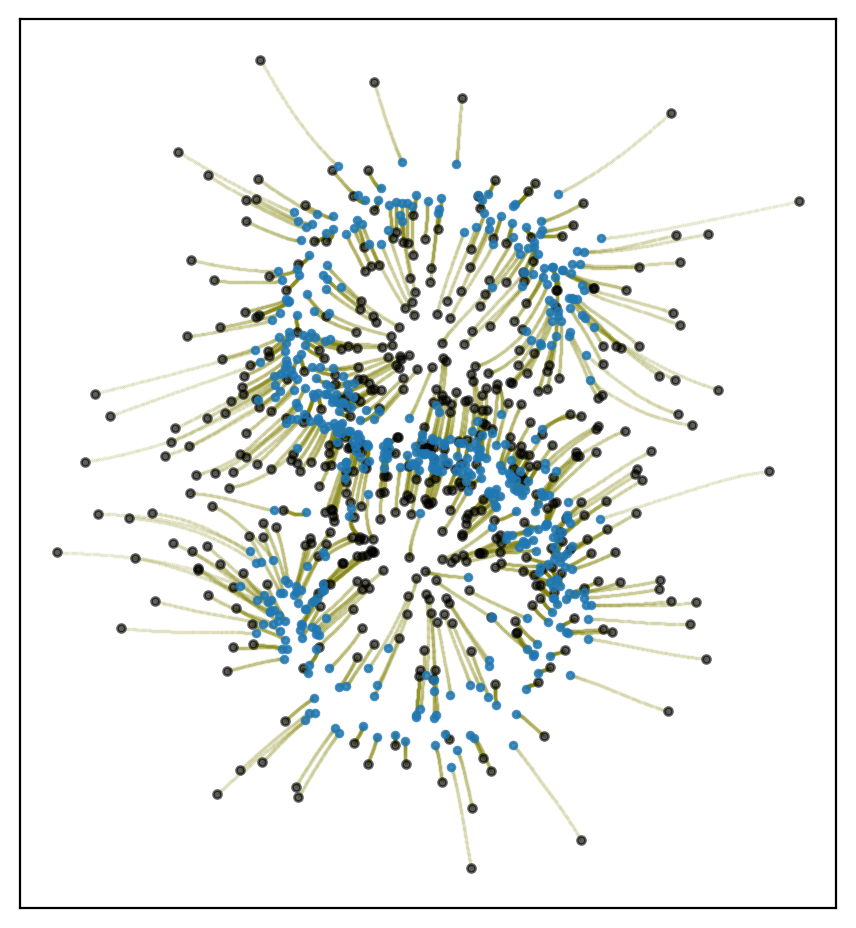} &
        \includegraphics[width=0.23\textwidth]{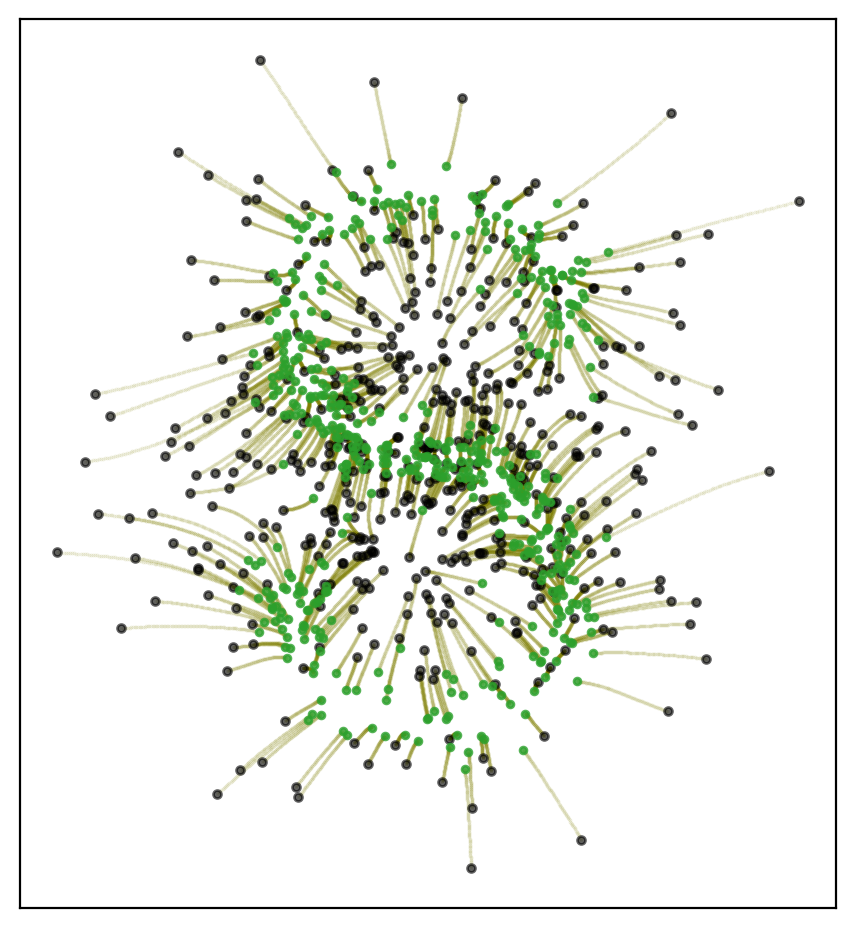} 
    \end{tabular}
    \vspace{-1em} 
    \caption{\footnotesize {Visualized flow trajectories mapping a 2D Gaussian prior into the \textbf{S-Curve} distribution. (Left to Right) I-CFM, OT-CFM, RA-OT, and OA-OT. Our amortized approaches robustly straighten the generative pathways, practically bypassing the severe integration turbulence traditionally observed in uncoupled I-CFM trajectories.}}
    \label{fig:traj_scurve}
\end{figure}

OT-CFM \citep{tong2024improving} leverages mini-batch OT to straighten flow trajectories in flow matching, thereby reducing inference cost. However, computing exact OT remains expensive and constitutes a major training bottleneck. To address this, we introduce a more efficient alternative by integrating RA-OT and OA-OT into the conditional flow matching framework, significantly accelerating training while preserving near-straight trajectories. Following \citep{tong2024improving}, we map $\mathcal{N}(0,I)$ to 2D targets (8gaussians, moons, scurve) using a network with three 64-unit SELU layers and a linear head, trained for 20000 steps (batch size 512, $\sigma=0.1$) with AdamW (learning rate $10^{-3}$, weight decay $10^{-5}$). We compare RA-OT and OA-OT against I-CFM and OT-CFM, where amortized methods use $M=50$ pretraining batches and $L=100$ projections. All methods employ fixed-step RK4 (101 steps); performance is evaluated using Sinkhorn $W_2^2$ at $t=1$ and path straightness via NPE. As shown in Table \ref{tab:fm_results}, our methods achieve $2.5\times$–$4.5\times$ training speedups over OT-CFM while maintaining comparable trajectory quality, with NPE close to OT-CFM and up to 1–2 orders of magnitude lower than I-CFM. Although slight degradation in Sinkhorn $W_2^2$ is observed in more complex settings, the overall trade-off highlights the scalability of our approach. The visualizations in Figures~\ref{fig:traj_8gauss}, \ref{fig:traj_moons}, and \ref{fig:traj_scurve} further illustrate the learned trajectories across the three 2D targets. Both RA-OT and OA-OT produce straighter and less entangled trajectories than I-CFM while closely resembling those of OT-CFM, consistent with the low NPE values in Table~\ref{tab:fm_results}. To further examine scalability, we additionally evaluate our methods on high-dimensional CIFAR-10 image generation. The results in Appendix~\ref{ab: high_dim_cfm} show that RA-OT and OA-OT reduce the computational overhead of OT-based flow matching while maintaining competitive generation quality.

\section{Conclusion}
\label{sec:conclusion}

The paper presents two novel amortized OT methods, RA-OT and OA-OT, which leverage sliced OT to efficiently predict OT plans across multiple measure pairs. RA-OT uses regression to map sliced OT potentials to original OT potentials, while OA-OT optimizes the Kantorovich dual within the span of sliced OT potentials. Both approaches enable rapid and accurate solutions for repeated OT problems, are parsimonious, and do not depend on the structure or size of the measures. Experiments on MNIST digit transport, color transfer, spherical supply–demand problems, and conditional flow matching demonstrate the practical effectiveness and generality of the methods.

\clearpage

\section*{Acknowledgement}
Portions of this research were conducted with the advanced computing resources provided by Texas A\&M High Performance Research Computing.
\bibliographystyle{abbrv}
\bibliography{bib}

@inproceedings{sejourne2022faster,
  title={Faster unbalanced optimal transport: Translation invariant sinkhorn and 1-d frank-wolfe},
  author={S{\'e}journ{\'e}, Thibault and Vialard, Fran{\c{c}}ois-Xavier and Peyr{\'e}, Gabriel},
  booktitle={International Conference on Artificial Intelligence and Statistics},
  pages={4995--5021},
  year={2022},
  organization={PMLR}
}

@article{nguyen2026summarizing,
  title={Summarizing Nonparametric {B}ayesian Mixture Posteriors--Sliced Optimal Transport Metrics for {G}aussian Mixtures},
  author={Nguyen, Khai and Mueller, Peter},
  journal={Journal of Computational and Graphical Statistics},
  pages={1--22},
  year={2026},
  publisher={Taylor \& Francis}
}

@inproceedings{rabin2012wasserstein,
  title={Wasserstein barycenter and its application to texture mixing},
  author={Rabin, Julien and Peyr{\'e}, Gabriel and Delon, Julie and Bernot, Marc},
  booktitle={Scale Space and Variational Methods in Computer Vision: Third International Conference, SSVM 2011, Ein-Gedi, Israel, May 29--June 2, 2011, Revised Selected Papers 3},
  pages={435--446},
  year={2012},
  organization={Springer, One New York Plaza, Suite 4600, New York, NY 10004-1562}
}

@inproceedings{fatras2020learning,
  title={Learning with minibatch {W}asserstein: asymptotic and gradient properties},
  author={Fatras, Kilian and Zine, Younes and Flamary, R{\'e}mi and Gribonval, R{\'e}mi and Courty, Nicolas},
  booktitle={AISTATS 2020-23nd International Conference on Artificial Intelligence and Statistics},
  volume={108},
  pages={1--20},
  year={2020}
}

@InProceedings{fatras2021unbalanced,
  title = 	 {Unbalanced minibatch Optimal Transport; applications to Domain Adaptation},
  author =       {Fatras, Kilian and Sejourne, Thibault and Flamary, R{\'e}mi and Courty, Nicolas},
  booktitle = 	 {Proceedings of the 38th International Conference on Machine Learning},
  pages = 	 {3186--3197},
  year = 	 {2021},
  editor = 	 {Meila, Marina and Zhang, Tong},
  volume = 	 {139},
  series = 	 {Proceedings of Machine Learning Research},
  month = 	 {18--24 Jul},
  publisher =    {PMLR},
  url = 	 {http://proceedings.mlr.press/v139/fatras21a.html}
}

@book{villani2009optimal,
  title={Optimal transport: old and new},
  author={Villani, C{\'e}dric},
  volume={338},
  year={2009},
  publisher={Springer, One New York Plaza, Suite 4600, New York, NY 10004-1562}
}

@inproceedings{genevay2018learning,
  title={Learning generative models with {S}inkhorn divergences},
  author={Genevay, Aude and Peyr{\'e}, Gabriel and Cuturi, Marco},
  booktitle={International Conference on Artificial Intelligence and Statistics},
  pages={1608--1617},
  year={2018},
  organization={PMLR}
}

@inproceedings{damodaran2018deepjdot,
  title={Deepjdot: Deep joint distribution optimal transport for unsupervised domain adaptation},
  author={Damodaran, Bharath Bhushan and Kellenberger, Benjamin and Flamary, R{\'e}mi and Tuia, Devis and Courty, Nicolas},
  booktitle={Proceedings of the European Conference on Computer Vision (ECCV)},
  pages={447--463},
  year={2018}
}

@article{sommerfeld2019optimal,
  title={Optimal Transport: {F}ast Probabilistic Approximation with Exact Solvers.},
  author={Sommerfeld, Max and Schrieber, J{\"o}rn and Zemel, Yoav and Munk, Axel},
  journal={Journal of Machine Learning Research},
  volume={20},
  pages={105--1},
  year={2019}
}

@article{bernton2019parameter,
  title={On parameter estimation with the {W}asserstein distance},
  author={Bernton, Espen and Jacob, Pierre E and Gerber, Mathieu and Robert, Christian P},
  journal={Information and Inference: A Journal of the IMA},
  volume={8},
  number={4},
  pages={657--676},
  year={2019},
  publisher={Oxford University Press}
}

@inproceedings{arjovsky2017wasserstein,
  title={{W}asserstein Generative Adversarial Networks},
  author={Arjovsky, Martin and Chintala, Soumith and Bottou, L{\'e}on},
  booktitle={International Conference on Machine Learning},
  pages={214--223},
  year={2017}
}

@blog{ruishu2017,
  title={Amortized Optimization \url{http://ruishu.io/2017/11/07/amortized-optimization/}},
  author={Shu, Rui},
  booktitle={http://ruishu.io/2017/11/07/amortized-optimization/},
  year={2017}
}

@inproceedings{altschuler2017near,
  title={Near-linear time approximation algorithms for optimal transport via {S}inkhorn iteration},
  author={Altschuler, Jason and Niles-Weed, Jonathan and Rigollet, Philippe},
  booktitle={Advances in Neural Information Processing Systems},
  pages={1964--1974},
  year={2017}
}

@inproceedings{courty2017joint,
  title={Joint distribution optimal transportation for domain adaptation},
  author={Courty, Nicolas and Flamary, R{\'e}mi and Habrard, Amaury and Rakotomamonjy, Alain},
  booktitle={Advances in Neural Information Processing Systems},
  pages={3730--3739},
  year={2017}
}

@inproceedings{cuturi2013sinkhorn,
 author = {Cuturi, Marco},
 booktitle = {Advances in Neural Information Processing Systems},
 editor = {C.J. Burges and L. Bottou and M. Welling and Z. Ghahramani and K.Q. Weinberger},
 pages = {},
 publisher = {Curran Associates, Inc.},
 title = {Sinkhorn Distances: Lightspeed Computation of Optimal Transport},
 url = {https://proceedings.neurips.cc/paper_files/paper/2013/file/af21d0c97db2e27e13572cbf59eb343d-Paper.pdf},
 volume = {26},
 year = {2013}
}

@article{bonneel2015sliced,
  title={Sliced and {R}adon {W}asserstein Barycenters of Measures},
  author={Bonneel, Nicolas and Rabin, Julien and Peyr{\'e}, Gabriel and Pfister, Hanspeter},
  journal={Journal of Mathematical Imaging and Vision},
  volume={1},
  number={51},
  pages={22--45},
  year={2015}
}

@article{peyre2020computational,
  title={Computational optimal transport: With applications to data science},
  author={Peyr{\'e}, Gabriel and Cuturi, Marco and others},
  journal={Foundations and Trends{\textregistered} in Machine Learning},
  volume={11},
  number={5-6},
  pages={355--607},
  year={2019},
  publisher={Now Publishers, Inc.}
}

@book{villani2003topics,
  title={Topics in optimal transportation},
  author={Villani, C{\'e}dric},
  number={58},
  year={2003},
  publisher={American Mathematical Soc.}
}

@article{courty2016optimal,
  title={Optimal transport for domain adaptation},
  author={Courty, Nicolas and Flamary, R{\'e}mi and Tuia, Devis and Rakotomamonjy, Alain},
  journal={IEEE transactions on pattern analysis and machine intelligence},
  volume={39},
  number={9},
  pages={1853--1865},
  year={2016},
  publisher={IEEE}
}

@article{nguyen2025introduction,
  title={An introduction to sliced optimal transport: foundations, advances, extensions, and applications},
  author={Nguyen, Khai},
  journal={Foundations and Trends{\textregistered} in Computer Graphics and Vision},
  volume={17},
  number={3-4},
  pages={171--391},
  year={2025},
  publisher={Emerald Publishing Limited}
}

@inproceedings{patrini2020sinkhorn,
  title={Sinkhorn autoencoders},
  author={Patrini, Giorgio and van den Berg, Rianne and Forre, Patrick and Carioni, Marcello and Bhargav, Samarth and Welling, Max and Genewein, Tim and Nielsen, Frank},
  booktitle={Uncertainty in Artificial Intelligence},
  pages={733--743},
  year={2020},
  organization={PMLR}
}

@article{bernton2019approximate,
  title={Approximate {B}ayesian computation with the {W}asserstein distance},
  author={Bernton, Espen and Jacob, Pierre E and Gerber, Mathieu and Robert, Christian P},
  journal={Journal of the Royal Statistical Society Series B: Statistical Methodology},
  volume={81},
  number={2},
  pages={235--269},
  year={2019},
  publisher={Oxford University Press}
}

@article{catalano2024wasserstein,
  title={A {W}asserstein index of dependence for random measures},
  author={Catalano, Marta and Lavenant, Hugo and Lijoi, Antonio and Pr{\"u}nster, Igor},
  journal={Journal of the American Statistical Association},
  volume={119},
  number={547},
  pages={2396--2406},
  year={2024},
  publisher={Taylor \& Francis}
}

@inproceedings{scetbon2021low,
  title={Low-rank {S}inkhorn factorization},
  author={Scetbon, Meyer and Cuturi, Marco and Peyr{\'e}, Gabriel},
  booktitle={International Conference on Machine Learning},
  pages={9344--9354},
  year={2021},
  organization={PMLR}
}

@article{scetbon2022low,
  title={Low-rank optimal transport: Approximation, statistics and debiasing},
  author={Scetbon, Meyer and Cuturi, Marco},
  journal={Advances in Neural Information Processing Systems},
  volume={35},
  pages={6802--6814},
  year={2022}
}

@article{orlin1997polynomial,
  title={A polynomial time primal network simplex algorithm for minimum cost flows},
  author={Orlin, James B},
  journal={Mathematical Programming},
  volume={78},
  number={2},
  pages={109--129},
  year={1997},
  publisher={Springer}
}

@article{rigollet2025sample,
  title={On the sample complexity of entropic optimal transport},
  author={Rigollet, Philippe and Stromme, Austin J},
  journal={The Annals of Statistics},
  volume={53},
  number={1},
  pages={61--90},
  year={2025},
  publisher={Institute of Mathematical Statistics}
}

@inproceedings{rubner1998metric,
  title={A metric for distributions with applications to image databases},
  author={Rubner, Yossi and Tomasi, Carlo and Guibas, Leonidas J},
  booktitle={Sixth International Conference on Computer Vision (IEEE Cat. No. 98CH36271)},
  pages={59--66},
  year={1998},
  organization={IEEE}
}

@inproceedings{
kolouri2021wasserstein,
title={Wasserstein Embedding for Graph Learning},
author={Soheil Kolouri and Navid Naderializadeh and Gustavo K. Rohde and Heiko Hoffmann},
booktitle={International Conference on Learning Representations},
year={2021}
}

@article{genevay2016stochastic,
  title={Stochastic optimization for large-scale optimal transport},
  author={Genevay, Aude and Cuturi, Marco and Peyr{\'e}, Gabriel and Bach, Francis},
  journal={Advances in neural information processing systems},
  volume={29},
  year={2016}
}

@article{manole2024plugin,
  title={Plugin estimation of smooth optimal transport maps},
  author={Manole, Tudor and Balakrishnan, Sivaraman and Niles-Weed, Jonathan and Wasserman, Larry},
  journal={The Annals of Statistics},
  volume={52},
  number={3},
  pages={966--998},
  year={2024},
  publisher={Institute of Mathematical Statistics}
}

@inproceedings{genevay2019sample,
  title={Sample complexity of sinkhorn divergences},
  author={Genevay, Aude and Chizat, L{\'e}naic and Bach, Francis and Cuturi, Marco and Peyr{\'e}, Gabriel},
  booktitle={The 22nd international conference on artificial intelligence and statistics},
  pages={1574--1583},
  year={2019},
  organization={PMLR}
}

@inproceedings{bonneel2023survey,
  title={A survey of optimal transport for computer graphics and computer vision},
  author={Bonneel, Nicolas and Digne, Julie},
  booktitle={Computer Graphics Forum},
  volume={42},
  number={2},
  pages={439--460},
  year={2023},
  organization={Wiley Online Library}
}

@article{sinkhorn1967concerning,
  title={Concerning nonnegative matrices and doubly stochastic matrices},
  author={Sinkhorn, Richard and Knopp, Paul},
  journal={Pacific Journal of Mathematics},
  volume={21},
  number={2},
  pages={343--348},
  year={1967},
  publisher={Mathematical Sciences Publishers}
}

@article{kolouri2017optimal,
  title={Optimal mass transport: Signal processing and machine-learning applications},
  author={Kolouri, Soheil and Park, Se Rim and Thorpe, Matthew and Slepcev, Dejan and Rohde, Gustavo K},
  journal={IEEE signal processing magazine},
  volume={34},
  number={4},
  pages={43--59},
  year={2017},
  publisher={IEEE}
}

@inproceedings{jiang2020wasserstein,
  title={Wasserstein fair classification},
  author={Jiang, Ray and Pacchiano, Aldo and Stepleton, Tom and Jiang, Heinrich and Chiappa, Silvia},
  booktitle={Uncertainty in Artificial Intelligence},
  pages={862--872},
  year={2020},
  organization={PMLR}
}

@article{zhu2024functional,
  title={Functional optimal transport: regularized map estimation and domain adaptation for functional data},
  author={Zhu, Jiacheng and Guha, Aritra and Do, Dat and Xu, Mengdi and Nguyen, XuanLong and Zhao, Ding},
  journal={Journal of Machine Learning Research},
  volume={25},
  number={276},
  pages={1--49},
  year={2024}
}

@article{catalano2021measuring,
  title={Measuring dependence in the {W}asserstein distance for {B}ayesian nonparametric models},
  author={Catalano, Marta and Lijoi, Antonio and Pr{\"u}nster, Igor},
  journal={The Annals of Statistics},
  volume={49},
  number={5},
  pages={2916--2947},
  year={2021},
  publisher={Institute of Mathematical Statistics}
}

@article{solomon2015convolutional,
  title={Convolutional {W}asserstein distances: Efficient optimal transportation on geometric domains},
  author={Solomon, Justin and De Goes, Fernando and Peyr{\'e}, Gabriel and Cuturi, Marco and Butscher, Adrian and Nguyen, Andy and Du, Tao and Guibas, Leonidas},
  journal={ACM Transactions on Graphics (ToG)},
  volume={34},
  number={4},
  pages={1--11},
  year={2015},
  publisher={ACM New York, NY, USA}
}

@article{solomon2016entropic,
  title={Entropic metric alignment for correspondence problems},
  author={Solomon, Justin and Peyr{\'e}, Gabriel and Kim, Vladimir G and Sra, Suvrit},
  journal={ACM Transactions on Graphics (TOG)},
  volume={35},
  number={4},
  pages={72},
  year={2016},
  publisher={ACM}
}

@inproceedings{makkuva2020optimal,
  title={Optimal transport mapping via input convex neural networks},
  author={Makkuva, Ashok and Taghvaei, Amirhossein and Oh, Sewoong and Lee, Jason},
  booktitle={International Conference on Machine Learning},
  pages={6672--6681},
  year={2020},
  organization={PMLR}
}

@article{amos2022tutorial,
  title={Tutorial on amortized optimization},
  author={Amos, Brandon},
  journal={Foundations and Trends in Machine Learning},
  volume={16},
  number={5},
  pages={592--732},
  year={2023},
  publisher={Emerald Publishing Limited}
}

@inproceedings{
nguyen2026fast,
title={Fast Estimation of {W}asserstein Distances via Regression on Sliced {W}asserstein Distances},
author={Khai Nguyen and Hai Nguyen and Nhat Ho},
booktitle={The Fourteenth International Conference on Learning Representations},
year={2026},
url={https://openreview.net/forum?id=oa7L4vcJ77}
}

@inproceedings{bunne2022proximal,
  title={Proximal optimal transport modeling of population dynamics},
  author={Bunne, Charlotte and Papaxanthos, Laetitia and Krause, Andreas and Cuturi, Marco},
  booktitle={International Conference on Artificial Intelligence and Statistics},
  pages={6511--6528},
  year={2022},
  organization={PMLR}
}

@article{engquist2014application,
  title={Application of the {W}asserstein metric to seismic signals},
  author={Engquist, Bj{\"o}rn and Froese, Brittany D},
  journal={Communications in Mathematical Sciences},
  volume={12},
  number={5},
  pages={979--988},
  year={2014},
  publisher={International Press, Inc.}
}

@InProceedings{nguyen2022improving,
    title={Improving Mini-batch Optimal Transport via Partial Transportation}, 
    author={Khai Nguyen and Dang Nguyen and Tung Pham and Nhat Ho},
    booktitle = {Proceedings of the 39th International Conference on Machine Learning},
    year={2022},
}

@article{bonet2022spherical,
  title={Spherical Sliced-{W}asserstein},
  author={Bonet, Cl{\'e}ment and Berg, Paul and Courty, Nicolas and Septier, Fran{\c{c}}ois and Drumetz, Lucas and Pham, Minh-Tan},
  journal={International Conference on Learning Representations},
  year={2023}
}

@article{moosmuller2023linear,
  title={Linear optimal transport embedding: provable {W}asserstein classification for certain rigid transformations and perturbations},
  author={Moosm{\"u}ller, Caroline and Cloninger, Alexander},
  journal={Information and Inference: A Journal of the IMA},
  volume={12},
  number={1},
  pages={363--389},
  year={2023},
  publisher={Oxford University Press}
}

@inproceedings{
haviv2024wasserstein,
title={Wasserstein Wormhole: Scalable Optimal Transport Distance with {T}ransformer},
author={Doron Haviv and Russell Zhang Kunes and Thomas Dougherty and Cassandra Burdziak and Tal Nawy and Anna Gilbert and Dana Pe'er},
booktitle={Forty-first International Conference on Machine Learning},
year={2024}
}

@inproceedings{
courty2018learning,
title={Learning {W}asserstein Embeddings},
author={Nicolas Courty and Rémi Flamary and Mélanie Ducoffe},
booktitle={International Conference on Learning Representations},
year={2018},
url={https://openreview.net/forum?id=SJyEH91A-},
}

@article{santambrogio2015optimal,
  title={Optimal transport for applied mathematicians},
  author={Santambrogio, Filippo},
  journal={Birk{\"a}user, NY},
  volume={55},
  number={58-63},
  pages={94},
  year={2015},
  publisher={Springer}
}

@article{dowson1982frechet,
  title={The {F}r{\'e}chet distance between multivariate {N}ormal distributions},
  author={Dowson, DC and Landau, BV666017},
  journal={Journal of Multivariate Analysis},
  volume={12},
  number={3},
  pages={450--455},
  year={1982},
  publisher={Elsevier}
}

@inproceedings{achlioptas2018learning,
  title={Learning representations and generative models for 3d point clouds},
  author={Achlioptas, Panos and Diamanti, Olga and Mitliagkas, Ioannis and Guibas, Leonidas},
  booktitle={International Conference on Machine Learning},
  pages={40--49},
  year={2018},
  organization={PMLR}
}

@inproceedings{feydy2017optimal,
  title={Optimal transport for diffeomorphic registration},
  author={Feydy, Jean and Charlier, Benjamin and Vialard, Fran{\c{c}}ois-Xavier and Peyr{\'e}, Gabriel},
  booktitle={Medical Image Computing and Computer Assisted Intervention- MICCAI 2017: 20th International Conference, Quebec City, QC, Canada, September 11-13, 2017, Proceedings, Part I 20},
  pages={291--299},
  year={2017},
  organization={Springer}
}

@article{alvarez2020geometric,
  title={Geometric dataset distances via optimal transport},
  author={Alvarez-Melis, David and Fusi, Nicolo},
  journal={Advances in Neural Information Processing Systems},
  volume={33},
  pages={21428--21439},
  year={2020}
}

@article{quellmalz2023sliced,
  title={Sliced optimal transport on the sphere},
  author={Quellmalz, Michael and Beinert, Robert and Steidl, Gabriele},
  journal={Inverse Problems},
  volume={39},
  number={10},
  pages={105005},
  year={2023},
  publisher={IOP Publishing}
}

@article{tran2024stereographic,
  title={Stereographic Spherical Sliced {W}asserstein Distances},
  author={Tran, Huy and Bai, Yikun and Kothapalli, Abihith and Shahbazi, Ashkan and Liu, Xinran and Martin, Rocio Diaz and Kolouri, Soheil},
  journal={International Conference on Machine Learning},
  year={2024}
}

@article{wu2023improving,
  title={Improving molecular representation learning with metric learning-enhanced optimal transport},
  author={Wu, Fang and Courty, Nicolas and Jin, Shuting and Li, Stan Z},
  journal={Patterns},
  volume={4},
  number={4},
  year={2023},
  publisher={Elsevier}
}

@article{schiebinger2019optimal,
  title={Optimal-transport analysis of single-cell gene expression identifies developmental trajectories in reprogramming},
  author={Schiebinger, Geoffrey and Shu, Jian and Tabaka, Marcin and Cleary, Brian and Subramanian, Vidya and Solomon, Aryeh and Gould, Joshua and Liu, Siyan and Lin, Stacie and Berube, Peter and others},
  journal={Cell},
  volume={176},
  number={4},
  pages={928--943},
  year={2019},
  publisher={Elsevier}
}

@article{he2026sinkhorn,
  title={Sinkhorn-Drifting Generative Models},
  author={He, Ping and Khangaonkar, Om and Pirsiavash, Hamed and Bai, Yikun and Kolouri, Soheil},
  journal={arXiv preprint arXiv:2603.12366},
  year={2026}
}

@article{bunne2023learning,
  title={Learning single-cell perturbation responses using neural optimal transport},
  author={Bunne, Charlotte and Stark, Stefan G and Gut, Gabriele and Del Castillo, Jacobo Sarabia and Levesque, Mitch and Lehmann, Kjong-Van and Pelkmans, Lucas and Krause, Andreas and R{\"a}tsch, Gunnar},
  journal={Nature methods},
  volume={20},
  number={11},
  pages={1759--1768},
  year={2023},
  publisher={Nature Publishing Group US New York}
}

@article{
tong2024improving,
title={Improving and generalizing flow-based generative models with minibatch optimal transport},
author={Alexander Tong and Kilian FATRAS and Nikolay Malkin and Guillaume Huguet and Yanlei Zhang and Jarrid Rector-Brooks and Guy Wolf and Yoshua Bengio},
journal={Transactions on Machine Learning Research},
issn={2835-8856},
year={2024},
url={https://openreview.net/forum?id=CD9Snc73AW},
note={Expert Certification}
}

@inproceedings{pooladian2023multisample,
  title={Multisample Flow Matching: Straightening Flows with Minibatch Couplings},
  author={Pooladian, Aram-Alexandre and Ben-Hamu, Heli and Domingo-Enrich, Carles and Amos, Brandon and Lipman, Yaron and Chen, Ricky TQ},
  booktitle={International Conference on Machine Learning},
  pages={28100--28127},
  year={2023},
  organization={PMLR}
}

@inproceedings{
lipman2023flow,
title={Flow Matching for Generative Modeling},
author={Yaron Lipman and Ricky T. Q. Chen and Heli Ben-Hamu and Maximilian Nickel and Matthew Le},
booktitle={The Eleventh International Conference on Learning Representations },
year={2023},
url={https://openreview.net/forum?id=PqvMRDCJT9t}
}

@inproceedings{tolstikhin2018wasserstein,
  title={{W}asserstein Auto-Encoders},
  author={Tolstikhin, Ilya and Bousquet, Olivier and Gelly, Sylvain and Schoelkopf, Bernhard},
  booktitle={International Conference on Learning Representations},
  year={2018}
}

@article{garrett2024validating,
  title={Validating climate models with spherical convolutional {W}asserstein distance},
  author={Garrett, Robert C and Harris, Trevor and Wang, Zhuo and Li, Bo},
  journal={Advances in Neural Information Processing Systems},
  volume={37},
  pages={59119--59149},
  year={2024}
}

@article{tanguy2025sliced,
  title={Sliced optimal transport plans},
  author={Tanguy, Eloi and Chapel, Laetitia and Delon, Julie},
  journal={arXiv preprint arXiv:2508.01243},
  year={2025}
}

@inproceedings{
liu2025expected,
title={Expected Sliced Transport Plans},
author={Xinran Liu and Rocio Diaz Martin and Yikun Bai and Ashkan Shahbazi and Matthew Thorpe and Akram Aldroubi and Soheil Kolouri},
booktitle={The Thirteenth International Conference on Learning Representations},
year={2025},
url={https://openreview.net/forum?id=P7O1Vt1BdU}
}

@article{bonet2024sliced,
  title={Sliced-{W}asserstein distances and flows on {C}artan-{H}adamard manifolds},
  author={Bonet, Cl{\'e}ment and Drumetz, Lucas and Courty, Nicolas},
  journal={Journal of Machine Learning Research},
  volume={26},
  number={32},
  pages={1--76},
  year={2025}
}

@article{benamou2015iterative,
  title={Iterative {B}regman projections for regularized transportation problems},
  author={Benamou, Jean-David and Carlier, Guillaume and Cuturi, Marco and Nenna, Luca and Peyr{\'e}, Gabriel},
  journal={SIAM Journal on Scientific Computing},
  volume={37},
  number={2},
  pages={A1111--A1138},
  year={2015},
  publisher={SIAM}
}

@inproceedings{cuturi2014fast,
  title={Fast computation of Wasserstein barycenters},
  author={Cuturi, Marco and Doucet, Arnaud},
  booktitle={International Conference on Machine Learning},
  pages={685--693},
  year={2014},
  organization={PMLR}
}

@inproceedings{
nguyen2024quasimonte,
title={Quasi-{M}onte {C}arlo for 3D Sliced {W}asserstein},
author={Khai Nguyen and Nicola Bariletto and Nhat Ho},
booktitle={The Twelfth International Conference on Learning Representations},
year={2024},
url={https://openreview.net/forum?id=Wd47f7HEXg}
}

@inproceedings{mahey2023fast,
  title={Fast optimal transport through sliced {W}asserstein generalized geodesics},
  author={Mahey, Guillaume and Chapel, Laetitia and Gasso, Gilles and Bonet, Cl{\'e}ment and Courty, Nicolas},
  booktitle={Proceedings of the 37th International Conference on Neural Information Processing Systems},
  pages={35350--35385},
  year={2023}
}

@article{liu2025efficient,
  title={Efficient Transferable Optimal Transport via Min-Sliced Transport Plans},
  author={Liu, Xinran and Akbari, Elaheh and Diaz Martin, Rocio and NaderiAlizadeh, Navid and Kolouri, Soheil},
  journal={arXiv e-prints},
  pages={arXiv--2511},
  year={2025}
}

@inproceedings{amos2023meta,
  title={Meta Optimal Transport},
  author={Amos, Brandon and Luise, Giulia and Cohen, Samuel and Redko, Ievgen},
  booktitle={International Conference on Machine Learning},
  pages={791--813},
  year={2023},
  organization={PMLR}
}

@article{Doxsey-Whitfield03072015,
author = {Erin Doxsey-Whitfield and Kytt MacManus and Susana B. Adamo and Linda Pistolesi and John Squires and Olena Borkovska and Sandra R. Baptista},
title = {Taking Advantage of the Improved Availability of Census Data: A First Look at the Gridded Population of the World, Version 4},
journal = {Papers in Applied Geography},
volume = {1},
number = {3},
pages = {226--234},
year = {2015},
publisher = {Routledge},
doi = {10.1080/23754931.2015.1014272},


URL = { 
    
        https://doi.org/10.1080/23754931.2015.1014272
    
    

},
eprint = { 
    
        https://doi.org/10.1080/23754931.2015.1014272
    
    

}

}


\appendix

\section{Experiment Details}
\label{app:impl}

In this section, we detail the architectures, hyperparameter settings, and optimization constraints for all methods utilized in the main experiments. All configuration values are dynamically loaded from a centralized configuration file to ensure exact reproducibility across multiple experimental runs. To guarantee a fair evaluation, methods requiring gradient-based optimization are allocated an equivalent training budget (5,000 steps). Optimizers (Adam) and learning rates werefixed per algorithm based on their respective convergence properties. The entropic regularization parameter ($\varepsilon$) defines the inherent Sinkhorn OT problem and is maintained uniformly across all baselines for consistent ground truth definition ($\varepsilon=0.1$ for Grayscale, $\varepsilon=0.5$ for spherical transport, and $\varepsilon=0.005$ for color transfer). All experiments were conducted on a single Google Colab T4 GPU, except for the CIFAR-10 fine-tuning, which utilized a 40GB A100 GPU.

\begin{table}[htbp]
\centering
\caption{Hyperparameters for \textbf{RA-OT} (Regression-based Amortized OT). }
\label{tab:cfg_ra_ot}
\small
\begin{tabular}{l c c c}
\toprule
\textbf{Parameter} & \textbf{MNIST Grayscale} & \textbf{World Pair} & \textbf{Color Transfer} \\
\midrule
Number of Projections ($L$) & 100 & 100 & 100 \\
Ridge Coefficient ($\lambda$) & $1 \times 10^{-3}$ & $1 \times 10^{-3}$ & $1 \times 10^{-3}$ \\

\bottomrule
\end{tabular}
\end{table}

\begin{table}[htbp]
\centering
\caption{Hyperparameters for \textbf{OA-OT} (Objective-based Amortized OT).}
\label{tab:cfg_oa_ot}
\small
\begin{tabular}{l c c c}
\toprule
\textbf{Parameter} & \textbf{MNIST Grayscale} & \textbf{World Pair} & \textbf{Color Transfer} \\
\midrule
Number of Projections ($L$) & 100 & 100 & 100 \\
Training Iterations ($T$) & 5,000 & 5,000 & 5,000 \\
Learning Rate & $1 \times 10^{-3}$ & $1 \times 10^{-3}$ & $1 \times 10^{-3}$ \\

\bottomrule
\end{tabular}
\end{table}

\begin{table}[htbp]
\centering
\caption{Hyperparameters for the \textbf{Meta-OT} baseline \citep{amos2023meta}.}
\label{tab:cfg_meta_ot}
\small
\begin{tabular}{p{3.5cm} c p{3.2cm} p{3.5cm}}
\toprule
\textbf{Parameter} & \textbf{MNIST Grayscale} & \textbf{World Pair} & \textbf{Color Transfer} \\
\midrule
Architecture Model & MLP & MLP & PointCloud Encoder \\
Hidden Layers / Units & 3 Layers & 3 Layers, 512 Units & 256 Enc Dim, 512 Head \\
Training Iterations ($T$) & 5,000 & 5,000  & 5,000  \\
Learning Rate & $1 \times 10^{-3}$ & $1 \times 10^{-3}$ & $1 \times 10^{-3}$ \\

\bottomrule
\end{tabular}
\end{table}

\begin{table}[!ht] 
\centering
\caption{Hyperparameters for the amortized \textbf{Min-STP} baseline \citep{liu2025efficient}. $\alpha$ denotes the inherent temperature parameter used during continuous soft-sorting.}
\label{tab:cfg_min_stp}
\small
\begin{tabular}{l c c c}
\toprule
\textbf{Parameter} & \textbf{MNIST Grayscale} & \textbf{World Pair} & \textbf{Color Transfer} \\
\midrule
Encoder/Context Dim & 32 & 32 & 32 \\
Hidden Dim & 128 & 128 & 64 \\
Training Iterations ($T$) & 5,000 & 5,000 & 5,000 \\
Learning Rate & $1 \times 10^{-3}$ & $1 \times 10^{-3}$ & $1 \times 10^{-3}$ \\
Soft-Sort Temperature ($\alpha$) & 0.05 & 0.05 & 0.05 \\
\bottomrule
\end{tabular}
\end{table}


\section{Ablation Study: Effect of the Number of Projections L}
\label{Vary_L}

\begin{table}[htbp]
\centering
\caption{Ablation study on the effect of the number of sliced projections $L \in \{3, 5, 10, 20, 50, 100\}$. The performance of RA-OT and OA-OT is evaluated across three tasks: \textbf{MNIST} grayscale transport, \textbf{Spherical} supply-demand, and \textbf{Color Transfer}. Results are reported in terms of Plan RMSE ($\times 10^{-6}$), training time, and inference time. \textbf{Bold} indicates the best metric between the two methods}
\label{tab:ablation_L}
\resizebox{\linewidth}{!}{%
\begin{tabular}{@{} c l l c c c @{\hspace{1.5em}} c l l c c c @{}}
\toprule
$L$ & Task & Method & RMSE\,($\times 10^{-6}$, $\downarrow$) & Train\,(s) & Infer\,(ms) & 
$L$ & Task & Method & RMSE\,($\times 10^{-6}$, $\downarrow$) & Train\,(s) & Infer\,(ms) \\
\cmidrule(r){1-6} \cmidrule(l){7-12}

\multirow{6}{*}{3} 
 & \multirow{2}{*}{MNIST} & RA-OT & $9.25 \pm 3.73$ & $\mathbf{3.22}$ & $37.47 \pm 3.91$ & 
\multirow{6}{*}{20} 
 & \multirow{2}{*}{MNIST} & RA-OT & $7.62 \pm 3.11$ & $\mathbf{3.18}$ & $38.11 \pm 2.45$ \\
 & & OA-OT & $\mathbf{8.50 \pm 3.55}$ & $15.35$ & $\mathbf{37.38 \pm 2.65}$ & 
 & & OA-OT & $\mathbf{6.05 \pm 2.53}$ & $15.35$ & $\mathbf{37.83 \pm 2.23}$ \\
 \cmidrule(r){2-6} \cmidrule(l){8-12}
 
 & \multirow{2}{*}{Spherical} & RA-OT & $0.71 \pm 0.17$ & $\mathbf{2.17}$ & $\mathbf{30.51 \pm 4.43}$ & 
 & \multirow{2}{*}{Spherical} & RA-OT & $0.74 \pm 0.19$ & $\mathbf{2.33}$ & $32.06 \pm 4.97$ \\
 & & OA-OT & $\mathbf{0.41 \pm 0.16}$ & $18.62$ & $34.20 \pm 5.70$ & 
 & & OA-OT & $\mathbf{0.38 \pm 0.16}$ & $18.23$ & $\mathbf{30.89 \pm 4.24}$ \\
 \cmidrule(r){2-6} \cmidrule(l){8-12}
 
 & \multirow{2}{*}{Color} & RA-OT & $25.60 \pm 10.42$ & $\mathbf{6.68}$ & $\mathbf{16.48 \pm 0.96}$ & 
 & \multirow{2}{*}{Color} & RA-OT & $9.39 \pm 5.50$ & $\mathbf{7.03}$ & $17.64 \pm 1.28$ \\
 & & OA-OT & $\mathbf{23.96 \pm 6.79}$ & $16.86$ & $17.29 \pm 1.15$ & 
 & & OA-OT & $\mathbf{9.11 \pm 5.11}$ & $17.31$ & $\mathbf{17.48 \pm 1.37}$ \\
\cmidrule(r){1-6} \cmidrule(l){7-12}

\multirow{6}{*}{5} 
 & \multirow{2}{*}{MNIST} & RA-OT & $8.80 \pm 3.53$ & $\mathbf{3.62}$ & $40.89 \pm 3.02$ & 
\multirow{6}{*}{50} 
 & \multirow{2}{*}{MNIST} & RA-OT & $7.66 \pm 3.05$ & $\mathbf{2.98}$ & $37.73 \pm 3.34$ \\
 & & OA-OT & $\mathbf{8.01 \pm 3.41}$ & $16.10$ & $\mathbf{40.76 \pm 3.06}$ & 
 & & OA-OT & $\mathbf{6.03 \pm 2.52}$ & $15.19$ & $\mathbf{37.71 \pm 2.40}$ \\
 \cmidrule(r){2-6} \cmidrule(l){8-12}
 
 & \multirow{2}{*}{Spherical} & RA-OT & $0.70 \pm 0.18$ & $\mathbf{2.37}$ & $\mathbf{31.90 \pm 4.62}$ & 
 & \multirow{2}{*}{Spherical} & RA-OT & $0.78 \pm 0.19$ & $\mathbf{2.46}$ & $35.53 \pm 4.06$ \\
 & & OA-OT & $\mathbf{0.41 \pm 0.16}$ & $18.60$ & $33.11 \pm 6.03$ & 
 & & OA-OT & $\mathbf{0.39 \pm 0.19}$ & $18.47$ & $\mathbf{33.04 \pm 3.95}$ \\
 \cmidrule(r){2-6} \cmidrule(l){8-12}
 
 & \multirow{2}{*}{Color} & RA-OT & $\mathbf{13.16 \pm 6.95}$ & $\mathbf{7.25}$ & $18.69 \pm 1.75$ & 
 & \multirow{2}{*}{Color} & RA-OT & $9.76 \pm 5.62$ & $\mathbf{6.91}$ & $17.52 \pm 1.51$ \\
 & & OA-OT & $14.83 \pm 6.22$ & $18.93$ & $\mathbf{18.67 \pm 0.89}$ & 
 & & OA-OT & $\mathbf{9.04 \pm 5.02}$ & $17.48$ & $\mathbf{17.51 \pm 1.17}$ \\
\cmidrule(r){1-6} \cmidrule(l){7-12}

\multirow{6}{*}{10} 
 & \multirow{2}{*}{MNIST} & RA-OT & $7.85 \pm 3.22$ & $\mathbf{3.20}$ & $\mathbf{37.56 \pm 2.80}$ & 
\multirow{6}{*}{100} 
 & \multirow{2}{*}{MNIST} & RA-OT & $7.77 \pm 3.06$ & $\mathbf{3.03}$ & $39.36 \pm 3.61$ \\
 & & OA-OT & $\mathbf{6.34 \pm 2.63}$ & $14.44$ & $37.90 \pm 2.74$ & 
 & & OA-OT & $\mathbf{6.02 \pm 2.52}$ & $15.78$ & $\mathbf{38.92 \pm 2.23}$ \\
 \cmidrule(r){2-6} \cmidrule(l){8-12}
 
 & \multirow{2}{*}{Spherical} & RA-OT & $0.67 \pm 0.18$ & $\mathbf{2.19}$ & $\mathbf{32.01 \pm 4.35}$ & 
 & \multirow{2}{*}{Spherical} & RA-OT & $0.78 \pm 0.19$ & $\mathbf{2.53}$ & $41.96 \pm 6.29$ \\
 & & OA-OT & $\mathbf{0.42 \pm 0.17}$ & $18.54$ & $35.13 \pm 5.99$ & 
 & & OA-OT & $\mathbf{0.39 \pm 0.19}$ & $19.37$ & $\mathbf{41.03 \pm 2.76}$ \\
 \cmidrule(r){2-6} \cmidrule(l){8-12}
 
 & \multirow{2}{*}{Color} & RA-OT & $9.63 \pm 5.47$ & $\mathbf{6.48}$ & $17.14 \pm 1.59$ & 
 & \multirow{2}{*}{Color} & RA-OT & $9.99 \pm 5.61$ & $\mathbf{7.40}$ & $\mathbf{17.40 \pm 0.83}$ \\
 & & OA-OT & $\mathbf{9.38 \pm 5.23}$ & $17.14$ & $\mathbf{17.08 \pm 1.03}$ & 
 & & OA-OT & $\mathbf{9.00 \pm 5.02}$ & $18.13$ & $17.76 \pm 0.79$ \\
\bottomrule
\end{tabular}%
}

\end{table}

To study the effect of the number of projections, we vary $L \in \{3, 5, 10, 20, 50, 100\}$. As shown in Table~\ref{tab:ablation_L}, both RA-OT and OA-OT benefit from increasing $L$, with clear improvements when moving away from very small values such as $L=3$ or $5$. With too few projections, the model cannot capture enough structural information, which leads to weaker performance. This is most noticeable in the Color Transfer task, where the data distribution is more complex and requires a richer representation. As $L$ increases, the error decreases quickly at first, and the results also become more stable across runs. However, after around $L=20$, the improvements start to level off, and adding more projections only brings small gains. This suggests that a moderate number of projections is already enough to capture most of the important structure in the data. Importantly, increasing $L$ does not significantly affect efficiency. Both training and inference times remain nearly the same across all values of $L$, showing that the method scales well with the number of projections. In practice, this means we can use a reasonably large $L$ to ensure good performance without worrying about  computational cost.

\section{High-dimensional Flow Matching Fine-tuning on CIFAR-10}
\label{ab: high_dim_cfm}

In addition to 2D  experiments, we investigate the scalability and practical advantages of our amortized methods on high-dimensional image generation. Specifically, we focus on fine-tuning a converged Continuous Normalizing Flow model on the CIFAR-10 dataset using an large batch setting.

We employ a widely adopted U-Net architecture containing approximately 35M parameters. Following the standard protocol \citep{tong2024improving}, we utilize a converged Independent CFM (I-CFM) checkpoint natively pretrained from scratch for 400,000 steps with a learning rate of $2 \times 10^{-4}$ and a batch size of $128$. Rather than training all configurations from scratch (which incurs extreme computational cost), we establish a rigorous fine-tuning protocol. All comparative methods natively load the equivalent I-CFM converged checkpoint (400000 steps training) and fine-tune for strictly 10 epochs. This setup establishes an absolutely fair ground where every method originates from an identical prior capacity and uniquely differs strictly by the assigned mini-batch coupling strategies (I-CFM, OT-CFM, OA-OT, and RA-OT). To avoid catastrophic forgetting, we reduce the learning rate to $5 \times 10^{-5}$, apply a short 100-step warmup, tighten gradient clipping to $0.5$, and lower the EMA decay from $0.9999$ to $0.999$ for faster adaptation. For amortized methods, predictors are constructed from $M=50$ mini-batches with $L=100$ sliced projections. All runtime results are reported under a large batch size ($\mathcal{B}=2048$).
About evaluation, we sample 50,000 outputs per model and evaluate the FID and NFE via the adaptive \texttt{dopri5} numerical solver (with tolerance $10^{-5}$). 
As shown in Table \ref{tab:cifar10_finetune}, directly applying the same optimal transport setup (OT-CFM) to large-batch training only increases the computational burden. In contrast, our precomputation strategies (RA-OT and OA-OT) effectively bypass this bottleneck by generating coupling plans much more efficiently, significantly reducing fine-tuning time. More interestingly, amortizing optimal transport does more than just speed things up, it also improves performance. OA-OT produces straighter trajectories, achieving the best NFE of $146.00$, while RA-OT delivers the strongest perceptual quality, reaching an FID of $3.543$.

\begin{table}[ht]
\caption{Adaptive sampling generation results (\texttt{dopri5}) fine-tuning Flow Matching models on CIFAR-10. Despite operating on an extreme scaling batch ($\mathcal{B}=2048$), our amortized paradigms (OA-OT, RA-OT) drastically accelerate iterative epochs relative to exact OT-CFM while yielding highly superior FID scores and tightly constrained NFE parameters.}
\label{tab:cifar10_finetune}
\centering
\begin{tabular}{l cccc}
\toprule
\textbf{Method} & \textbf{FID} $\downarrow$ & \textbf{NFE/sample} $\downarrow$ & \textbf{Training Time (s)} $\downarrow$ & \textbf{Pretrain Time (s)} $\downarrow$ \\
\midrule
ICFM & 3.638 & 146.61 & \textbf{481.9} & 0.0 \\
OT-CFM & 3.630 & 146.86 & 744.2 & 0.0 \\
OA-OT & 3.575 & \textbf{146.00} & 607.0 & \textbf{12.0} \\
RA-OT & \textbf{3.543} & 147.10 & 618.1 & 16.4 \\
\bottomrule
\end{tabular}
\end{table}

\section{Additional Experiments}
\label{sec:additional_experiments}

In this section, we provide additional experiments to further examine several properties of the proposed amortized OT framework. In particular, we study its ability to handle measures with varying numbers of atoms, the expressivity of using a global coefficient vector across heterogeneous measure pairs, its use for warm-starting iterative Sinkhorn solvers, robustness under distribution shift, the marginal feasibility of the predicted transportation plans, and the trade-off between training and inference time across different numbers of test problems.

\paragraph{Transportation between measures with different numbers of atoms.}
A key property of our amortized model is that its parameters are independent of the number of atoms in the input measures. To empirically evaluate this property, we construct a multi-resolution MNIST benchmark in which each image is randomly resized to one of three resolutions, namely $28\times 28$, $20\times 20$, and $14\times 14$, corresponding to measures with $784$, $400$, and $196$ atoms, respectively. For both the training set ($M=50$) and the test set ($N=300$), the resolutions of the source and target images are sampled independently. As a result, the benchmark contains both same-resolution and cross-resolution transportation problems, including, for example, transportation between a $784$-atom measure and a $196$-atom measure. We train a single RA-OT or OA-OT model on the resulting mixed-resolution training set and evaluate the same model on all test pairs without any architectural modification. This setting is particularly suitable for our construction since the coefficient vector $\omega\in\mathbb{R}^{L}$ does not depend on the dimensions of the input measures. In contrast, the fixed-dimensional MLP architecture used by Meta-OT requires a fixed number of input atoms and therefore cannot be directly applied to this heterogeneous setting.

Table~\ref{tab:multires_breakdown} reports the prediction errors separately for same-resolution and cross-resolution pairs. Both RA-OT and OA-OT remain stable across the two cases, indicating that the proposed representation can naturally accommodate transportation problems with different support sizes. Table~\ref{tab:multires_overall} further reports the overall performance. Both methods use only $L=100$ learned coefficients while operating on all combinations of resolutions within the same amortized model.

\begin{table}[t]
    \centering
    \caption{Plan RMSE on the multi-resolution MNIST benchmark, separated into same-resolution and cross-resolution source-target pairs.}
    \label{tab:multires_breakdown}
    \begin{tabular}{lcc}
        \toprule
        Method & Same-resolution RMSE & Cross-resolution RMSE \\
        \midrule
        RA-OT & $7.37\times 10^{-5} \pm 8.33\times 10^{-5}$
              & $4.20\times 10^{-5} \pm 3.02\times 10^{-5}$ \\
        OA-OT & $4.37\times 10^{-5} \pm 4.69\times 10^{-5}$
              & $2.50\times 10^{-5} \pm 1.55\times 10^{-5}$ \\
        \bottomrule
    \end{tabular}
\end{table}

\begin{table}[t]
    \centering
    \caption{Overall performance on the multi-resolution MNIST benchmark. A single amortized model is used for measure pairs with different numbers of atoms.}
    \label{tab:multires_overall}
    \begin{tabular}{lcccc}
        \toprule
        Method & Plan RMSE & Train (s) & Infer (ms) & \# Params \\
        \midrule
        RA-OT
        & $5.18\times 10^{-5} \pm 5.50\times 10^{-5}$
        & 2.3
        & $15.50 \pm 11.79$
        & 100 \\
        OA-OT
        & $3.08\times 10^{-5} \pm 3.00\times 10^{-5}$
        & 17.1
        & $15.31 \pm 11.01$
        & 100 \\
        \bottomrule
    \end{tabular}
\end{table}

\paragraph{Expressivity of a global coefficient vector on heterogeneous measures.}
The linear amortized model considered in this work uses a single coefficient vector $\omega$ across measure pairs drawn from the underlying meta-distribution. We further investigate whether sharing this coefficient vector introduces an expressivity bottleneck when the measure pairs are heterogeneous. To this end, we partition MNIST digits into two disjoint groups: Group A containing digits $0$-$4$ and Group B containing digits $5$-$9$. For each group, we use $M=20$ training pairs and $N=25$ test pairs.

We first estimate two group-specific RA-OT models, yielding coefficient vectors $\omega_A$ and $\omega_B$, using only training pairs from their corresponding groups. We then estimate a global coefficient vector $\omega_{\mathrm{global}}$ using the combined training set of $M=40$ pairs from both groups. Table~\ref{tab:global_coeff_rmse} compares the resulting plan prediction errors. Despite the substantial heterogeneity between the two subsets, the global coefficient performs comparably to, and in some cases slightly better than, the corresponding group-specific coefficients. To further examine the learned representations, Table~\ref{tab:global_coeff_cosine} reports the cosine similarities between the coefficient vectors. Interestingly, $\omega_A$ and $\omega_B$ are nearly orthogonal, with a cosine similarity of $-0.0294$, suggesting that the two subsets favor substantially different combinations of sliced potentials. Nevertheless, $\omega_{\mathrm{global}}$ maintains good prediction accuracy on both groups. These results indicate that, in this experiment, using a shared global coefficient does not introduce a substantial expressivity bottleneck and provides a parsimonious model across heterogeneous measures. We emphasize that the amortized formulation in Section~3.2 is not restricted to linear operators, and more expressive choices of $\gamma_\phi$ can be considered when needed.

\begin{table}[htbp]
    \centering
    \caption{Plan RMSE obtained using a global RA-OT coefficient vector and group-specific coefficient vectors on heterogeneous MNIST subsets. The global model is trained on the combined training pool from both groups.}
    \label{tab:global_coeff_rmse}
    \resizebox{\textwidth}{!}{%
    \begin{tabular}{lccc}
        \toprule
        Test set
        & $\omega_{\mathrm{global}}$ (Mixed)
        & $\omega_A$ (Digits 0-4)
        & $\omega_B$ (Digits 5-9) \\
        \midrule
        Test A (Digits 0-4)
        & $8.11\times 10^{-6} \pm 3.40\times 10^{-6}$
        & $8.95\times 10^{-6} \pm 3.09\times 10^{-6}$
        & $9.65\times 10^{-6} \pm 6.05\times 10^{-6}$ \\
        Test B (Digits 5-9)
        & $7.14\times 10^{-6} \pm 2.73\times 10^{-6}$
        & $9.13\times 10^{-6} \pm 3.91\times 10^{-6}$
        & $7.40\times 10^{-6} \pm 3.13\times 10^{-6}$ \\
        Test Combined
        & $7.62\times 10^{-6} \pm 3.12\times 10^{-6}$
        & - & - \\
        \bottomrule
    \end{tabular}%
    }
\end{table}

\begin{table}[htbp]
    \centering
    \caption{Cosine similarity between the RA-OT coefficient vectors learned from the two MNIST subsets and the combined training pool.}
    \label{tab:global_coeff_cosine}
    \begin{tabular}{lc}
        \toprule
        Vector pair & Cosine similarity \\
        \midrule
        $(\omega_A,\omega_B)$ & $-0.0294$ \\
        $(\omega_A,\omega_{\mathrm{global}})$ & $0.7520$ \\
        $(\omega_B,\omega_{\mathrm{global}})$ & $0.5762$ \\
        \bottomrule
    \end{tabular}
\end{table}

\paragraph{Warm-starting Sinkhorn with predicted Kantorovich potentials.}
Although our main experiments directly recover transportation plans from the predicted potentials, the amortized potentials can also be used to initialize iterative OT solvers. We investigate this use case by using the potentials predicted by RA-OT, OA-OT, and Meta-OT to warm-start the standard Sinkhorn iterations. We compare these initializations against the conventional cold start using zero potentials.

Using $N=100$ test pairs, we track the $L_1$ marginal error at fixed iteration checkpoints up to $1{,}000$ Sinkhorn iterations. The results are reported in Table~\ref{tab:sinkhorn_warmstart}. The predicted potentials provide substantially better initializations than a cold start. In particular, after one Sinkhorn iteration, OA-OT obtains a marginal error of $1.41\times 10^{-1}$, compared with $2.55\times 10^{-1}$ for Meta-OT and $5.35\times 10^{-1}$ for the cold start. This advantage remains visible during the early iterations: at iteration $10$, OA-OT reaches a marginal error of $1.63\times 10^{-2}$, which is below the errors obtained by the other initializations at the same iteration. As expected, all initialization schemes eventually converge to machine precision when sufficiently many Sinkhorn iterations are performed. These results show that the proposed amortized potentials can also serve as useful warm-starts when an exact or more tightly converged Sinkhorn solution is desired.

\begin{table*}[htbp]
    \centering
    \caption{$L_1$ marginal error during Sinkhorn iterations under different initialization strategies. The predicted Kantorovich potentials from Meta-OT, RA-OT, and OA-OT are used as warm-starts, while the cold-start baseline initializes the potentials at zero. \textbf{Bold} indicates the best performance.}
    \label{tab:sinkhorn_warmstart}
    \resizebox{\textwidth}{!}{
    \begin{tabular}{lcccccccccc}
        \toprule
        Init. method
        & Iter. 0 & Iter. 1 & Iter. 2 & Iter. 5 & Iter. 10
        & Iter. 20 & Iter. 50 & Iter. 100 & Iter. 200 & Iter. 1000 \\
        \midrule
        Cold Start
        & $1.61\times 10^{4}$
        & $5.35\times 10^{-1}$
        & $2.08\times 10^{-1}$
        & $6.93\times 10^{-2}$
        & $2.81\times 10^{-2}$
        & $9.78\times 10^{-3}$
        & $8.72\times 10^{-4}$
        & $2.83\times 10^{-5}$
        & $\bm{1.22\times 10^{-7}}$
        & $4.54\times 10^{-16}$ \\
        Meta-OT
        & $4.18\times 10^{-1}$
        & $2.55\times 10^{-1}$
        & $1.52\times 10^{-1}$
        & $6.23\times 10^{-2}$
        & $2.69\times 10^{-2}$
        & $9.42\times 10^{-3}$
        & $8.18\times 10^{-4}$
        & $2.76\times 10^{-5}$
        & $1.63\times 10^{-7}$
        & $6.98\times 10^{-16}$ \\
        RA-OT
        & $3.24\times 10^{-1}$
        & $1.79\times 10^{-1}$
        & $1.09\times 10^{-1}$
        & $5.23\times 10^{-2}$
        & $2.62\times 10^{-2}$
        & $9.70\times 10^{-3}$
        & $8.23\times 10^{-4}$
        & $2.97\times 10^{-5}$
        & $2.88\times 10^{-7}$
        & $4.44\times 10^{-16}$ \\
        OA-OT
        & $\bm{2.84\times 10^{-1}}$
        & $\bm{1.41\times 10^{-1}}$
        & $\bm{7.95\times 10^{-2}}$
        & $\bm{3.46\times 10^{-2}}$
        & $\bm{1.63\times 10^{-2}}$
        & $\bm{5.96\times 10^{-3}}$
        & $\bm{5.32\times 10^{-4}}$
        & $\bm{2.09\times 10^{-5}}$
        & $1.95\times 10^{-7}$
        & $\bm{4.29\times 10^{-16}}$ \\
        \bottomrule
    \end{tabular}
    }
\end{table*}

\paragraph{Robustness under rotational distribution shift.}
The amortized OT formulation assumes that future measure pairs are sampled from a meta-distribution related to that observed during training. We do not assume that the learned amortized model is invariant to arbitrary distribution shifts. Nevertheless, we conduct an additional experiment to examine its behavior under a simple geometric shift. Following the Rotated MNIST setting, we train all amortized models using only unrotated images ($0^\circ$) and evaluate them without further adaptation on images rotated by $30^\circ$, $60^\circ$, and $90^\circ$. Table~\ref{tab:rotated_mnist} reports the resulting plan RMSE. Both RA-OT and OA-OT remain relatively stable across the evaluated rotations. For example, the RMSE of OA-OT varies from $6.02\times 10^{-6}$ at $0^\circ$ to $6.45\times 10^{-6}$ at $90^\circ$. RA-OT exhibits a similarly moderate change. Both methods also retain lower reconstruction errors than Meta-OT across all evaluated rotation angles. These observations suggest that the sliced potentials preserve useful geometric information under this particular rotational shift. However, this experiment only considers one structured form of distribution shift, and a systematic study of OOD generalization for amortized OT remains an interesting direction for future work.

\begin{table}[htbp]
    \centering
    \caption{Plan RMSE under rotational distribution shift on MNIST. All amortized models are trained using only images at $0^\circ$ and evaluated zero-shot at the indicated rotation angles.}
    \label{tab:rotated_mnist}
    \begin{tabular}{lcccc}
        \toprule
        Method & $0^\circ$ & $30^\circ$ & $60^\circ$ & $90^\circ$ \\
        \midrule
        RA-OT
        & $7.77\times 10^{-6}$
        & $7.21\times 10^{-6}$
        & $7.31\times 10^{-6}$
        & $8.34\times 10^{-6}$ \\
        OA-OT
        & $\bm{6.02\times 10^{-6}}$
        & $\bm{5.19\times 10^{-6}}$
        & $\bm{5.31\times 10^{-6}}$
        & $\bm{6.45\times 10^{-6}}$ \\
        Meta-OT
        & $1.84\times 10^{-5}$
        & $1.63\times 10^{-5}$
        & $1.63\times 10^{-5}$
        & $1.84\times 10^{-5}$ \\
        \bottomrule
    \end{tabular}
\end{table}

\begin{table}[!b]
    \centering
    \caption{Plan RMSE and $L_1$ marginal violations before and after applying the rounding procedure of \citep{altschuler2017near}. Results are reported for the MNIST experiment with $M=50$, $N=300$, and $\epsilon=10^{-2}$.}
    \label{tab:marginal_rounding}
    \begin{tabular}{llccc}
        \toprule
        Method & Stage & Plan RMSE & $\mathrm{MargErr}_a$ & $\mathrm{MargErr}_b$ \\
        \midrule
        RA-OT
        & Before
        & $7.77\times 10^{-6} \pm 3.1\times 10^{-6}$
        & $1.818\times 10^{-1}$
        & $4.225\times 10^{-9}$ \\
        & After
        & $6.52\times 10^{-6} \pm 2.5\times 10^{-6}$
        & $1.358\times 10^{-16}$
        & $2.941\times 10^{-16}$ \\
        \midrule
        OA-OT
        & Before
        & $6.02\times 10^{-6} \pm 2.5\times 10^{-6}$
        & $1.441\times 10^{-1}$
        & $4.171\times 10^{-9}$ \\
        & After
        & $\bm{4.91\times 10^{-6} \pm 2.0\times 10^{-6}}$
        & $\bm{1.340\times 10^{-16}}$
        & $2.998\times 10^{-16}$ \\
        \midrule
        Meta-OT
        & Before
        & $1.55\times 10^{-5} \pm 4.7\times 10^{-6}$
        & $8.503\times 10^{-16}$
        & $4.300\times 10^{-1}$ \\
        & After
        & $1.11\times 10^{-5} \pm 3.0\times 10^{-6}$
        & $1.350\times 10^{-16}$
        & $\bm{2.514\times 10^{-16}}$ \\
        \bottomrule
    \end{tabular}
\end{table}

\paragraph{Marginal violations and rounding to valid transportation plans.}
The transportation plan reconstructed from an amortized prediction is obtained from a predicted Kantorovich potential together with the corresponding update of the other potential. Consequently, the resulting matrix is not necessarily guaranteed to satisfy both marginal constraints exactly before further normalization. We explicitly quantify this effect by measuring the $L_1$ marginal violations with respect to the source and target measures, denoted by $\mathrm{MargErr}_a$ and $\mathrm{MargErr}_b$, respectively.

We additionally apply the standard rounding procedure of \citep{altschuler2017near} to project each predicted matrix onto the transportation polytope with the prescribed marginals. As shown in Table~\ref{tab:marginal_rounding}, before rounding, each method satisfies one marginal up to numerical precision, while the other may exhibit a noticeable violation due to the single-potential update. After rounding, both marginal errors are reduced to approximately $10^{-16}$ for all methods, yielding valid transportation plans up to machine precision. Importantly, enforcing the marginal constraints does not compromise prediction quality: the plan RMSE decreases after rounding, and RA-OT and OA-OT continue to achieve lower reconstruction errors than Meta-OT.

\paragraph{Training-inference time trade-off.}
Although RA-OT and OA-OT require substantially less training time than Meta-OT, their per-pair inference time is higher due primarily to the computation of the sliced potentials. This additional inference cost comes with several properties of the sliced-potential representation: the resulting amortized model is independent of the number of atoms in the input measures, can naturally accommodate measures with varying support sizes, and uses only a small number of learned parameters. We further investigate the resulting trade-off between the one-time training cost and the per-pair inference cost through a break-even analysis.

In particular, we consider the MNIST setting with $M=50$ and $\epsilon=0.01$ and measure all runtimes sequentially on the GPU, without parallelizing inference across test pairs. For an amortized method, we define the total computational cost for processing $N$ test pairs as
\begin{equation}
    \mathrm{Cost}(N)
    =
    T_{\mathrm{train}}
    +
    N t_{\mathrm{infer}},
    \label{eq:break_even_cost}
\end{equation}
where $T_{\mathrm{train}}$ denotes the one-time training cost and $t_{\mathrm{infer}}$ is the average inference time per test pair. We compare this cost with solving every OT problem independently from scratch using Sinkhorn, whose average runtime is $t_{\mathrm{scratch}}=158.10$ ms per pair. We denote by $N^\star$ the number of test pairs at which the total costs of two methods become equal.

Table~\ref{tab:break_even_scratch} shows that RA-OT, OA-OT, and Meta-OT break even with solving OT from scratch after $34$, $137$, and $238$ test pairs, respectively. Comparing amortized methods directly, Table~\ref{tab:break_even_amortized} shows that RA-OT and OA-OT remain cheaper than Meta-OT up to approximately $1{,}013$ and $599$ test pairs, after which Meta-OT benefits from its lower per-pair inference cost.
\begin{table}[htbp]
    \centering
    \caption{Training time, per-pair inference time, and break-even point relative to independently solving each OT problem using Sinkhorn. Results are measured on MNIST with $M=50$ and $\epsilon=0.01$. The timing is sequential and does not exploit parallelization across test pairs.}
    \label{tab:break_even_scratch}
    \begin{tabular}{lccc}
        \toprule
        Method
        & $T_{\mathrm{train}}$ (s)
        & $t_{\mathrm{infer}}$ (ms/pair)
        & $N^\star$ vs.\ scratch \\
        \midrule
        RA-OT   & $4.24$  & $34.48$  & $34$  \\
        OA-OT   & $16.68$ & $36.23$  & $137$ \\
        Meta-OT & $37.23$ & $1.92$   & $238$ \\
        Sinkhorn from scratch & - & $158.10$ & - \\
        \bottomrule
    \end{tabular}
\end{table}
\begin{table}[htbp]
    \centering
    \caption{Break-even points between the proposed methods and Meta-OT under sequential inference. A method in the left column has a lower total computational cost below the corresponding  $N^\star$.}
    \label{tab:break_even_amortized}
    \begin{tabular}{lcl}
        \toprule
        Comparison & $N^\star$ & Interpretation \\
        \midrule
        RA-OT vs.\ Meta-OT
        & $1013.5$
        & RA-OT lower cost for $N < 1013$ \\
        OA-OT vs.\ Meta-OT
        & $599.2$
        & OA-OT lower cost for $N < 599$ \\
        \bottomrule
    \end{tabular}
\end{table}
\begin{table}[hpbt]
    \centering
    \caption{Total computational cost, in seconds, at representative numbers of test pairs $N$. The cost of each amortized method includes its one-time training cost.}
    \label{tab:total_cost_scales}
    \begin{tabular}{rcccc}
        \toprule
        $N$ & RA-OT & OA-OT & Meta-OT & Sinkhorn from scratch \\
        \midrule
        $1$     & $4.27$  & $16.71$ & $37.24$ & $0.16$   \\
        $100$   & $7.68$  & $20.30$ & $37.42$ & $15.81$  \\
        $1{,}000$ & $38.71$ & $52.90$ & $39.15$ & $158.10$ \\
        \bottomrule
    \end{tabular}
\end{table}

These results highlight a clear training--inference trade-off. Meta-OT offers faster sequential inference at large scale, whereas RA-OT and OA-OT require less training time and achieve lower plan reconstruction errors at a higher per-pair inference cost. The preferred regime therefore depends on the balance between training cost, throughput, and accuracy. \textbf{Since test pairs can be processed independently, parallel inference may further shift the practical crossover points}; we leave a detailed throughput study to future work.

\section{Additional Visualizations}
\label{ab:add_vis}

\begin{figure}[htbp]
    \centering

     \begin{minipage}{0.48\textwidth}
        \centering
        \begin{tikzpicture}
            \node[inner sep=0pt] (img3) at (0,0)
                {\includegraphics[width=0.87\linewidth]{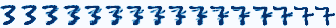}};
            \node[above=0.1cm of img3.north, font=\normalsize]
                {\textbf{Sinkhorn} \textcolor{gray}{(converged, ground-truth)}};
            \coordinate (bl3) at ([xshift=0.3cm,  yshift=-0.1cm]img3.south west);
            \coordinate (br3) at ([xshift=-0.3cm, yshift=-0.1cm]img3.south east);
            \coordinate (bc3) at ([yshift=-0.1cm]img3.south);
            \draw[<->, >=Latex, thick] (bl3) -- (br3);
            \draw[thick] (bc3) -- ([yshift=0.15cm]bc3);
            \node[left=0.05cm  of bl3] {$\alpha_0$};
            \node[below=0cm    of bc3] {$\alpha_1$};
            \node[right=0.05cm of br3] {$\alpha_2$};
        \end{tikzpicture}
    \end{minipage}\hfill
    \begin{minipage}{0.48\textwidth}
        \centering
        \begin{tikzpicture}
            \node[inner sep=0pt] (img3) at (0,0)
                {\includegraphics[width=0.87\linewidth]{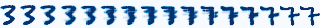}};
            \node[above=0.1cm of img3.north, font=\normalsize]
                {\textbf{Meta-OT}};
            \coordinate (bl3) at ([xshift=0.3cm,  yshift=-0.1cm]img3.south west);
            \coordinate (br3) at ([xshift=-0.3cm, yshift=-0.1cm]img3.south east);
            \coordinate (bc3) at ([yshift=-0.1cm]img3.south);
            \draw[<->, >=Latex, thick] (bl3) -- (br3);
            \draw[thick] (bc3) -- ([yshift=0.15cm]bc3);
            \node[left=0.05cm  of bl3] {$\alpha_0$};
            \node[below=0cm    of bc3] {$\alpha_1$};
            \node[right=0.05cm of br3] {$\alpha_2$};
        \end{tikzpicture}
    \end{minipage}

    \vspace{0.45cm}

    \begin{minipage}{0.48\textwidth}
        \centering
        \begin{tikzpicture}
            \node[inner sep=0pt] (img3) at (0,0)
                {\includegraphics[width=0.87\linewidth]{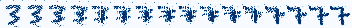}};
            \node[above=0.1cm of img3.north, font=\normalsize]
                {\textbf{Min-STP}};
            \coordinate (bl3) at ([xshift=0.3cm,  yshift=-0.1cm]img3.south west);
            \coordinate (br3) at ([xshift=-0.3cm, yshift=-0.1cm]img3.south east);
            \coordinate (bc3) at ([yshift=-0.1cm]img3.south);
            \draw[<->, >=Latex, thick] (bl3) -- (br3);
            \draw[thick] (bc3) -- ([yshift=0.15cm]bc3);
            \node[left=0.05cm  of bl3] {$\alpha_0$};
            \node[below=0cm    of bc3] {$\alpha_1$};
            \node[right=0.05cm of br3] {$\alpha_2$};
        \end{tikzpicture}
    \end{minipage}\hfill
    \begin{minipage}{0.48\textwidth}
        \centering
        \begin{tikzpicture}
            \node[inner sep=0pt] (img3) at (0,0)
                {\includegraphics[width=0.87\linewidth]{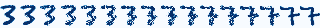}};
            \node[above=0.1cm of img3.north, font=\normalsize]
                {\textbf{min-SWGG}};
            \coordinate (bl3) at ([xshift=0.3cm,  yshift=-0.1cm]img3.south west);
            \coordinate (br3) at ([xshift=-0.3cm, yshift=-0.1cm]img3.south east);
            \coordinate (bc3) at ([yshift=-0.1cm]img3.south);
            \draw[<->, >=Latex, thick] (bl3) -- (br3);
            \draw[thick] (bc3) -- ([yshift=0.15cm]bc3);
            \node[left=0.05cm  of bl3] {$\alpha_0$};
            \node[below=0cm    of bc3] {$\alpha_1$};
            \node[right=0.05cm of br3] {$\alpha_2$};
        \end{tikzpicture}
    \end{minipage}

    \vspace{0.45cm}

    \begin{minipage}{0.48\textwidth}
        \centering
        \begin{tikzpicture}
            \node[inner sep=0pt] (img1) at (0,0)
                {\includegraphics[width=0.87\linewidth]{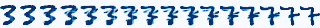}};
            \node[above=0.1cm of img1.north, font=\normalsize]
                {\textbf{RA-OT}};
            \coordinate (bl1) at ([xshift=0.3cm,  yshift=-0.1cm]img1.south west);
            \coordinate (br1) at ([xshift=-0.3cm, yshift=-0.1cm]img1.south east);
            \coordinate (bc1) at ([yshift=-0.1cm]img1.south);
            \draw[<->, >=Latex, thick] (bl1) -- (br1);
            \draw[thick] (bc1) -- ([yshift=0.15cm]bc1);
            \node[left=0.05cm  of bl1] {$\alpha_0$};
            \node[below=0cm    of bc1] {$\alpha_1$};
            \node[right=0.05cm of br1] {$\alpha_2$};
        \end{tikzpicture}
    \end{minipage}\hfill
    \begin{minipage}{0.48\textwidth}
        \centering
        \begin{tikzpicture}
            \node[inner sep=0pt] (img2) at (0,0)
                {\includegraphics[width=0.87\linewidth]{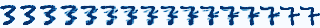}};
            \node[above=0.1cm of img2.north, font=\normalsize]
                {\textbf{OA-OT}};
            \coordinate (bl2) at ([xshift=0.3cm,  yshift=-0.1cm]img2.south west);
            \coordinate (br2) at ([xshift=-0.3cm, yshift=-0.1cm]img2.south east);
            \coordinate (bc2) at ([yshift=-0.1cm]img2.south);
            \draw[<->, >=Latex, thick] (bl2) -- (br2);
            \draw[thick] (bc2) -- ([yshift=0.15cm]bc2);
            \node[left=0.05cm  of bl2] {$\alpha_0$};
            \node[below=0cm    of bc2] {$\alpha_1$};
            \node[right=0.05cm of br2] {$\alpha_2$};
        \end{tikzpicture}
    \end{minipage}

    \vspace{0.45cm}

    \begin{minipage}{0.48\textwidth}
        \centering
        \begin{tikzpicture}
            \node[inner sep=0pt] (img3) at (0,0)
                {\includegraphics[width=0.87\linewidth]{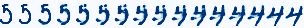}};
            \node[above=0.1cm of img3.north, font=\normalsize]
                {\textbf{Sinkhorn} \textcolor{gray}{(converged, ground-truth)}};
            \coordinate (bl3) at ([xshift=0.3cm,  yshift=-0.1cm]img3.south west);
            \coordinate (br3) at ([xshift=-0.3cm, yshift=-0.1cm]img3.south east);
            \coordinate (bc3) at ([yshift=-0.1cm]img3.south);
            \draw[<->, >=Latex, thick] (bl3) -- (br3);
            \draw[thick] (bc3) -- ([yshift=0.15cm]bc3);
            \node[left=0.05cm  of bl3] {$\alpha_0$};
            \node[below=0cm    of bc3] {$\alpha_1$};
            \node[right=0.05cm of br3] {$\alpha_2$};
        \end{tikzpicture}
    \end{minipage}\hfill
    \begin{minipage}{0.48\textwidth}
        \centering
        \begin{tikzpicture}
            \node[inner sep=0pt] (img3) at (0,0)
                {\includegraphics[width=0.87\linewidth]{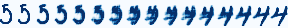}};
            \node[above=0.1cm of img3.north, font=\normalsize]
                {\textbf{Meta-OT}};
            \coordinate (bl3) at ([xshift=0.3cm,  yshift=-0.1cm]img3.south west);
            \coordinate (br3) at ([xshift=-0.3cm, yshift=-0.1cm]img3.south east);
            \coordinate (bc3) at ([yshift=-0.1cm]img3.south);
            \draw[<->, >=Latex, thick] (bl3) -- (br3);
            \draw[thick] (bc3) -- ([yshift=0.15cm]bc3);
            \node[left=0.05cm  of bl3] {$\alpha_0$};
            \node[below=0cm    of bc3] {$\alpha_1$};
            \node[right=0.05cm of br3] {$\alpha_2$};
        \end{tikzpicture}
    \end{minipage}

    \vspace{0.45cm}

    \begin{minipage}{0.48\textwidth}
        \centering
        \begin{tikzpicture}
            \node[inner sep=0pt] (img3) at (0,0)
                {\includegraphics[width=0.87\linewidth]{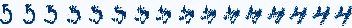}};
            \node[above=0.1cm of img3.north, font=\normalsize]
                {\textbf{Min-STP}};
            \coordinate (bl3) at ([xshift=0.3cm,  yshift=-0.1cm]img3.south west);
            \coordinate (br3) at ([xshift=-0.3cm, yshift=-0.1cm]img3.south east);
            \coordinate (bc3) at ([yshift=-0.1cm]img3.south);
            \draw[<->, >=Latex, thick] (bl3) -- (br3);
            \draw[thick] (bc3) -- ([yshift=0.15cm]bc3);
            \node[left=0.05cm  of bl3] {$\alpha_0$};
            \node[below=0cm    of bc3] {$\alpha_1$};
            \node[right=0.05cm of br3] {$\alpha_2$};
        \end{tikzpicture}
    \end{minipage}\hfill
    \begin{minipage}{0.48\textwidth}
        \centering
        \begin{tikzpicture}
            \node[inner sep=0pt] (img3) at (0,0)
                {\includegraphics[width=0.87\linewidth]{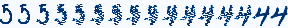}};
            \node[above=0.1cm of img3.north, font=\normalsize]
                {\textbf{min-SWGG}};
            \coordinate (bl3) at ([xshift=0.3cm,  yshift=-0.1cm]img3.south west);
            \coordinate (br3) at ([xshift=-0.3cm, yshift=-0.1cm]img3.south east);
            \coordinate (bc3) at ([yshift=-0.1cm]img3.south);
            \draw[<->, >=Latex, thick] (bl3) -- (br3);
            \draw[thick] (bc3) -- ([yshift=0.15cm]bc3);
            \node[left=0.05cm  of bl3] {$\alpha_0$};
            \node[below=0cm    of bc3] {$\alpha_1$};
            \node[right=0.05cm of br3] {$\alpha_2$};
        \end{tikzpicture}
    \end{minipage}

    \vspace{0.45cm}

    \begin{minipage}{0.48\textwidth}
        \centering
        \begin{tikzpicture}
            \node[inner sep=0pt] (img1) at (0,0)
                {\includegraphics[width=0.87\linewidth]{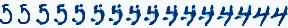}};
            \node[above=0.1cm of img1.north, font=\normalsize]
                {\textbf{RA-OT}};
            \coordinate (bl1) at ([xshift=0.3cm,  yshift=-0.1cm]img1.south west);
            \coordinate (br1) at ([xshift=-0.3cm, yshift=-0.1cm]img1.south east);
            \coordinate (bc1) at ([yshift=-0.1cm]img1.south);
            \draw[<->, >=Latex, thick] (bl1) -- (br1);
            \draw[thick] (bc1) -- ([yshift=0.15cm]bc1);
            \node[left=0.05cm  of bl1] {$\alpha_0$};
            \node[below=0cm    of bc1] {$\alpha_1$};
            \node[right=0.05cm of br1] {$\alpha_2$};
        \end{tikzpicture}
    \end{minipage}\hfill
    \begin{minipage}{0.48\textwidth}
        \centering
        \begin{tikzpicture}
            \node[inner sep=0pt] (img2) at (0,0)
                {\includegraphics[width=0.87\linewidth]{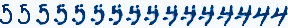}};
            \node[above=0.1cm of img2.north, font=\normalsize]
                {\textbf{OA-OT}};
            \coordinate (bl2) at ([xshift=0.3cm,  yshift=-0.1cm]img2.south west);
            \coordinate (br2) at ([xshift=-0.3cm, yshift=-0.1cm]img2.south east);
            \coordinate (bc2) at ([yshift=-0.1cm]img2.south);
            \draw[<->, >=Latex, thick] (bl2) -- (br2);
            \draw[thick] (bc2) -- ([yshift=0.15cm]bc2);
            \node[left=0.05cm  of bl2] {$\alpha_0$};
            \node[below=0cm    of bc2] {$\alpha_1$};
            \node[right=0.05cm of br2] {$\alpha_2$};
        \end{tikzpicture}
    \end{minipage}

    \vspace{0.45cm}

    \begin{minipage}{0.48\textwidth}
        \centering
        \begin{tikzpicture}
            \node[inner sep=0pt] (img3) at (0,0)
                {\includegraphics[width=0.87\linewidth]{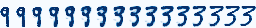}};
            \node[above=0.1cm of img3.north, font=\normalsize]
                {\textbf{Sinkhorn} \textcolor{gray}{(converged, ground-truth)}};
            \coordinate (bl3) at ([xshift=0.3cm,  yshift=-0.1cm]img3.south west);
            \coordinate (br3) at ([xshift=-0.3cm, yshift=-0.1cm]img3.south east);
            \coordinate (bc3) at ([yshift=-0.1cm]img3.south);
            \draw[<->, >=Latex, thick] (bl3) -- (br3);
            \draw[thick] (bc3) -- ([yshift=0.15cm]bc3);
            \node[left=0.05cm  of bl3] {$\alpha_0$};
            \node[below=0cm    of bc3] {$\alpha_1$};
            \node[right=0.05cm of br3] {$\alpha_2$};
        \end{tikzpicture}
    \end{minipage}\hfill
    \begin{minipage}{0.48\textwidth}
        \centering
        \begin{tikzpicture}
            \node[inner sep=0pt] (img3) at (0,0)
                {\includegraphics[width=0.87\linewidth]{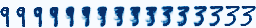}};
            \node[above=0.1cm of img3.north, font=\normalsize]
                {\textbf{Meta-OT}};
            \coordinate (bl3) at ([xshift=0.3cm,  yshift=-0.1cm]img3.south west);
            \coordinate (br3) at ([xshift=-0.3cm, yshift=-0.1cm]img3.south east);
            \coordinate (bc3) at ([yshift=-0.1cm]img3.south);
            \draw[<->, >=Latex, thick] (bl3) -- (br3);
            \draw[thick] (bc3) -- ([yshift=0.15cm]bc3);
            \node[left=0.05cm  of bl3] {$\alpha_0$};
            \node[below=0cm    of bc3] {$\alpha_1$};
            \node[right=0.05cm of br3] {$\alpha_2$};
        \end{tikzpicture}
    \end{minipage}

    \vspace{0.45cm}

    \begin{minipage}{0.48\textwidth}
        \centering
        \begin{tikzpicture}
            \node[inner sep=0pt] (img3) at (0,0)
                {\includegraphics[width=0.87\linewidth]{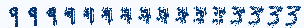}};
            \node[above=0.1cm of img3.north, font=\normalsize]
                {\textbf{Min-STP}};
            \coordinate (bl3) at ([xshift=0.3cm,  yshift=-0.1cm]img3.south west);
            \coordinate (br3) at ([xshift=-0.3cm, yshift=-0.1cm]img3.south east);
            \coordinate (bc3) at ([yshift=-0.1cm]img3.south);
            \draw[<->, >=Latex, thick] (bl3) -- (br3);
            \draw[thick] (bc3) -- ([yshift=0.15cm]bc3);
            \node[left=0.05cm  of bl3] {$\alpha_0$};
            \node[below=0cm    of bc3] {$\alpha_1$};
            \node[right=0.05cm of br3] {$\alpha_2$};
        \end{tikzpicture}
    \end{minipage}\hfill
    \begin{minipage}{0.48\textwidth}
        \centering
        \begin{tikzpicture}
            \node[inner sep=0pt] (img3) at (0,0)
                {\includegraphics[width=0.87\linewidth]{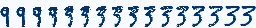}};
            \node[above=0.1cm of img3.north, font=\normalsize]
                {\textbf{min-SWGG}};
            \coordinate (bl3) at ([xshift=0.3cm,  yshift=-0.1cm]img3.south west);
            \coordinate (br3) at ([xshift=-0.3cm, yshift=-0.1cm]img3.south east);
            \coordinate (bc3) at ([yshift=-0.1cm]img3.south);
            \draw[<->, >=Latex, thick] (bl3) -- (br3);
            \draw[thick] (bc3) -- ([yshift=0.15cm]bc3);
            \node[left=0.05cm  of bl3] {$\alpha_0$};
            \node[below=0cm    of bc3] {$\alpha_1$};
            \node[right=0.05cm of br3] {$\alpha_2$};
        \end{tikzpicture}
    \end{minipage}

    \vspace{0.45cm}

    \begin{minipage}{0.48\textwidth}
        \centering
        \begin{tikzpicture}
            \node[inner sep=0pt] (img1) at (0,0)
                {\includegraphics[width=0.87\linewidth]{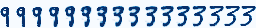}};
            \node[above=0.1cm of img1.north, font=\normalsize]
                {\textbf{RA-OT}};
            \coordinate (bl1) at ([xshift=0.3cm,  yshift=-0.1cm]img1.south west);
            \coordinate (br1) at ([xshift=-0.3cm, yshift=-0.1cm]img1.south east);
            \coordinate (bc1) at ([yshift=-0.1cm]img1.south);
            \draw[<->, >=Latex, thick] (bl1) -- (br1);
            \draw[thick] (bc1) -- ([yshift=0.15cm]bc1);
            \node[left=0.05cm  of bl1] {$\alpha_0$};
            \node[below=0cm    of bc1] {$\alpha_1$};
            \node[right=0.05cm of br1] {$\alpha_2$};
        \end{tikzpicture}
    \end{minipage}\hfill
    \begin{minipage}{0.48\textwidth}
        \centering
        \begin{tikzpicture}
            \node[inner sep=0pt] (img2) at (0,0)
                {\includegraphics[width=0.87\linewidth]{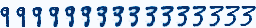}};
            \node[above=0.1cm of img2.north, font=\normalsize]
                {\textbf{OA-OT}};
            \coordinate (bl2) at ([xshift=0.3cm,  yshift=-0.1cm]img2.south west);
            \coordinate (br2) at ([xshift=-0.3cm, yshift=-0.1cm]img2.south east);
            \coordinate (bc2) at ([yshift=-0.1cm]img2.south);
            \draw[<->, >=Latex, thick] (bl2) -- (br2);
            \draw[thick] (bc2) -- ([yshift=0.15cm]bc2);
            \node[left=0.05cm  of bl2] {$\alpha_0$};
            \node[below=0cm    of bc2] {$\alpha_1$};
            \node[right=0.05cm of br2] {$\alpha_2$};
        \end{tikzpicture}
    \end{minipage}

    \vspace{0.2cm}
    \caption{ Additional qualitative visualizations of the predicted optimal transport plans on the MNIST dataset ($M=50, L=100$). The figure compares the structural alignment and digit matching quality across different amortized and sliced OT solvers (RA-OT, OA-OT, Meta-OT, Min-STP, and min-SWGG). Our proposed methods consistently preserve the topological structure of the digits.}

    \label{fig:mnist_interp}
\end{figure}
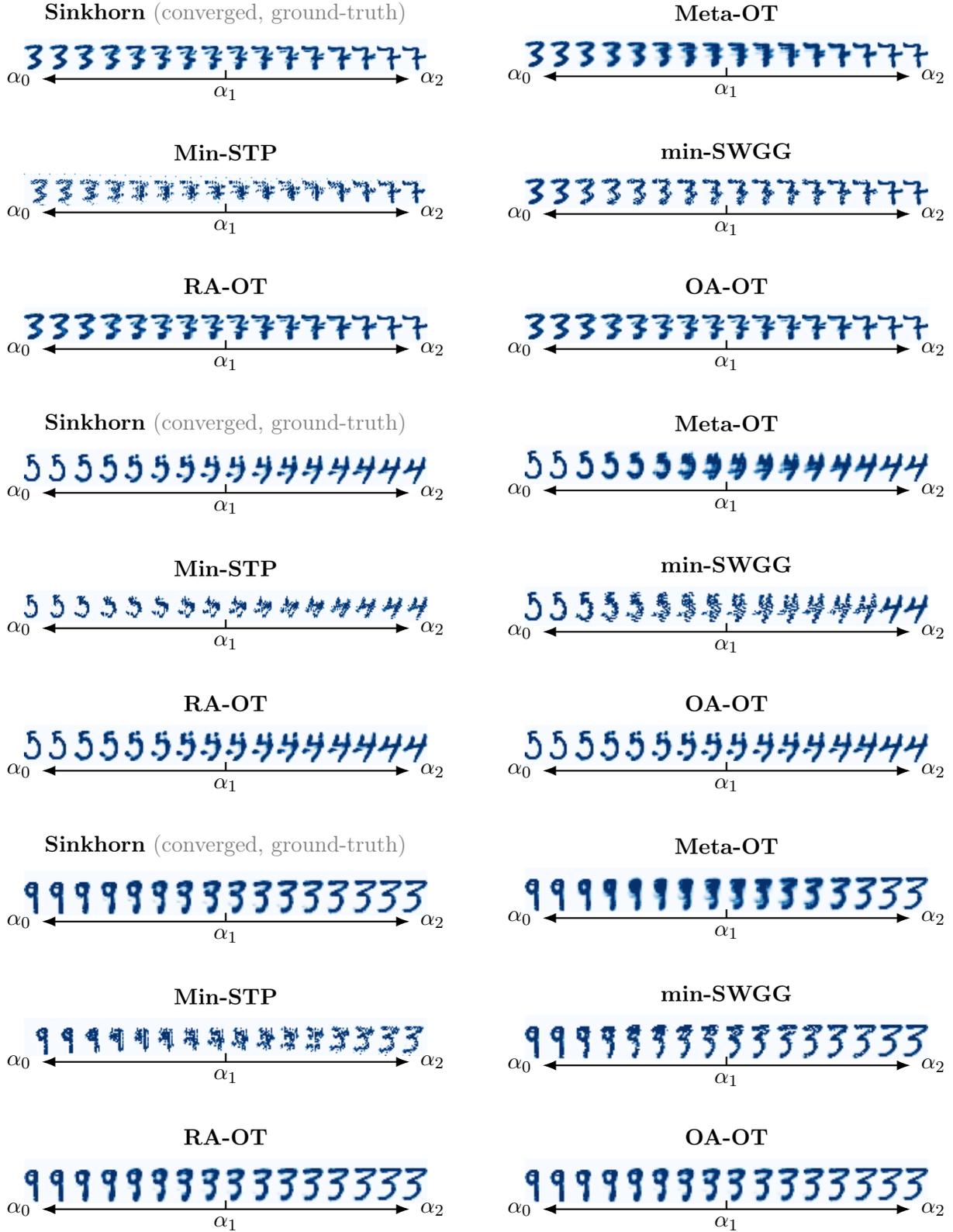

\begin{figure}[htbp]
    \centering

    \begin{minipage}{0.5\textwidth}
        \centering
        \textbf{Sinkhorn} \textcolor{gray}{(converged, ground-truth)} \\[1ex]
        \includegraphics[width=\textwidth]{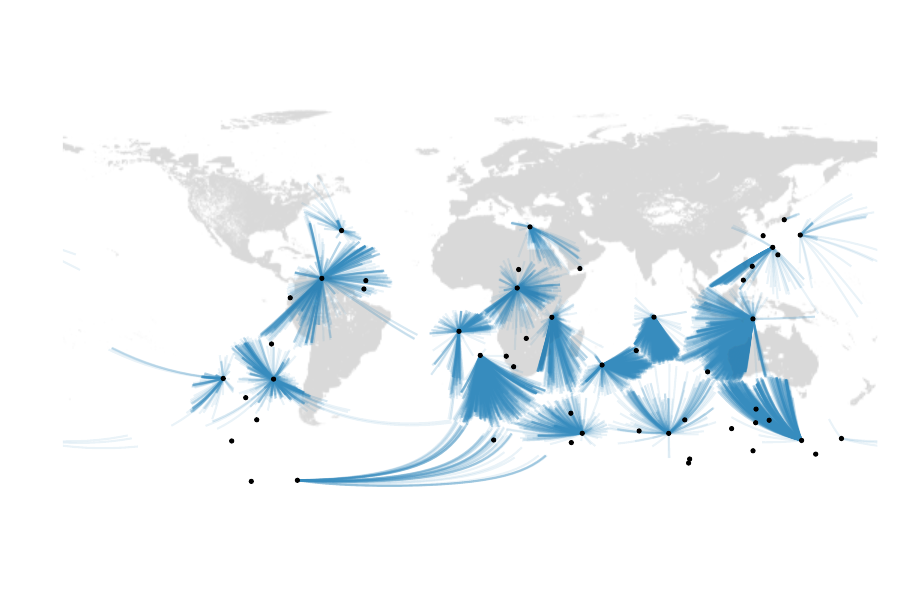}
    \end{minipage}\hfill
    \begin{minipage}{0.5\textwidth}
        \centering
        \textbf{Meta-OT} \\[1ex]
        \includegraphics[width=\textwidth]{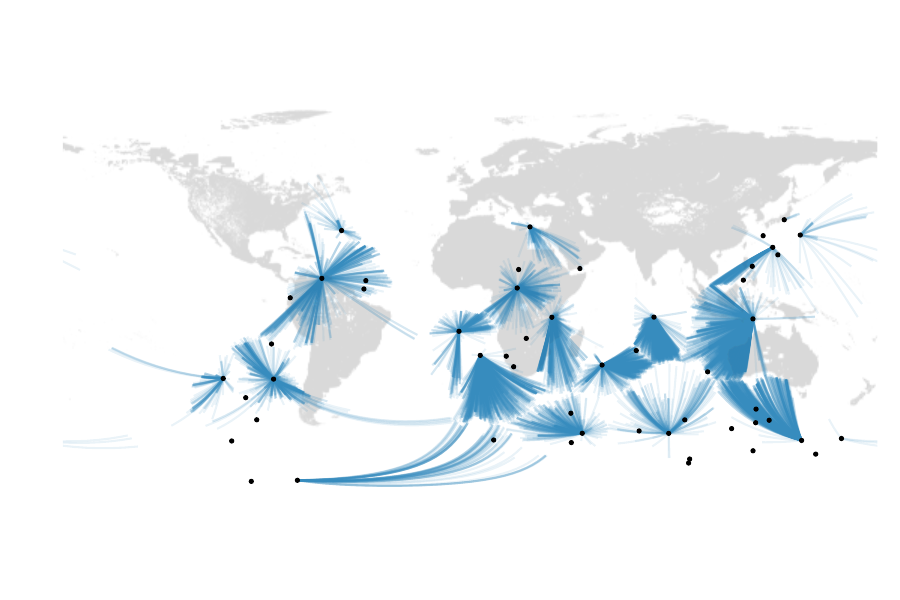}
    \end{minipage}

        \vspace{0.1cm}

    \begin{minipage}{0.5\textwidth}
        \centering
        \textbf{Min-STP} \\[1ex]
        \includegraphics[width=\textwidth]{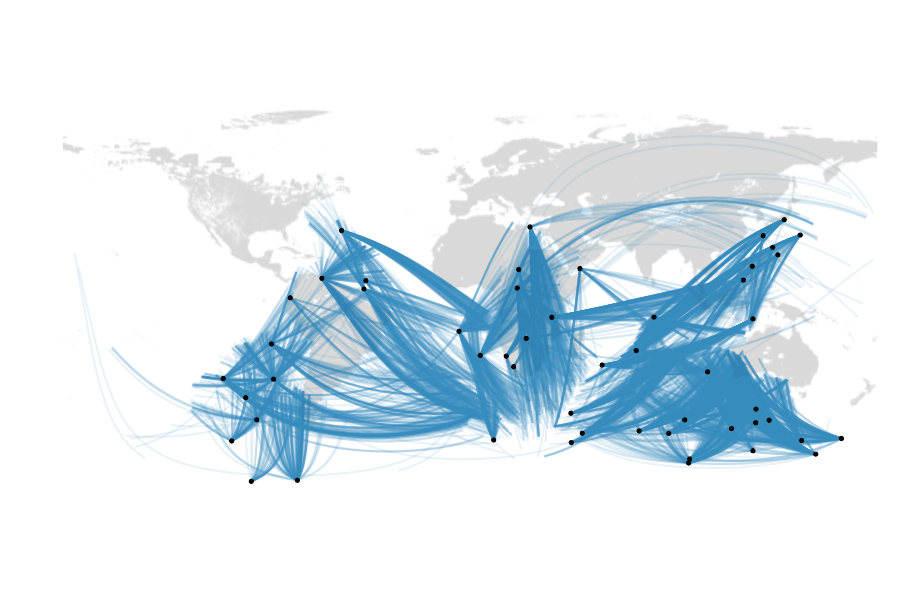}
    \end{minipage}\hfill
    \begin{minipage}{0.5\textwidth}
        \centering
        \textbf{min-SWGG} \\[1ex]
        \includegraphics[width=\textwidth]{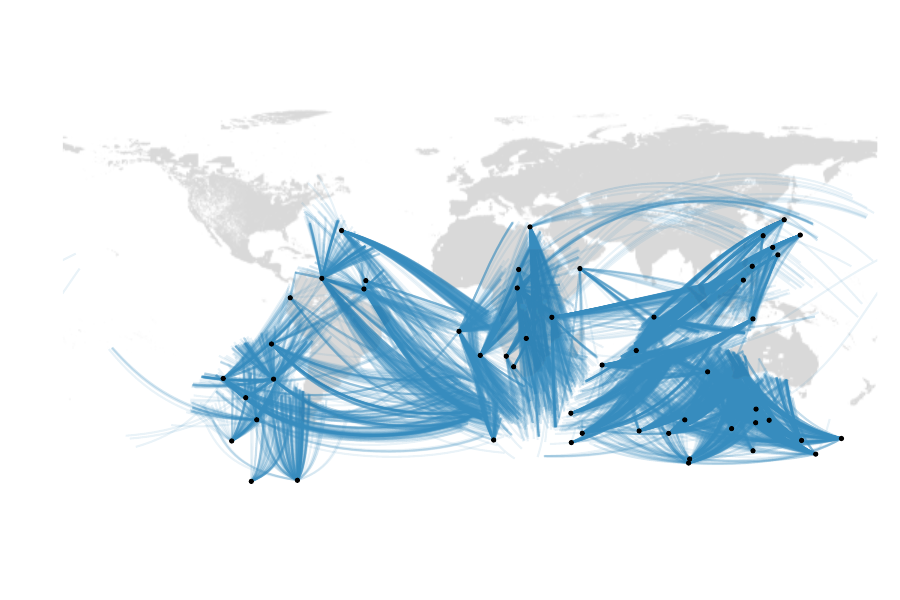}
    \end{minipage}

    \vspace{0.1cm}

    \begin{minipage}{0.5\textwidth}
        \centering
        \textbf{RA-OT} \\[1ex]
        \includegraphics[width=\linewidth]{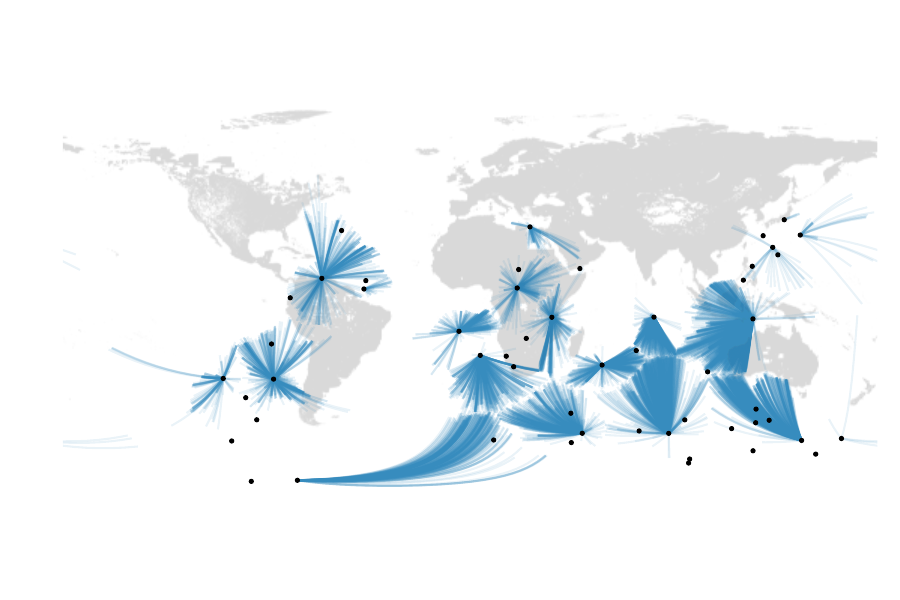}
    \end{minipage}\hfill
    \begin{minipage}{0.5\textwidth}
        \centering
        \textbf{OA-OT} \\[1ex]
        \includegraphics[width=\linewidth]{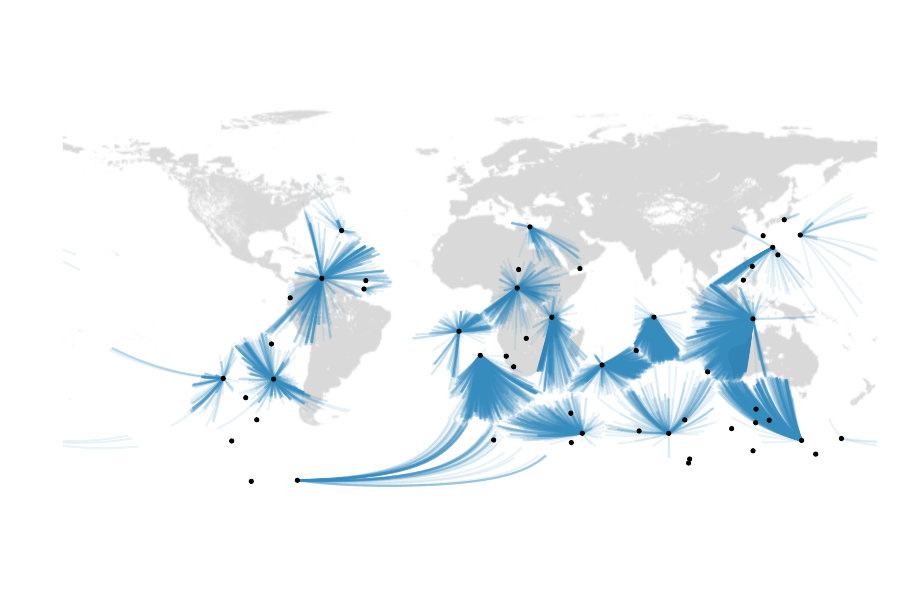}
    \end{minipage}

    \vspace{0.2cm}
    \caption{Extended qualitative results on the spherical world supply-demand transport task ($M=50, L=100$). The visualizations illustrate the transport assignments from dense global population distributions (demand) to uniformly scattered supply points across the globe. Observations indicate that RA-OT and OA-OT generate smoother and more globally coherent matching patterns, effectively avoiding the localized structural artifacts often observed in standard linear sliced estimation methods.}
    \label{fig:world_transport}
\end{figure}

\begin{figure}[htbp]
    \centering
    \setlength{\tabcolsep}{2pt} 
    \renewcommand{\arraystretch}{0}
    \newcommand{\imw}{0.24\textwidth} 

    \begin{tabular}{@{}cccc@{}}
        \small Source & \small Target & \small Sinkhorn \textcolor{gray}{(GT)} & \small Meta-OT \\[2pt]
        
        \includegraphics[width=\imw, height=\imw, keepaspectratio=false]{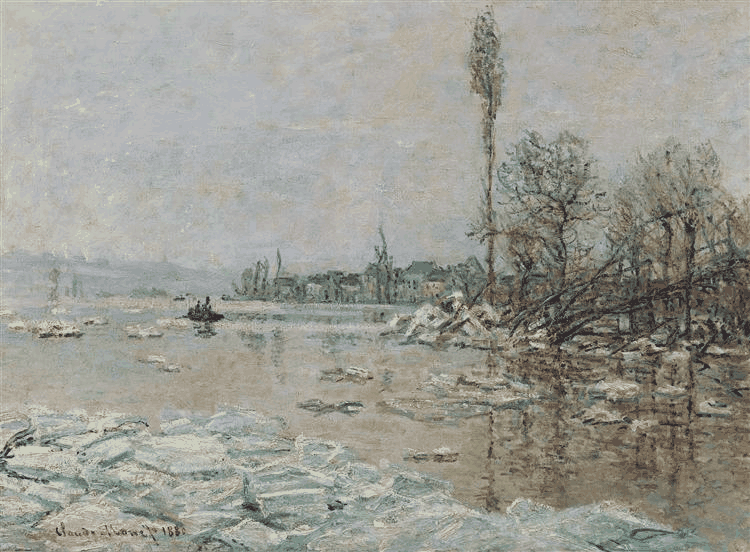} &
        \includegraphics[width=\imw, height=\imw, keepaspectratio=false]{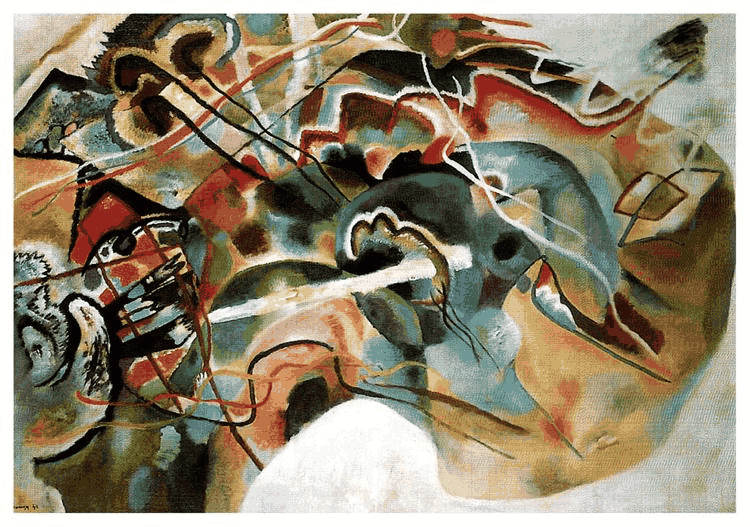} &
        \includegraphics[width=\imw, height=\imw, keepaspectratio=false]{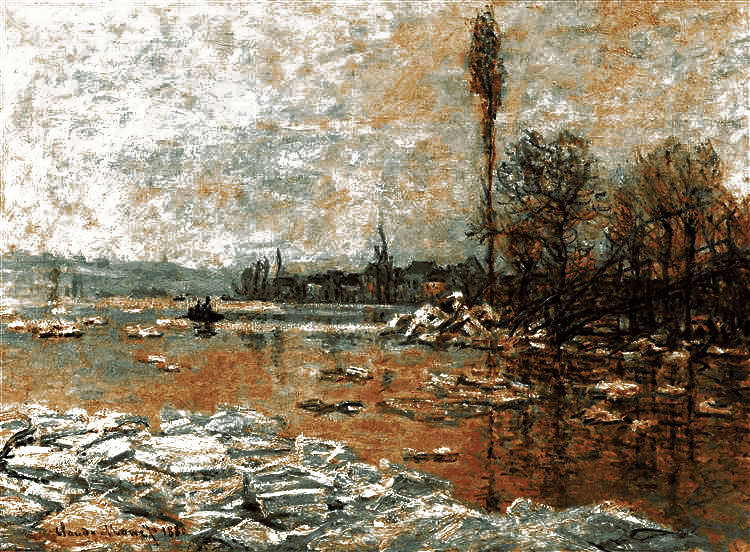} &
        \includegraphics[width=\imw, height=\imw, keepaspectratio=false]{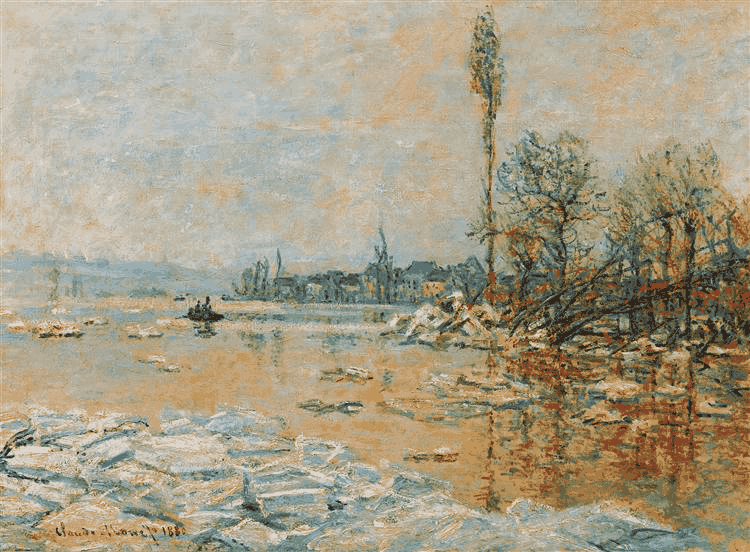} \\[6pt] 
        
        \small Min-STP & \small min-SWGG & \small \textbf{RA-OT} (ours) & \small \textbf{OA-OT} (ours) \\[2pt]
        
        \includegraphics[width=\imw, height=\imw, keepaspectratio=false]{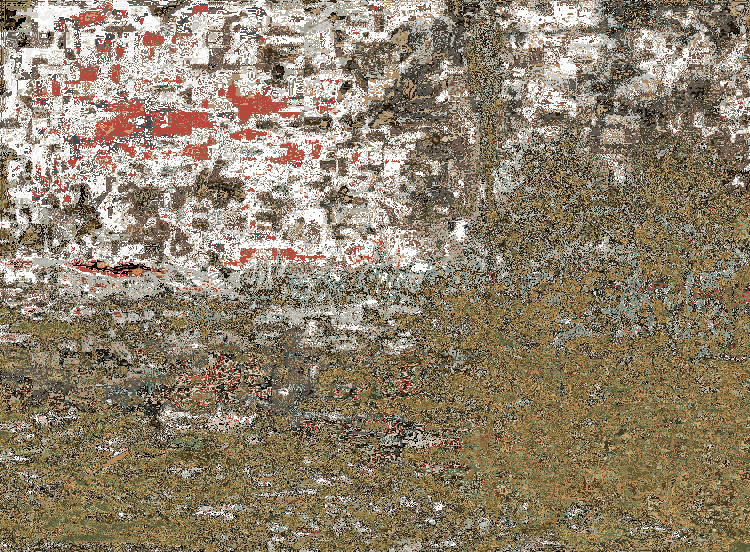} &
        \includegraphics[width=\imw, height=\imw, keepaspectratio=false]{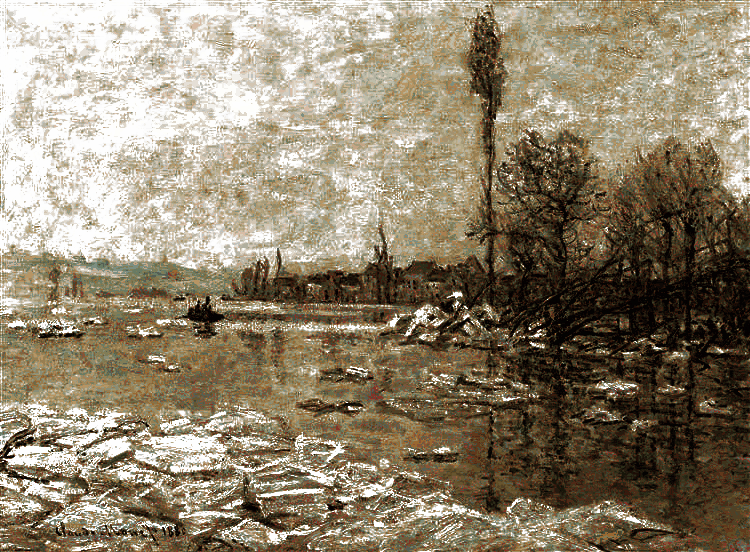} &
        \includegraphics[width=\imw, height=\imw, keepaspectratio=false]{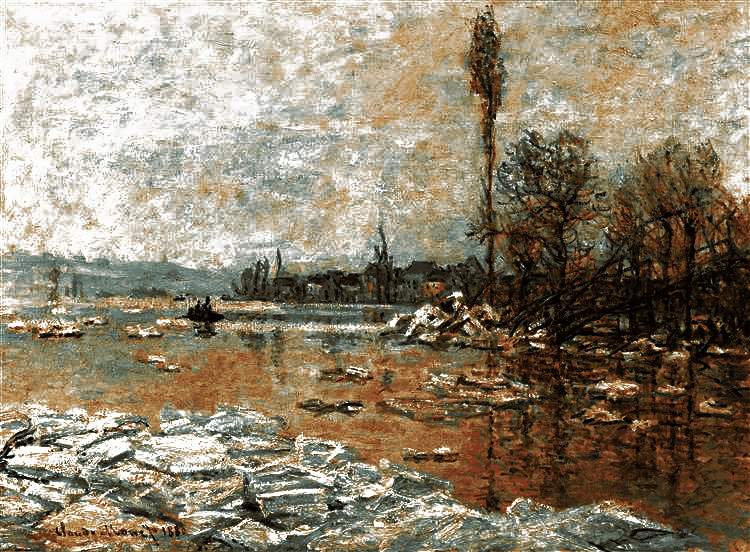} &
        \includegraphics[width=\imw, height=\imw, keepaspectratio=false]{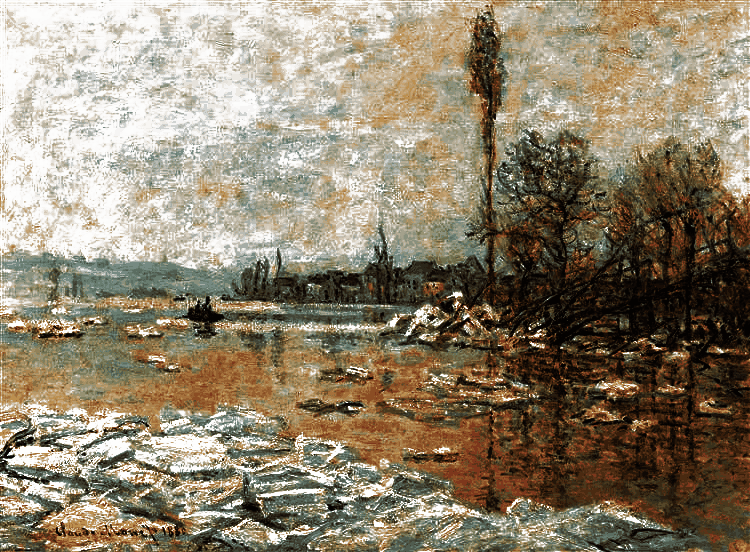}\\[6pt]

        \small Source & \small Target & \small Sinkhorn \textcolor{gray}{(GT)} & \small Meta-OT \\[2pt]
        
        \includegraphics[width=\imw, height=\imw, keepaspectratio=false]{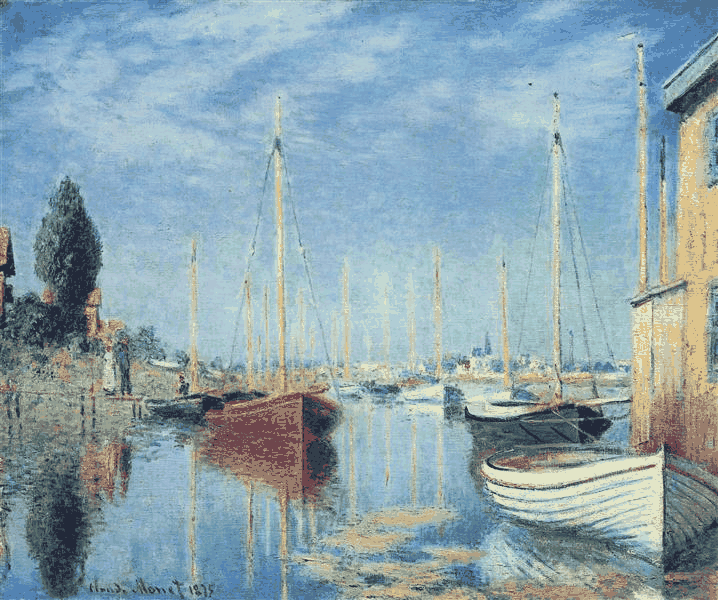} &
        \includegraphics[width=\imw, height=\imw, keepaspectratio=false]{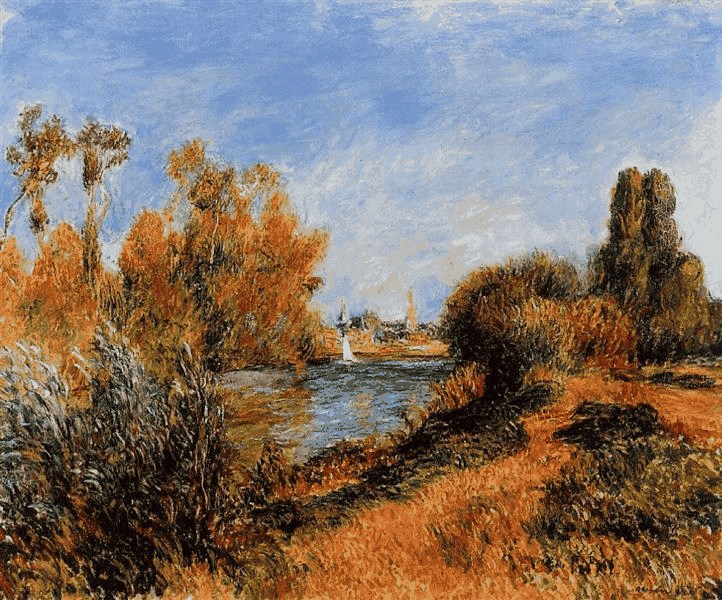} &
        \includegraphics[width=\imw, height=\imw, keepaspectratio=false]{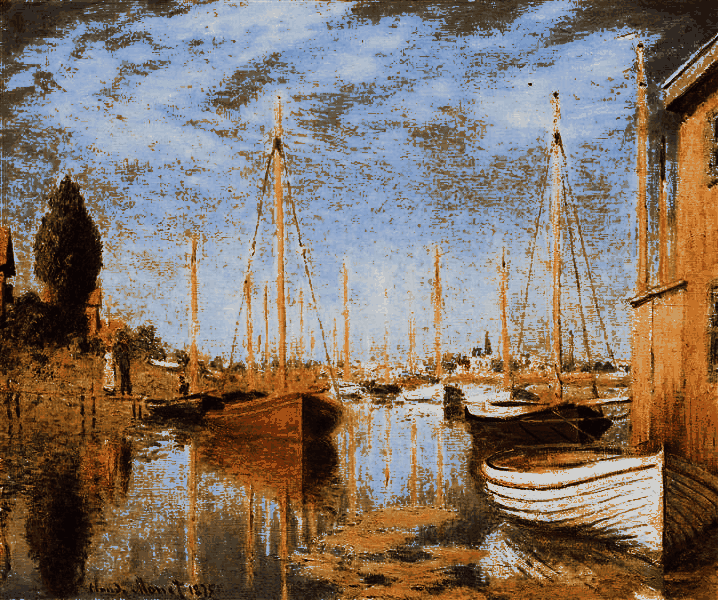} &
        \includegraphics[width=\imw, height=\imw, keepaspectratio=false]{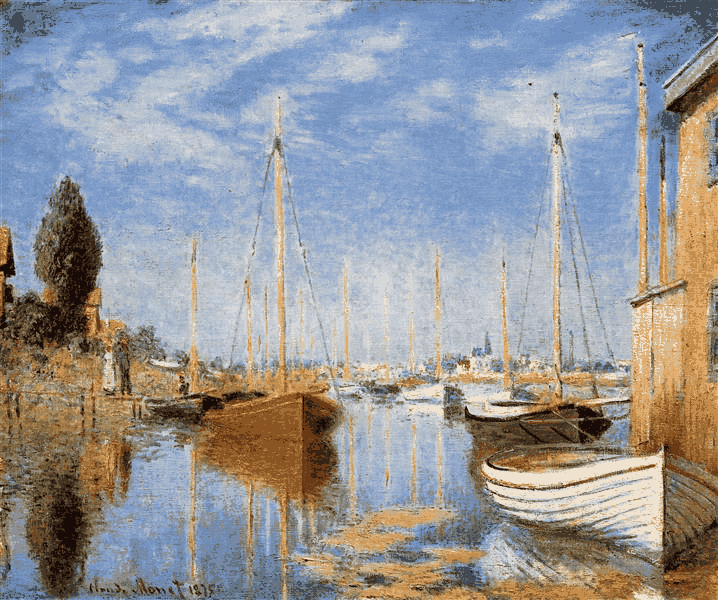} \\[6pt]
        
        \small Min-STP & \small min-SWGG & \small \textbf{RA-OT} (ours) & \small \textbf{OA-OT} (ours) \\[2pt]
        
        \includegraphics[width=\imw, height=\imw, keepaspectratio=false]{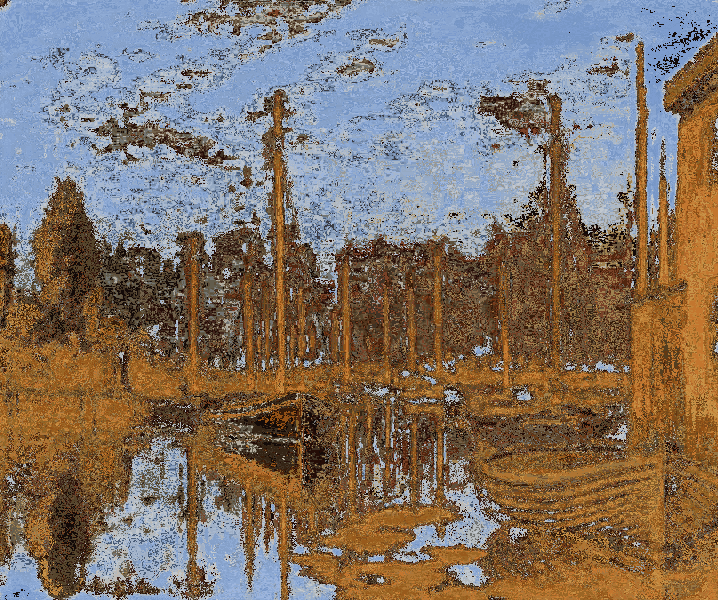} &
        \includegraphics[width=\imw, height=\imw, keepaspectratio=false]{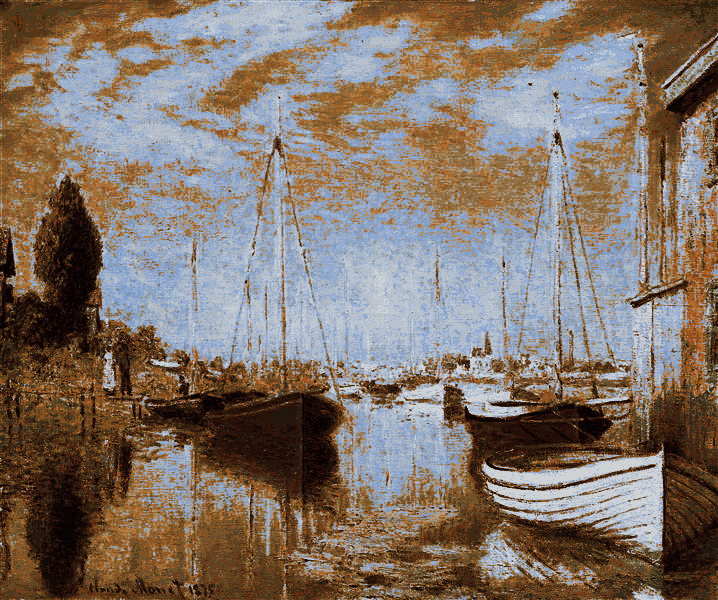} &
        \includegraphics[width=\imw, height=\imw, keepaspectratio=false]{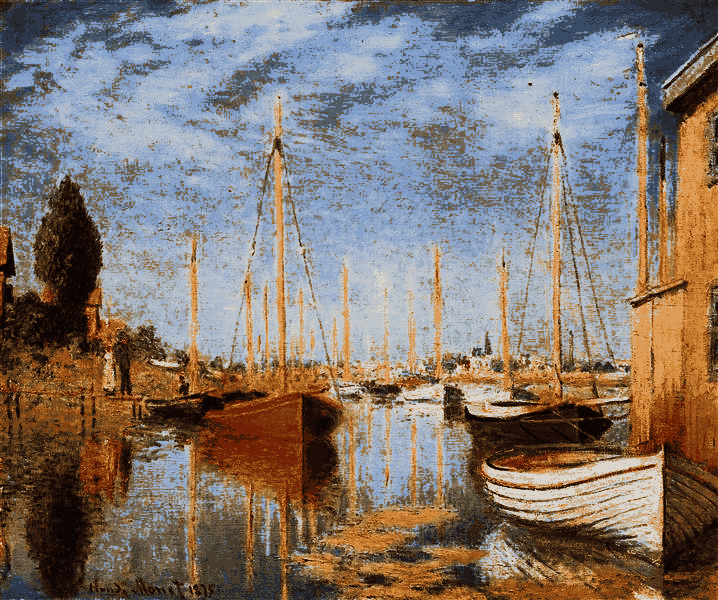} &
        \includegraphics[width=\imw, height=\imw, keepaspectratio=false]{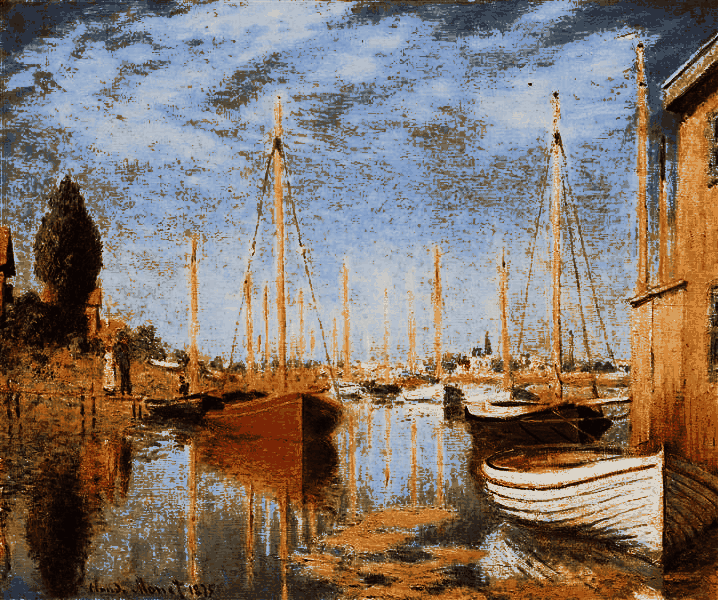} \\[6pt]

    \end{tabular}

    \vspace{0.2cm}
    \caption{Further qualitative examples of the color transfer task between natural images using different OT solvers ($M=50, L=100$). The results demonstrate the capability of our amortized solvers to robustly align high-dimensional RGB color histograms. The images transported by RA-OT and OA-OT strictly adhere to the target color palettes while faithfully preserving the underlying textural and geometric details of the source images, exhibiting highly competitive visual fidelity against established exact and amortized baselines.}
    \label{fig:color_transfer_all}
\end{figure}

\begin{figure}[htbp]
    \centering
    \setlength{\tabcolsep}{2pt} 
    \renewcommand{\arraystretch}{0}
    \newcommand{\imw}{0.24\textwidth} 

    \begin{tabular}{@{}cccc@{}}
        \small Source & \small Target & \small Sinkhorn \textcolor{gray}{(GT)} & \small Meta-OT \\[2pt]
        
        \includegraphics[width=\imw, height=\imw, keepaspectratio=false]{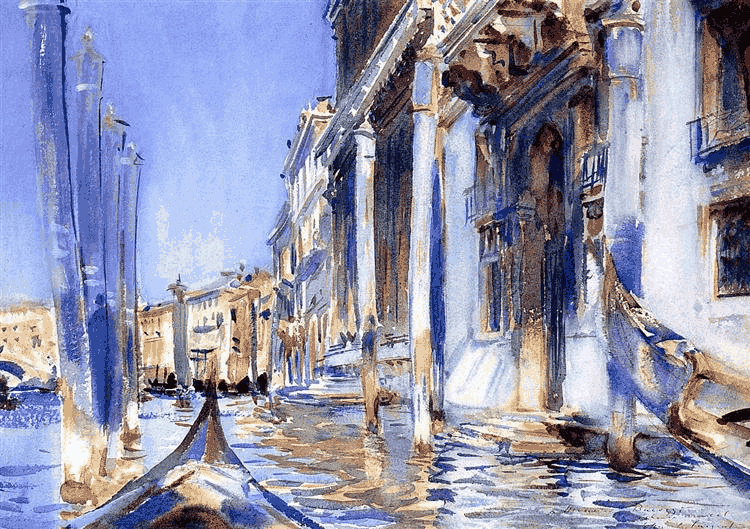} &
        \includegraphics[width=\imw, height=\imw, keepaspectratio=false]{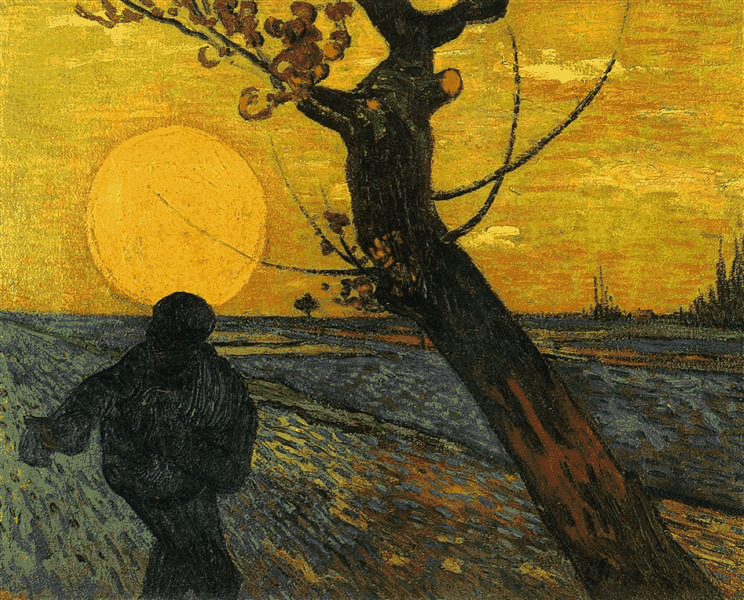} &
        \includegraphics[width=\imw, height=\imw, keepaspectratio=false]{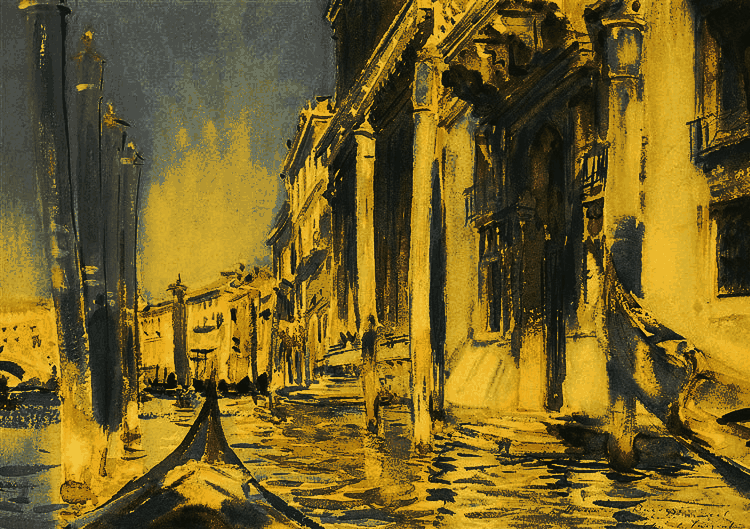} &
        \includegraphics[width=\imw, height=\imw, keepaspectratio=false]{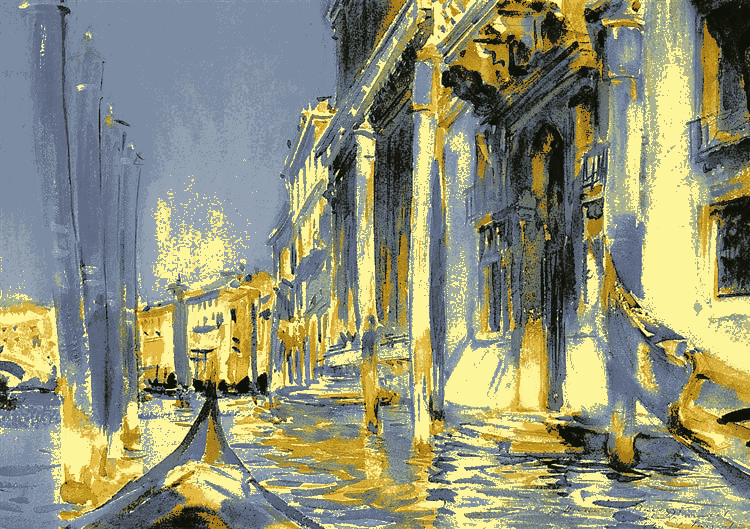} \\[6pt] 
        
        \small Min-STP & \small min-SWGG & \small \textbf{RA-OT} (ours) & \small \textbf{OA-OT} (ours) \\[2pt]
        
        \includegraphics[width=\imw, height=\imw, keepaspectratio=false]{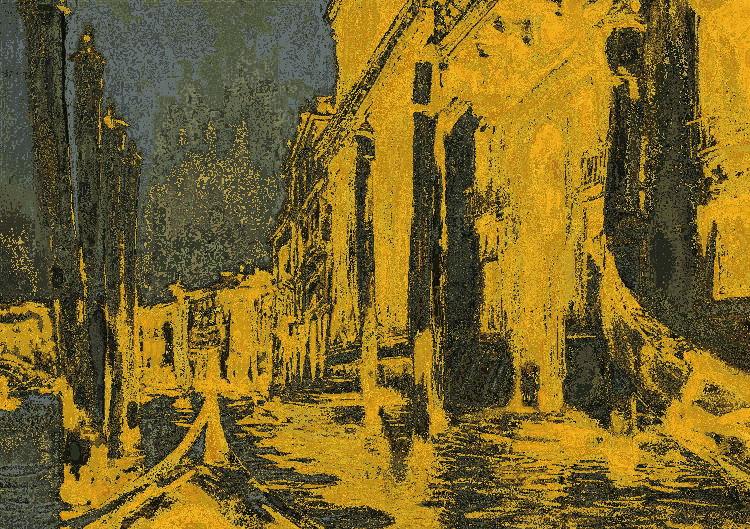} &
        \includegraphics[width=\imw, height=\imw, keepaspectratio=false]{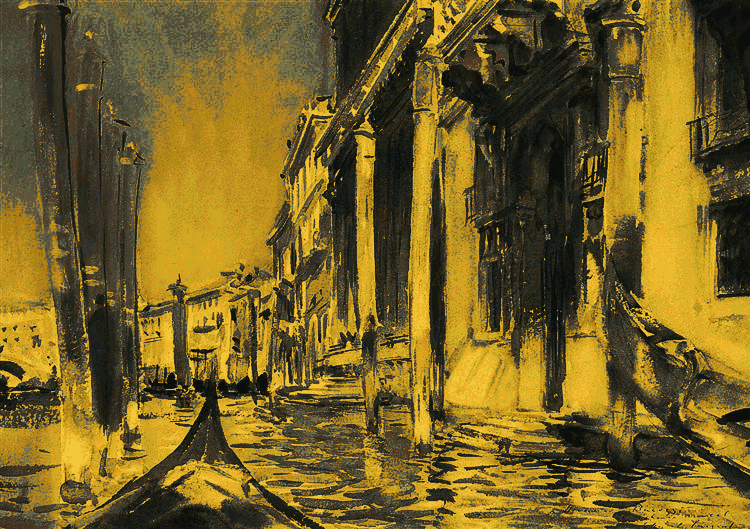} &
        \includegraphics[width=\imw, height=\imw, keepaspectratio=false]{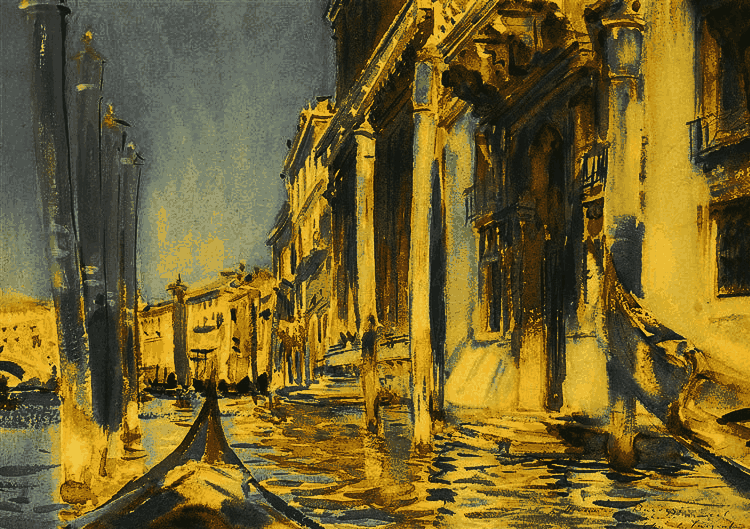} &
        \includegraphics[width=\imw, height=\imw, keepaspectratio=false]{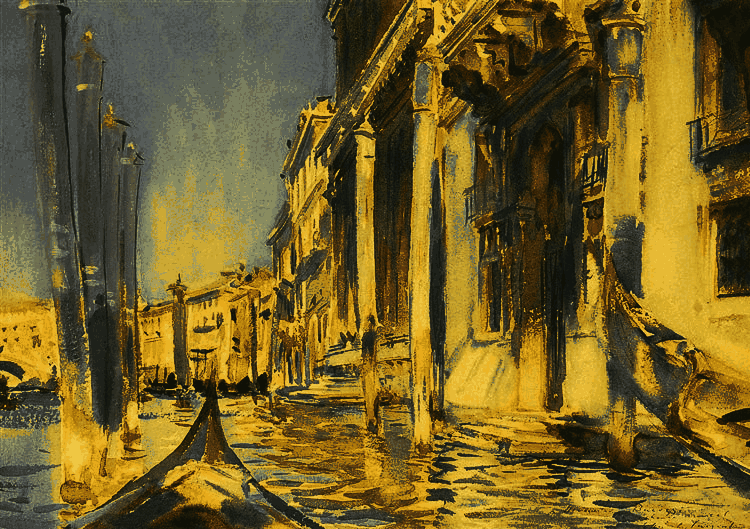}\\[6pt]

        \small Source & \small Target & \small Sinkhorn \textcolor{gray}{(GT)} & \small Meta-OT \\[2pt]
        
        \includegraphics[width=\imw, height=\imw, keepaspectratio=false]{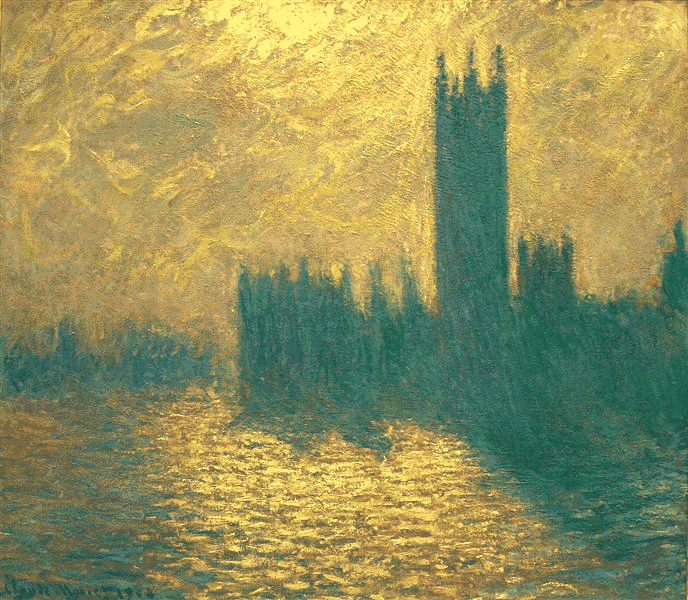} &
        \includegraphics[width=\imw, height=\imw, keepaspectratio=false]{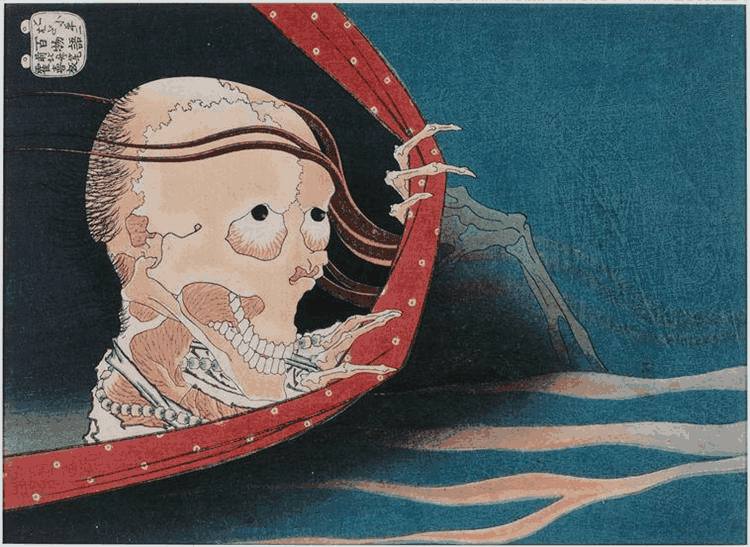} &
        \includegraphics[width=\imw, height=\imw, keepaspectratio=false]{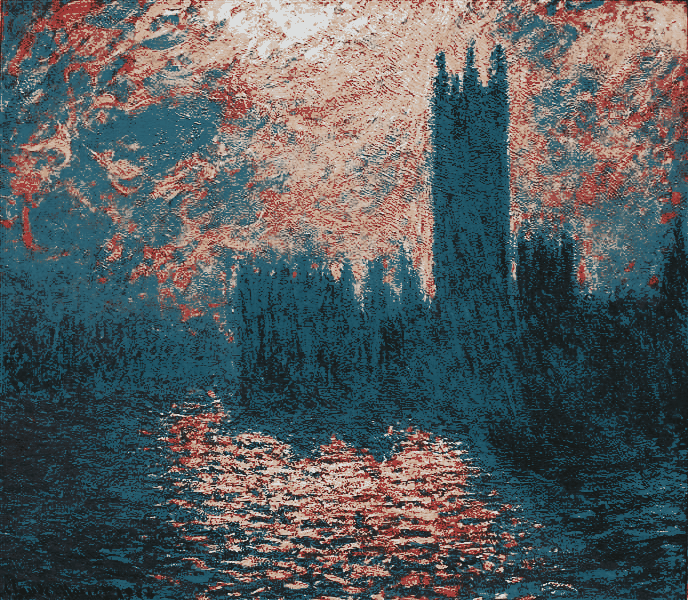} &
        \includegraphics[width=\imw, height=\imw, keepaspectratio=false]{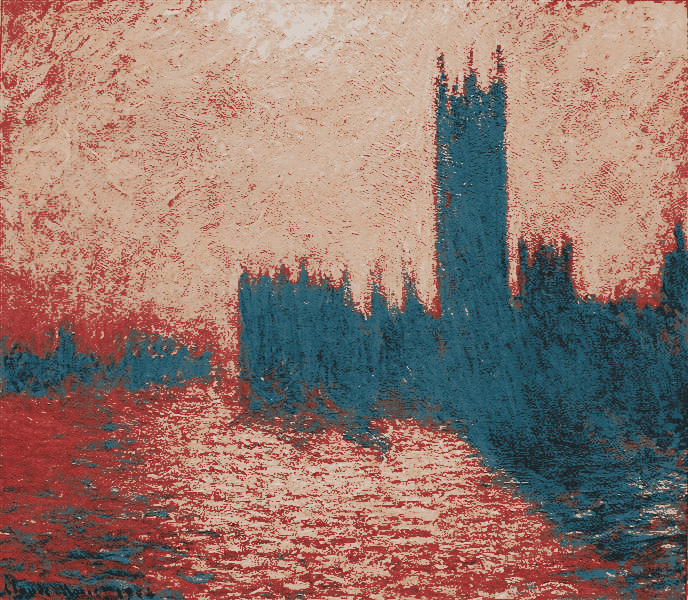} \\[6pt]
        
        \small Min-STP & \small min-SWGG & \small \textbf{RA-OT} (ours) & \small \textbf{OA-OT} (ours) \\[2pt]
        
        \includegraphics[width=\imw, height=\imw, keepaspectratio=false]{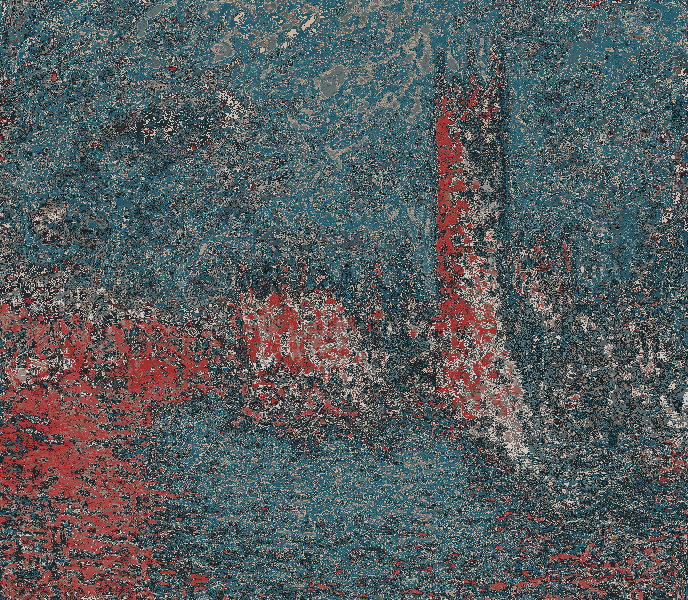} &
        \includegraphics[width=\imw, height=\imw, keepaspectratio=false]{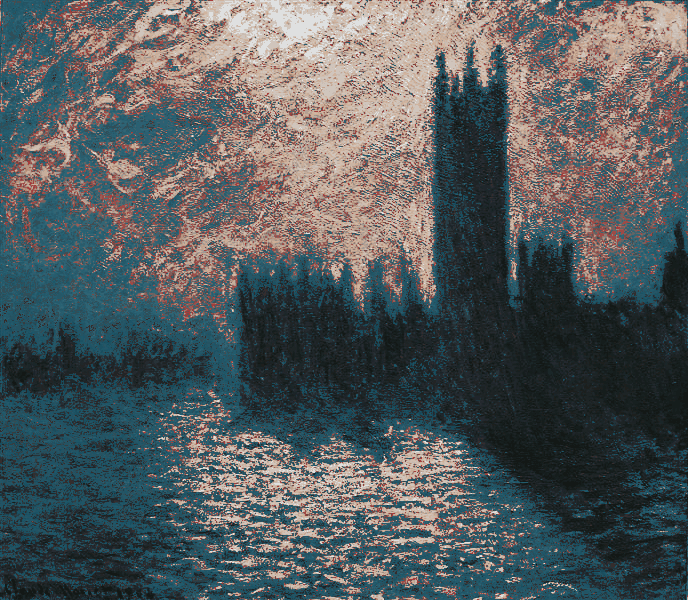} &
        \includegraphics[width=\imw, height=\imw, keepaspectratio=false]{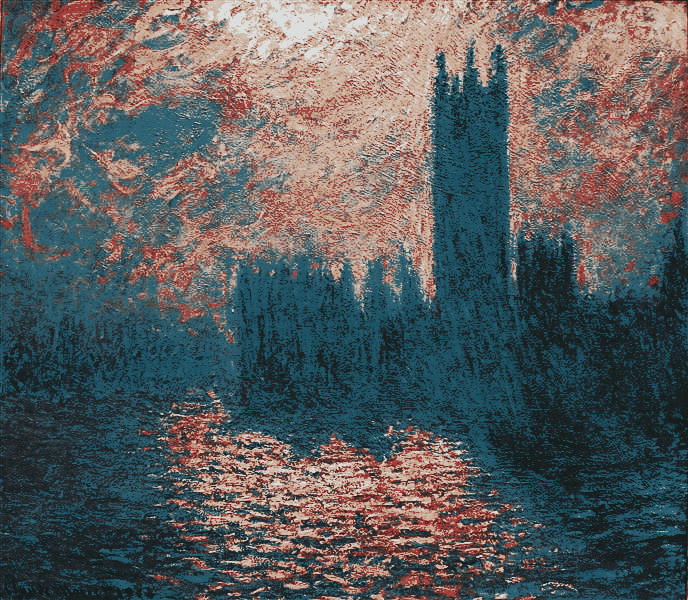} &
        \includegraphics[width=\imw, height=\imw, keepaspectratio=false]{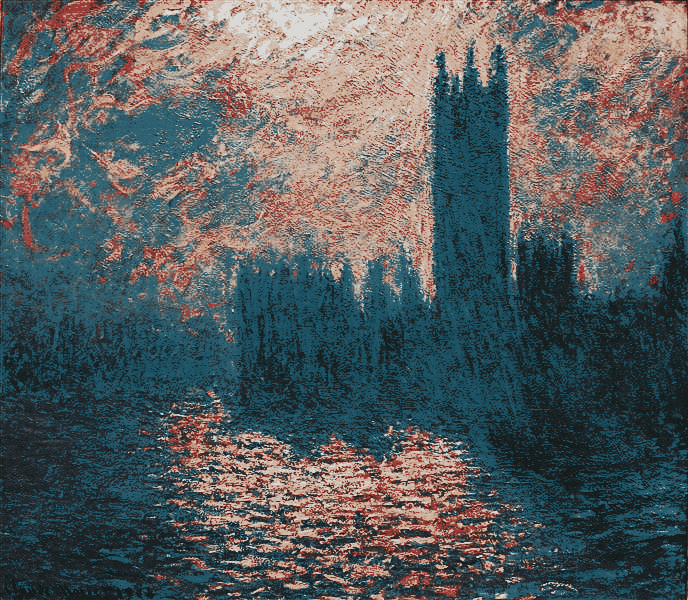} \\[6pt]

    \end{tabular}

    \vspace{0.2cm}
    \caption{Additional qualitative examples of the color transfer task ($M=50, L=100$). The figure presents further comparisons of the visual outputs generated by the evaluated methods when mapping the color palette of the target images onto the source images.}
    \label{fig:color_transfer_all}
\end{figure}

\newpage

\end{document}